\pdfoutput=1
\documentclass{article}

\usepackage[final]{colm2026_conference}

\usepackage[utf8]{inputenc}
\usepackage[T1]{fontenc}
\usepackage{hyperref}
\usepackage{url}
\usepackage{booktabs}
\usepackage{amsfonts}
\usepackage{nicefrac}
\usepackage{xcolor}

\definecolor{darkblue}{rgb}{0, 0, 0.5}
\hypersetup{
colorlinks=true,
citecolor=darkblue,
linkcolor=darkblue,
urlcolor=darkblue
}

\usepackage[utf8]{inputenc}

\usepackage{microtype} 

\usepackage{amsmath}
\usepackage{amssymb}

\usepackage{booktabs}  

\definecolor{red}{rgb}{0.74,0.08,0.10}
\definecolor{green}{rgb}{0.26,0.49,0.18}
\definecolor{blue}{rgb}{0.22,0.53,0.75}
\definecolor{Gray}{gray}{0.9}
\definecolor{LightCyan}{rgb}{0.75,1,1}

\makeatletter
\newcommand\notsotiny{\@setfontsize\notsotiny\@viiipt\@ixpt}
\makeatother

\newcommand{\ie}{\emph{i.e.}, }
\newcommand{\eg}{\emph{e.g.}, }

\newcommand{\makename}[3][s]{%
  \expandafter\newcommand\csname #2\endcsname{#3\xspace}%
  \expandafter\newcommand\csname #2s\endcsname{#3#1\xspace}%
}

\makename{mice}{model-internal confidence estimators}
\makename{miceCaps}{Model-Internal Confidence Estimators}
\makename{miceTitle}{Model-Internal Confidence Estimation}

\makename{logitLens}{logit lens}

\makename{metricName}{expected tool-calling utility}
\makename{metricNameFirstCaps}{Expected tool-calling utility}
\makename{metricNameCaps}{Expected Tool-Calling Utility}
\makename{metricNameAbbr}{ETCU}

\newcommand{\causalmetric}{necessity}
\newcommand{\Causalmetric}{Necessity}

\newcommand{\reuseat}[1]{\textbf{Reuse@}#1}

\usepackage{dsfont}
\newcommand{\indicator}[1]{\mathds{1}\{#1\}}    

\newcommand{\prob}[2][]{p\ifthenelse{\not\equal{}{#1}}{_{#1}}{}(#2)} 
\newcommand{\expect}[2][]{\text{\bf E}\ifthenelse{\not\equal{}{#1}}{_{#1}}{}\!\left[#2\right]}
\newcommand{\var}[2][]{\text{\bf Var}\ifthenelse{\not\equal{}{#1}}{_{#1}}{}\!\left[#2\right]}

\usepackage{listings}

\lstdefinestyle{simple_lst_style}{
    columns=flexible,
    keywordstyle=\color{red},
    numberstyle=\color{gray},
    stringstyle=\color{green},
    basicstyle=\ttfamily\small,
    identifierstyle=\color{black},
    commentstyle=\color{blue},
    breakatwhitespace=false,
    breaklines=true,
    captionpos=b,
    keepspaces=false,
    numbersep=5pt,
    showspaces=false,
    showstringspaces=false,
    showtabs=false,
    tabsize=2,
    frame=single,
}

\lstdefinelanguage{Prompt}{
    morekeywords={Human, Computer},
    sensitive=true,
    morestring=[b]",
}

\usepackage{graphicx}
\usepackage{enumitem}
\usepackage{booktabs}
\usepackage{ifthen}
\usepackage{adjustbox}
\usepackage{multirow}

\usepackage{xspace}

\usepackage[noabbrev, capitalize]{cleveref} 

\crefname{page}{page}{pages}
\crefname{footnote}{footnote}{footnotes}   
\crefname{equation}{Eq.}{Eqs.}
\Crefname{equation}{Eq.}{Eqs.}

\crefformat{equation}{Eq.~#2#1#3}
\Crefformat{equation}{Eq.~#2#1#3}
\crefname{line}{line}{lines}               
\crefrangeformat{line}{lines #3#1#4--#5#2#6}
\crefname{lstlsting}{Listing}{Listings}   
\crefname{section}{\S}{\S\S}
\Crefname{section}{\S}{\S\S}    
\crefformat{section}{#2\S#1#3}  
\Crefformat{section}{#2\S#1#3}
\crefrangeformat{section}{\S\S#3#1#4--#5#2#6}
\Crefrangeformat{section}{\S\S#3#1#4--#5#2#6}
\crefmultiformat{section}{\S#2#1#3}{ and~\S#2#1#3}{, \S#2#1#3}{ and~\S#2#1#3}
\Crefmultiformat{section}{\S#2#1#3}{ and~\S#2#1#3}{, \S#2#1#3}{ and~\S#2#1#3}
\crefrangemultiformat{section}{\S\S#3#1#4--#5#2#6}{ and~\S\S#3#1#4--#5#2#6}{, \S\S#3#1#4--#5#2#6}{ and~\S\S#3#1#4--#5#2#6}
\Crefrangemultiformat{section}{\S\S#3#1#4--#5#2#6}{ and~\S\S#3#1#4--#5#2#6}{, \S\S#3#1#4--#5#2#6}{ and~\S\S#3#1#4--#5#2#6}

\usepackage{subcaption}
\usepackage{wrapfig}
\usepackage{etoc}

\newcommand{\cmulogo}{%
\raisebox{-0.15ex}{%
\includegraphics[height=1.2em]{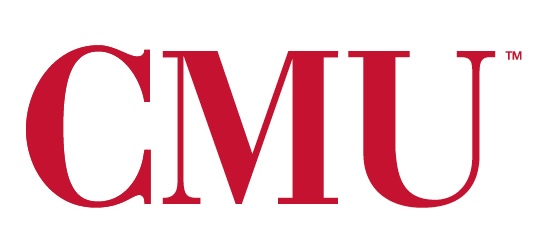}%
}%
\hspace{0.12em}%
\raisebox{0.1ex}{%
\includegraphics[height=0.9em]{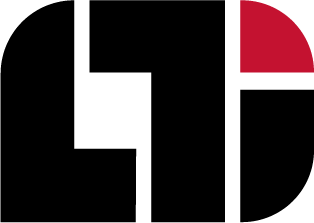}%
}%
}

\newcommand{\northeasternlogo}{%
\raisebox{-0.1ex}{%
\includegraphics[height=1.3em]{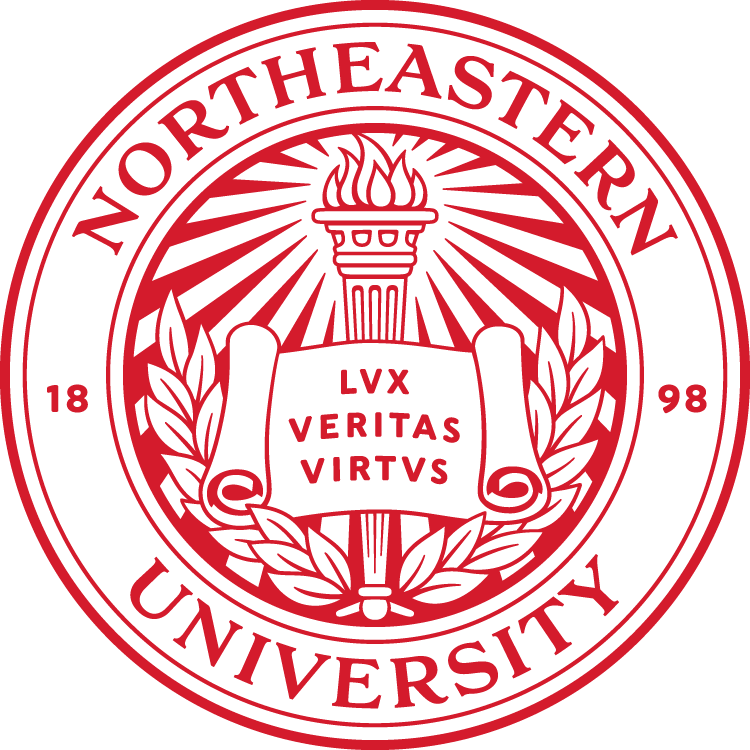}%
}%
}

\title{How Much Do Circuits Tell Us? Measuring the Consistency and Specificity of Language Model Circuits}

\author{
\textbf{Michael Li}\thanks{Work done while at Carnegie Mellon University.}\,$^{\text{\northeasternlogo}}$
\qquad
\textbf{Nishant Subramani}\,$^{\text{\cmulogo}}$
\\[0.4em]
$^{\text{\northeasternlogo}}$\,Northeastern University
\\[0.15em]
$^{\text{\cmulogo}}$\,Carnegie Mellon University -- Language Technologies Institute
\\[0.35em]
\texttt{\{li.michael@northeastern.edu, nishant2@cs.cmu.edu\}}
}

\begin{document}


\maketitle
\etocdepthtag.toc{mainbody}

\begin{abstract}
The circuits framework in mechanistic interpretability aims to identify sparse subgraphs of model components that are causally responsible for a behavior, typically evaluated by measuring \emph{necessity} and \emph{sufficiency}.
But these criteria say little about whether a circuit consistently captures how a model performs a task, or if it is specific to that task.
We study these two properties, \emph{consistency} and \emph{specificity}, across six tasks and five models, extracting circuits at the component level (attention heads and MLP blocks) and at the level of individual MLP neurons.
We find that component-level circuits are highly consistent and causally important for most tasks, but they are not specific: for a given task, ablating its own circuit drops accuracy by about as much as ablating another task's circuit.
Neuron-level circuits, on the other hand, exhibit higher specificity across tasks, but far less consistency within tasks.
This is explained by circuit overlap: component-level circuits share most of their components across tasks, related or not, while neuron-level circuits overlap only between closely related tasks.
In a case study of the components shared by the task circuits of \texttt{Llama-3.2-3B}, we show that they consist mostly of MLP blocks, while the few attention heads within turn out to have general-purpose roles.
Overall, our findings raise questions about the degree to which circuits can support targeted understanding of, and intervention on, model behavior.
\end{abstract}
\section{Introduction}
Neural networks are infamously black box; even when we can elicit strong performance on a task, it is unclear which internal computations are responsible. 
Mechanistic interpretability seeks to reverse-engineer these computations by identifying \emph{circuits}: sparse subgraphs of model components that are causally responsible for a particular behavior~\citep{elhage2021mathematical,wang2022interpretability}.
A growing body of work has developed methods for extracting such circuits~\citep{syed-etal-2024-attribution,marks2025sparse,jafari2025relp} and evaluating their \emph{necessity} (removing the circuit should degrade performance) and \emph{sufficiency} (the circuit alone should reproduce the behavior;~\citealp{shi2024hypothesis}).

\begin{figure*}[t]
\centering
\includegraphics[width=\textwidth]{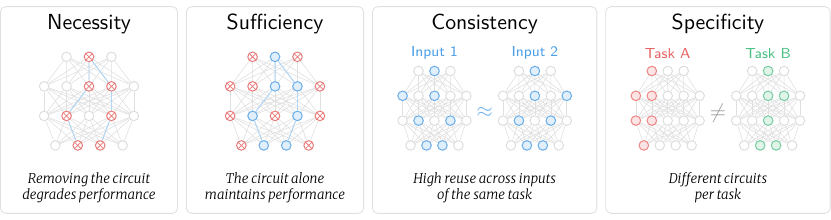}
\caption{\textbf{Circuit evaluation criteria.} We propose evaluating circuits for consistency across inputs and specificity across tasks, in addition to necessity and sufficiency.}
\label{fig:overview}
\end{figure*}

We argue that there are two additional properties crucial to consider (\cref{fig:overview}).
First, circuits should be \emph{consistent}: most of the same units should recur across individual examples of a task, or else the aggregated task circuit is too noisy to describe the task.
Second, circuits should be \emph{specific}: a task's circuit should be meaningfully different from the circuits of unrelated tasks, since a circuit that surfaces for any task says little about how the model solves one task in particular.

We test both properties at scale, using two automated methods for circuit discovery: edge attribution patching with integrated gradients (EAP-IG; \citealp{hanna2024have}), and relevance patching (RelP; \citealp{jafari2025relp}).
For each method, we extract circuits at the component level (attention heads and MLP blocks) and at the level of individual MLP neurons~\citep{arora2026language}, for $n$=1000 examples across six tasks spanning algorithmic reasoning (\texttt{Addition}, \texttt{Boolean Logic}), information retrieval (\texttt{IOI}, \texttt{CopyColors MCQA}), and knowledge-intensive benchmarks (\texttt{ARC Easy}, \texttt{ARC Challenge}). We study five models from three architecture families (\texttt{Gemma~2}, \texttt{Llama~3.2}, \texttt{Qwen3}), and additionally report results on \texttt{OLMo-2-1B} pre-training checkpoints. We find the following:

\begin{enumerate}[itemsep=2pt]
    \item \textbf{Component-level circuits are consistent but not specific.}     
    Across tasks and models, a large fraction of each per-example circuit is drawn from the same set of components, and ablating this set (knocking them out) causes larger accuracy drops than ablating a capacity-matched random set of the same size. But these components are not task-specific: ablating one task's circuit damages an unrelated task about as much as ablating that task's own circuit does. This is explained by the substantial overlap between component-level circuits, which are composed of largely the same MLP blocks.
    \item \textbf{Neuron-level circuits are specific but not consistent.} 
    Circuits derived from individual MLP neurons tell a different story. A task is damaged far more by ablating its own neuron-level circuit than by ablating any other task's circuit, but the neurons that recur across examples of a task cover only a small fraction of each per-example circuit. Because there are far more MLP neurons than transformer components, these circuits are also much smaller as a fraction of the model.
\end{enumerate}

We conclude that component-level circuits are too coarse to isolate task-specific mechanisms, while neuron-level circuits are too fine-grained to characterize the task in general.
We discuss several possible reasons for this - such as the role of shared MLP blocks and superposition/polysemanticity~\citep{elhage2022superposition} - and present a case study of the components shared across tasks in \texttt{Llama-3.2-3B}.
We also find preliminary evidence that feature-level circuits~\citep{dunefsky2024transcoders,marks2025sparse,ameisen2025circuit} do not recover these properties either (\cref{sec:sae_features}).
Finally, we discuss implications for applications that assume circuit-level modularity, including model editing~\citep{meng2022locating,dai2022knowledge} and safety interventions~\citep{li2024inference}.

\section{Background}
\label{sec:background}

\subsection{Transformer Circuits}

We adopt the residual-stream view of transformers introduced by \citet{elhage2021mathematical}, along with the computational graph formulation of \citet{conmy2023towards}, which represents decoder-only transformers as a directed acyclic computational graph.
Under the Transformer Circuits view, the \emph{residual stream} is interpreted as a shared communication channel running through the model. The token embeddings are written into it first, after which each attention head and MLP reads from the stream, performs a computation, and additively writes its output back.
Because these contributions are additive, the output of any component can influence any downstream component, resulting in connections between non-adjacent layers.
This naturally induces a graph in which nodes are connected whenever the output of one node can contribute to another. Following \citet{elhage2021mathematical}, nodes are commonly taken to be attention heads and MLP blocks.
Recent work has introduced other decompositions, however, such as individual neurons or sparse autoencoder features, each with its own  benefits~\citep{marks2025sparse,ameisen2025circuit,arora2026language}.
Edges represent the flow of information between components through the residual stream.
A \emph{circuit} is then a sparse subgraph of this computational graph that is responsible for a given behavior~\citep{wang2022interpretability,conmy2023towards}.

\subsection{Attribution Methods}

To identify components which are important for a given behavior, researchers typically use activation patching~\citep{vig2020causal,meng2022locating,wang2022interpretability}, which compares a \emph{clean} input that elicits the behavior of interest with a corresponding \emph{corrupted} input in which that behavior is different. Then, individual component's activation from the corrupted run are \emph{patched} to the clean run. The resulting change is then measured and used to estimate the components' importance.
However, this requires a separate forward pass per component, which is expensive.
Edge attribution patching with integrated gradients (EAP-IG;~\citealp{hanna2024have}) instead efficiently approximates the effect of patching a component by taking the product of its activation difference and the gradient of the metric, averaged over $m$ points along the straight-line path between the corrupted and clean inputs.
This allows all components to be scored simultaneously, making it suitable for the large-scale analysis we conduct here.
Relevance patching (RelP;~\citealp{jafari2025relp}) computes the same product, but replaces the gradient with a layer-wise relevance propagation signal that conserves attribution from one layer to the next, requiring a single backward pass.
See \cref{sec:attribution_details} for full details of both methods.
\section{Methodology}
\label{sec:method}

\subsection{Tasks and Models}

We evaluate six tasks spanning algorithmic reasoning (\texttt{Addition}, \texttt{Boolean Logic}), information retrieval from context (\texttt{IOI};~\citealp{wang2022interpretability}, \texttt{CopyColors MCQA};~\citealp{mueller2025mib}), and knowledge-intensive benchmarks (\texttt{ARC Easy}, \texttt{ARC Challenge};~\citealp{clark2018arc}).
We study five models from three architecture families: \texttt{Gemma~2} (2B, 2B~IT;~\citealp{gemmateam2024gemma2improvingopen}), \texttt{Llama~3.2} (3B, 3B~Instruct;~\citealp{grattafiori2024llama3herdmodels}), \texttt{Qwen3} (4B;~\citealp{qwen3}), as well as \texttt{OLMo-2-1B}~\citep{walsh2025}, which is used for pre-training analysis.

\subsection{Circuit Discovery}

For each task $T$, let $\mathcal{D}_T^{\mathrm{train}} = \{(x_i,y_i)\}_{i=1}^{n_{\mathrm{train}}}$ denote the training set from which circuits are extracted, and let $\mathcal{D}_T^{\mathrm{eval}}$ denote a disjoint held-out evaluation set on which all ablation accuracies are measured. Throughout, we use $n_{\mathrm{train}}=1000$ and $n_{\mathrm{eval}} =|\mathcal{D}_T^{\mathrm{eval}}|=200$. Here, $x_i$ denotes the input prompt and $y_i$ the target answer token. These are paired with a corrupted input $x_i'$ that minimally alters $x_i$, so that $y_i$ is different.\footnote{Corruption strategies are described in~\cref{sec:task_details}.}

Let $\mathcal{C} = (\mathcal{V}, \mathcal{E})$ be the model's computation graph, where $\mathcal{V}$ is the set of model components and $\mathcal{E}$ are the connections between them. For notational simplicity, we identify a circuit with its selected node set; the corresponding subgraph is obtained by inheriting the relevant edges from $\mathcal{C}$.
For each $(x_i, y_i) \in \mathcal{D}_T^{\mathrm{train}}$, we extract a per-input circuit $\mathcal{C}_i \subseteq \mathcal{V}$ consisting of the top-$K\%$ of nodes according to their attribution scores, computed using EAP-IG or RelP. We sweep $K \in \{1, 5, 10, 20, 30\}$.

We instantiate $\mathcal{V}$ at two levels of granularity. At the \emph{component} granularity, nodes are attention heads and whole MLP blocks. Our models have between 234 (\texttt{Gemma~2~2B}) and 1{,}188 (\texttt{Qwen3~4B} of these components).
At the \emph{neuron} granularity, nodes are the pre-down-projection MLP activations of each MLP block.\footnote{We discuss our motivation for this choice over other units, \eg transcoder latents, in \cref{sec:limitations}, which is based on findings from \citet{arora2026language}.}
With $d_{\mathrm{mlp}}$ neurons per block, this gives between 229{,}376 and 350{,}208 nodes in total, corresponding to a node set approximately 300--1{,}000 times as large as at the component granularity.

Using the per-input circuits $\mathcal{C}_i$, we define a task's circuit at consensus threshold $P \in [0,1]$ as
\begin{equation}
    S_P = \Bigl\{\, c \in \mathcal{V} \,:\, \tfrac{1}{n_{\mathrm{train}}}\textstyle\sum_{i=1}^{n_{\mathrm{train}}} \indicator{c \in \mathcal{C}_i} \,\geq\, P \,\Bigr\}.
\end{equation}
Thus, $S_P$ contains exactly the components that appear in at least $P\times100\%$ of the per-input circuits.\footnote{We report $P$ as a percentage in the text and figures.}

\subsection{Within-Task Experiments}
We define $\reuseat{P}$ as the mean fraction of a per-input circuit overlapping with the task circuit,
\begin{equation}
    \reuseat{P} = \frac{1}{n_{\mathrm{train}}}\sum_{i=1}^{n_{\mathrm{train}}} \frac{|S_P \cap \mathcal{C}_i|}{|\mathcal{C}_i|},
    \label{eq:overlap}
\end{equation}
Larger values indicate greater reuse of circuit components across examples at the specified consensus threshold.
We sweep $P \in \{50\%, 75\%, 85\%, 90\%, 95\%, 96\%, 97\%, 98\%, 99\%, 100\%\}$, but report $P$=50\% in the main text, since it is the only threshold at which the neuron-level circuit reuse is non-zero for most model-task pairs. The full sweep is reported in \cref{sec:full_within}.

We next test whether the task circuit is causally important to task performance. We zero-ablate $S_P$ and measure the accuracy drop on $\mathcal{D}_T^{\mathrm{eval}}$, formalizing this via do-calculus~\citep{pearl1995docalculus}. 
For a circuit $S \subseteq \mathcal{V}$, let $\mathrm{do}(S \leftarrow 0)$ denote the intervention that clamps the activations associated with every $c \in S$ to zero.\footnote{Zero-ablation is one of several methods used in the literature~\citep{wang2022interpretability,zhang2024towards,li2024optimal}. In \cref{sec:ablation_robustness} we repeat our experiments using mean ablation and find that our conclusions still hold.}
The model's output distribution on input $x$ under this intervention is $p_{\mathcal{M}}(\,\cdot \mid x;\, \mathrm{do}(S \leftarrow 0))$. Taking $S = \varnothing$ recovers the output distribution of the unablated model.

We define the \textbf{zero-ablated prediction} (ZAP) of $\mathcal{M}$ under ablation set $S$ to be the model's maximum-probability token under the intervention:\footnote{We use argmax decoding here, but any decoding rule could in principle be applied to $p_{\mathcal{M}}$.}
\begin{equation}
    \mathrm{ZAP}(\mathcal{M}, S, x) = \arg\max_{y'} p_{\mathcal{M}} \bigl(y' \mid x;\, \mathrm{do}(S \leftarrow 0) \bigr),
    \label{eq:zap}
\end{equation}
Accuracy of the model on task $T$ under ablation $S$ is then \footnote{$S$, $\mathrm{acc}$, and~\causalmetric~all depend implicitly on $T$ through $\mathcal{D}_T^{\mathrm{train}}$ and $\mathcal{D}_T^{\mathrm{eval}}$. We drop $T$ from the notation for readability.}
\begin{equation}
    \mathrm{acc}_T(\mathcal{M}, S) = \frac{1}{n_{\mathrm{eval}}}\sum_{(x, y)\, \in\, \mathcal{D}_T^{\mathrm{eval}}} \indicator{y = \mathrm{ZAP}(\mathcal{M}, S, x)}.
    \label{eq:acc}
\end{equation}
However, removing any set of components will reduce network capacity, so some degradation is expected even when the ablated components are not task-relevant. To account for this, we compare the consensus-circuit ablation against a \textbf{capacity-conserved control}, denoted $S_{C^3}$. The control is randomly sampled from $\mathcal{V} \setminus S_P$ and matched to $S_P$ both in cardinality and in component-type composition.
The difference in accuracy drop between $S_P$ and $S_{C^3}$ should therefore capture the effect of ablating the task circuit beyond degradation lost due to capacity. We define the causal effect of $S_P$ on task $T$ as
\begin{equation}
    \mathrm{\causalmetric}(\mathcal{M}, S_P)
    =
    \frac{\mathrm{acc}(\mathcal{M}, S_{C^3}) - \mathrm{acc}(\mathcal{M}, S_P)}{\mathrm{acc}(\mathcal{M}, \varnothing)}.
    \label{eq:causal_effect}
\end{equation}
A positive \causalmetric~means ablating $S_P$ hurts more than ablating $C^3$, which we interpret as evidence that the components shared across examples are \emph{causally important} for task $T$.
 
\subsection{Cross-Task Experiments}
 
The experiments above establish whether a task's circuit is consistent across inputs and causally important for that task, but they do not tell us whether that circuit is \emph{specific} to that task.
To probe specificity, we conduct three experiments. First, we compare the effect of ablating a task's own circuit with the effects of ablating other tasks' circuits. Second, we quantify the overlap between pairs of task circuits. Finally, we partition pairs of circuits into shared and task-exclusive components to localize any task-specific causal signal.
 
First, define $\Delta_A^B = \mathrm{acc}_A(\mathcal{M}, \varnothing) - \mathrm{acc}_A(\mathcal{M}, S_P^B)$, the accuracy drop on task $A$ when ablating task $B$'s circuit, where $\mathrm{acc}_A$ denotes accuracy evaluated on $\mathcal{D}_A^{\mathrm{eval}}$ and $S_P^B$ is task $B$'s circuit. If circuits are task-specific, ablating task $A$'s own circuit should hurt task $A$ more than ablating task $B$'s circuit. For each task and model, we therefore compare $\Delta_A^A$ against $\frac{1}{|\mathcal{T}|-1}\sum_{B \in \mathcal{T}\setminus\{A\}}\Delta_A^B$, the mean accuracy drop on $A$ when ablating each other task's circuit. We summarize this comparison with the \emph{specificity gap}, $\Delta_A^A - \frac{1}{|\mathcal{T}|-1}\sum_{B \in \mathcal{T}\setminus\{A\}}\Delta_A^B$, measured in percentage points of accuracy. A gap near zero indicates that ablating a task's own circuit is no more damaging to that task than ablating other tasks' circuits.

We then measure the similarity between two tasks' circuits using the Jaccard index $J(S_P^A, S_P^B) = |S_P^A \cap S_P^B| \,/\, |S_P^A \cup S_P^B|$, which is 1 when the two circuits are identical and 0 when they are disjoint.

Finally, we ask whether any task-specific causal signal is concentrated in components unique to a task or in components shared across tasks. For each task pair $(A,B)$, we partition $S_P^A \cup S_P^B$ into three disjoint sets: the \emph{shared core} $S_P^A \cap S_P^B$, the \emph{$A$-only} set $S_P^A \setminus S_P^B$, and the \emph{$B$-only} set $S_P^B \setminus S_P^A$. We ablate each set separately and measure its effect on task $A$. We then report the accuracy drop on the target task alongside the corresponding mean drop on the other tasks. We also compare the $A$-only set with a capacity-conserved control, constructed analogously to $S_{C^3}$ above.
Unlike our other experiments, we run these at $P$=100\% rather than $P$=50\%, and only at the component granularity, since at the neuron granularity the shared core is nearly empty, whereas component-level circuits remain non-empty even at $P$=100\%.

\begin{figure*}[t]
  \centering
  \includegraphics[width=\textwidth]{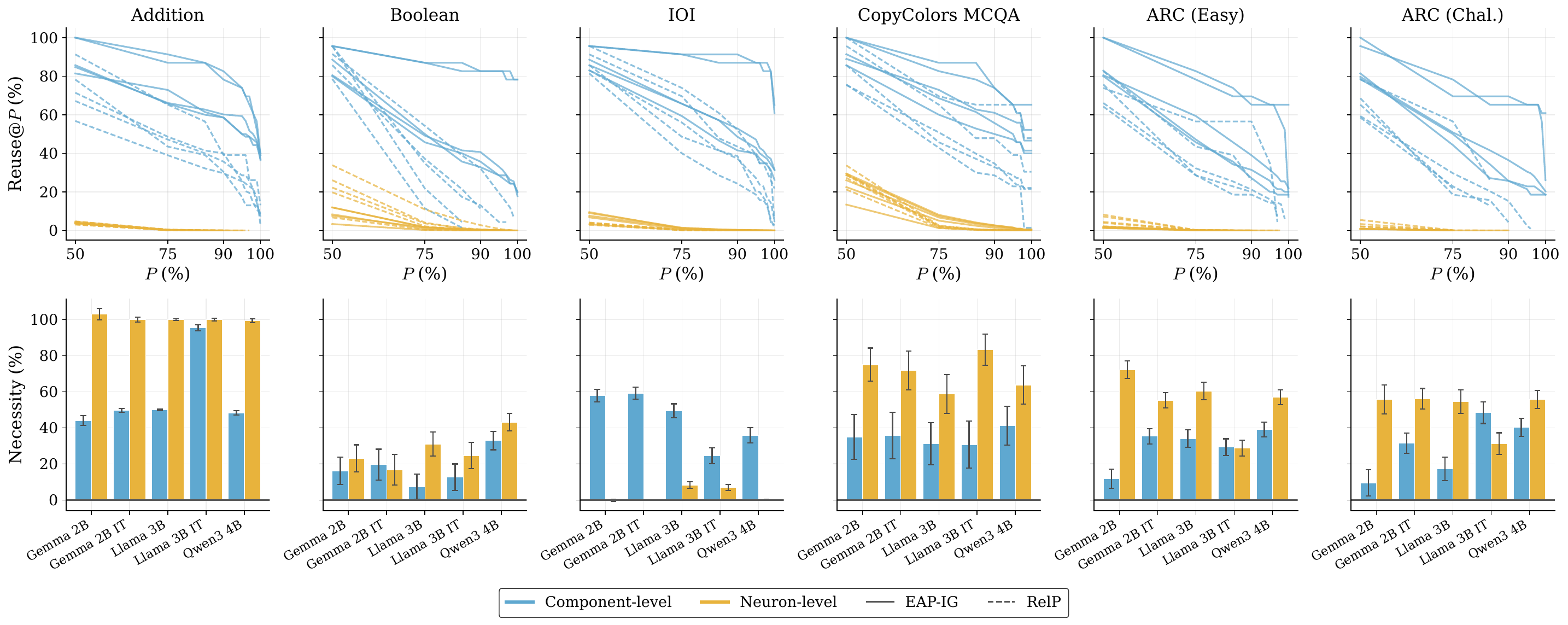}
  \caption{\textbf{Within-task circuit reuse and \causalmetric.} \emph{Top:}~\reuseat{P} at top-$K$=10\%, the fraction of each example's circuit covered by components that appear in at least $P$\% of examples, as $P$ sweeps from 50\% to 100\%. Curves show different models and attribution methods. \emph{Bottom:}~\Causalmetric~of circuits (at top-$K$=10\% and $P$=50\%) measures the difference in accuracy drop between ablating the task circuit and ablating the control set ($S_{C^3}$), with 95\% C.I.'s.}
  \label{fig:reuse_necessity}
\end{figure*}
\section{Within-Task Consistency}
\label{sec:within_task}

We now present the results of our within-task experiments, looking first at circuit reuse and \causalmetric, before examining the structure of these circuits, and how they evolve during \texttt{OLMo-2-1B} pretraining.

\paragraph{Component-level circuits reuse the same components across inputs.}
If a circuit captures how a model solves a task, most of the same components should appear regardless of which input is used.
\Cref{fig:reuse_necessity} (top row) shows \reuseat{P}, the average fraction of a per-example circuit that is covered by the task's consensus circuit (\cref{eq:overlap}).
At $K$=10\% and $P$=50\%, component-level circuits show high reuse across inputs, with a median \reuseat{P} of 88.6\% under EAP-IG and 78.4\% under RelP.
Reuse falls as $P$ becomes more stringent: under EAP-IG the mean is 60.8\% at $P$=85\% and 38.5\% at $P$=100\% (where a component must appear in every example's circuit).
Across the broader top-$K$ and $P$ sweeps, reuse remains high: under EAP-IG the median rises to 98.0\% at $K$=30\%, and no model--task pair falls below 50.0\% (\cref{fig:reuse_sweep,fig:reuse_all_eap_ig_head_mlp}).

\paragraph{Neuron-level circuits (mostly) do not reuse the same neurons across inputs.}
At the level of individual MLP neurons, per-example circuits are far less consistent.
At $K$=10\% and $P$=50\%, neuron-level circuits show little reuse across inputs, with a median \reuseat{P} of 3.9\% under EAP-IG and 4.5\% under RelP (\cref{fig:reuse_necessity}).
Neuron-level reuse is below 2\% by $P$=75\% and effectively zero by $P$=85\% (\cref{fig:reuse_necessity}), and remains much lower than component-level reuse at every circuit size for both attribution methods (\cref{sec:full_within,fig:reuse_sweep}).
Although neuron-level reuse is low throughout, it does vary, with \texttt{CopyColors~MCQA} and \texttt{Boolean~Logic} showing higher reuse than others (\cref{fig:reuse_necessity}).

\paragraph{Circuits are often causally important, but this varies by task.}
\Cref{fig:reuse_necessity} (bottom row) shows the \causalmetric~of each task's circuit (\cref{eq:causal_effect}).
At $K$=10\% and $P$=50\%, \causalmetric~is clearly positive for \texttt{IOI}, \texttt{Addition}, \texttt{CopyColors~MCQA}, \texttt{ARC~Easy}, and \texttt{ARC~Challenge}, but is low for \texttt{Boolean~Logic} at both granularities and for \texttt{IOI} at the neuron level.\footnote{All models have above chance baseline accuracy (\cref{tab:baseline_accuracy})}
Therefore, reuse alone does not necessarily establish that a consensus circuit matters for the task.
Neuron-level circuits exhibit far lower reuse than component-level circuits, but are still causally important at $P=50\%$, with higher \causalmetric~than component-level circuits.\footnote{At $P=100\%$, neuron-level circuits are nearly empty (\cref{fig:circuit_size}), causing \causalmetric~to drop from about 51 points at $P=50\%$ to 5--19 at $P=85\%$ (\cref{fig:necessity_sweep_neuron}).}
Conversely, component-level circuits have high reuse but low \causalmetric~for \texttt{Boolean~Logic}.
The full sweeps across top-$K$ and $P$ are reported in \cref{sec:full_within}.

\begin{figure*}[t]
  \centering
  \includegraphics[width=\textwidth]{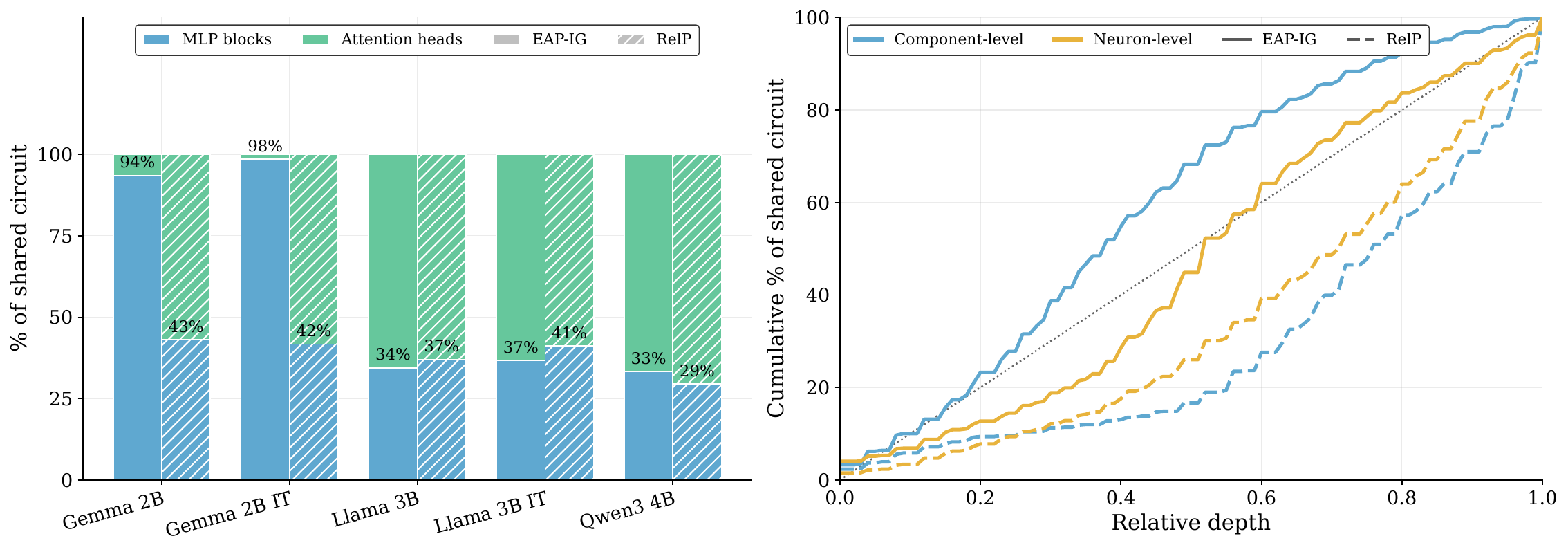}
  \caption{\textbf{Composition and depth of task circuits at $K$=10\%, $P$=50\%.} \emph{Left:}~the fraction of each component-level circuit made up of MLP blocks and attention heads. \emph{Right:}~the cumulative fraction of circuits at or below each relative depth. See \cref{sec:layer_distribution_cdf} for per-model and per-task versions.}
  \label{fig:circuit_anatomy}
  \vskip -1em
\end{figure*}

\paragraph{Different attribution methods find different circuits.}
EAP-IG and RelP scores are only weakly correlated (mean Spearman $\rho=0.36$ between the the two methods' component scores on the same example), and their circuits for the same task overlap with a Jaccard index of only 0.23 at the component level and 0.16 at the neuron level ($K$=10\%, $P$=50\%, per model and task in \cref{fig:method_agreement}), less than the EAP-IG circuits of two different tasks (0.56, \cref{sec:full_cross}).
The two methods' circuits also differ in structure (\cref{fig:circuit_anatomy}).
Relative to their base rates, MLPs are substantially overrepresented (\eg, in \texttt{Llama-3.2-3B}, they are only $\sim$4\% of nodes but up to 41\% of a circuit), most of all in the EAP-IG circuits of the \texttt{Gemma-2} models (93.5\% and 98.5\% MLP blocks).
Circuits are mostly distributed in the later layers, although EAP-IG draws more from earlier layers than RelP (see \cref{sec:circuit_anatomy} for sizes, composition tables and per-model depth distributions).

\paragraph{Consistency emerges early and degrades over training.}
Tracking circuit reuse across \texttt{OLMo-2-1B}'s full pretraining run shows that reuse is highest within the first 32B tokens and declines over the rest of pretraining (\cref{sec:pretraining_appendix,fig:pretraining}). We also measure \causalmetric, but find it hard to interpret as baseline accuracy is around chance for most of training. See \cref{sec:pretraining_appendix} for full results.

\section{Cross-Task Specificity}
\label{sec:cross_task}
We have shown that component-level circuits are consistent while neuron-level circuits are not, but that circuits at both granularities are causally important.
However, a component may be critical for task $A$ only because it is critical for every task, revealing little about how the model performs $A$ itself.
We now ask whether the circuits identified for one task are \emph{specific} to that task.

\begin{figure*}[t]
    \centering
    \includegraphics[width=\textwidth]{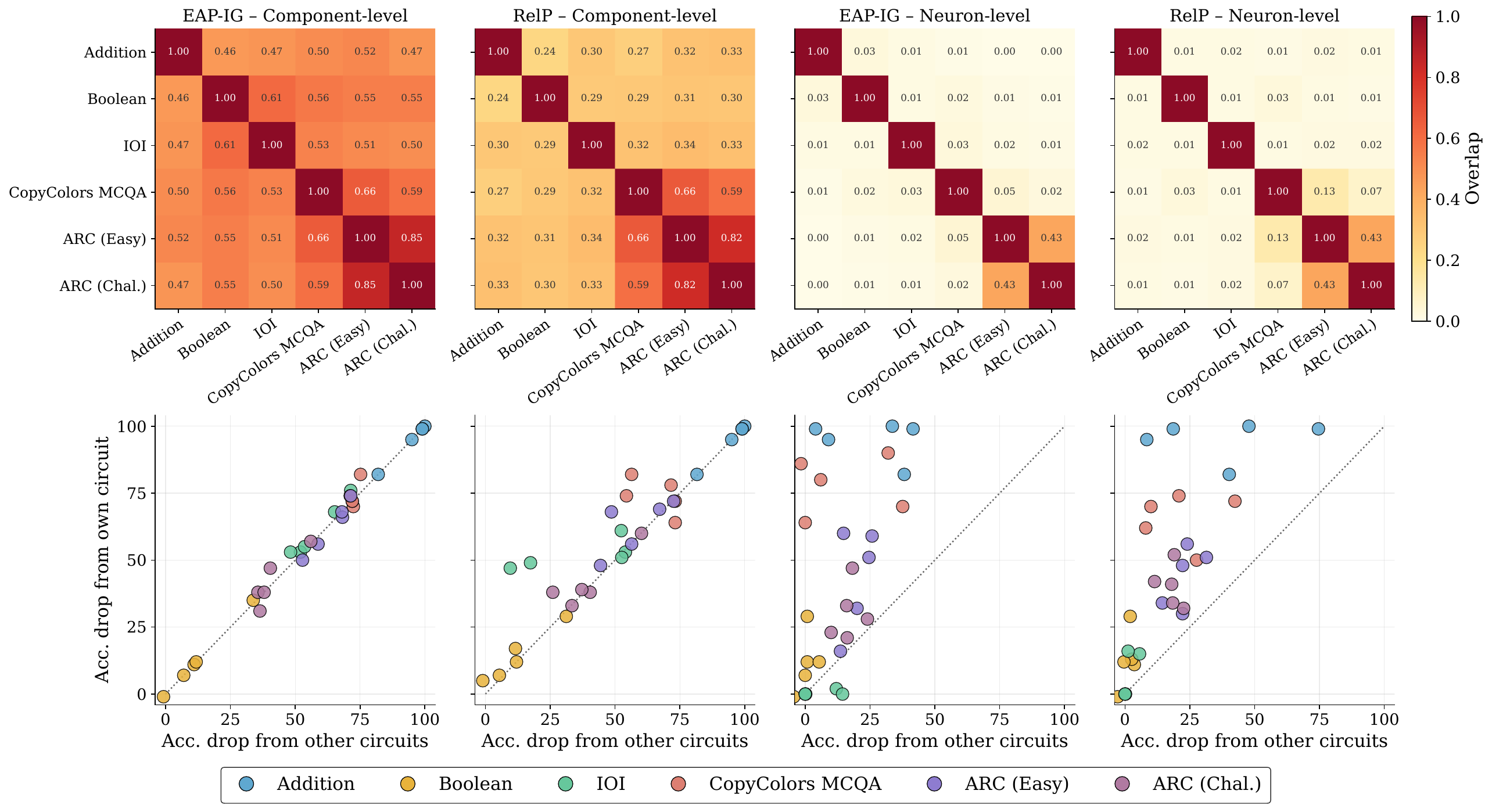}
    \caption{\textbf{Cross-task overlap and accuracy drops at $K$=10\%, $P$=50\%.} \emph{Top:}~Jaccard overlap between each pair of tasks' circuits, averaged over the five models. \emph{Bottom:}~For every model and task, we plot the accuracy drop from ablating that task's own circuit against the mean accuracy drop from ablating the other tasks' circuits. Per-model overlap across $K$ and $P$ and drops at every $K$ are in \cref{sec:full_cross}.}
    \label{fig:overlap_own_other}
\end{figure*}

\begin{figure*}[t]
    \centering
    \includegraphics[width=\textwidth]{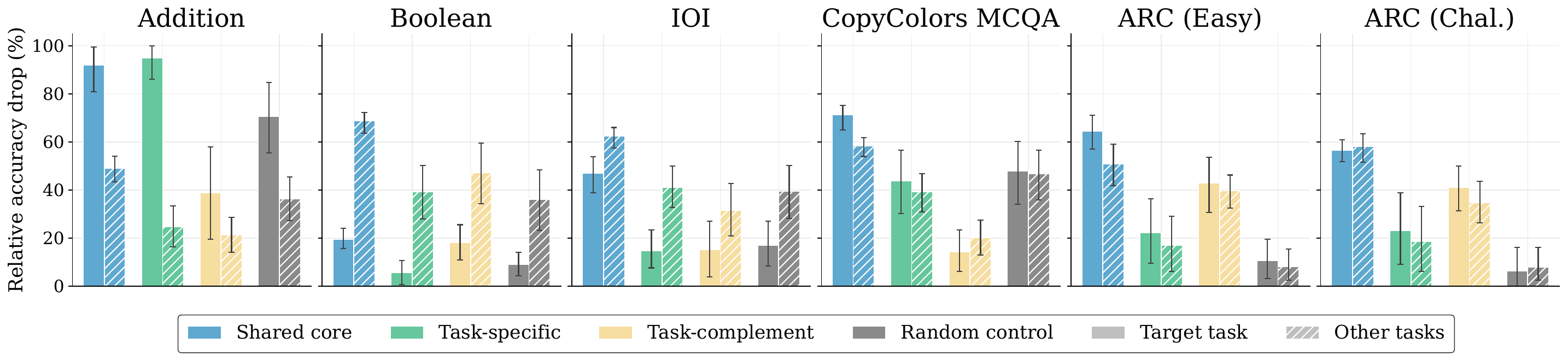}
    \caption{\textbf{Selective ablation of the component-level circuit partitions at $K$=10\%, $P$=100\%.} Solid bars represent the drop in accuracy on the target task (relative to baseline performance) and hatched bars the mean drop on the other five, averaged across models. See \cref{sec:selective_all_k} for all values of $K$.}
    \label{fig:selective}
    \vskip -1em
\end{figure*}

\paragraph{Neuron-level circuits are task-specific, while component-level circuits are not.}
If a circuit is specific to its task, ablating it should damage that task more than any other task's circuit.
\Cref{fig:overlap_own_other} (bottom) plots, for each task and model, the drop from ablating that task's own circuit against the mean drop from ablating the other five tasks' circuits, across granularity and attribution method.
We observe that component-level circuits under EAP-IG cluster along the diagonal (specificity gap of 0.7 points), and just above it for RelP (specificity gap of 5.4 points).
In contrast, neuron-level circuits fall well above the diagonal, with specificity gaps of 29.4 points for EAP-IG and 26.8 points for RelP.
The exception is \texttt{IOI}, whose neuron-level circuit leaves \texttt{IOI} accuracy unchanged when ablated.

\paragraph{Specificity is correlated with overlap between different task's circuits.}
Suppose two tasks' circuits were to overlap heavily. Then, if we ablated either circuit, we would expect this to cause similar drops in performance on the same task. We find that this is indeed the case: \Cref{fig:overlap_own_other} (top) shows the Jaccard overlap between each pair of tasks' circuits at the component- and neuron-level.
Component circuits overlap heavily under EAP-IG, and under RelP they have smaller but still substantial overlap.
For neuron circuits however, overlap is near zero for nearly every pair, with the exception of \texttt{ARC Easy} and \texttt{ARC Challenge}, which is unsurprising given that they come from the same benchmark.

\paragraph{Components exclusive to a task's circuit are not strongly task-specific.}
Although there is high overlap between component-level tasks, some components remain exclusive to a single task.
To test whether these components pass our specificity check, we split each pair of task circuits into a shared core, a task-specific set, and a task-complement set, and ablate each one on its own (\cref{fig:selective}). 
Ablating the shared core causes the largest drop on every task except \texttt{Addition}, consistent with our prior findings of low specificity.
The task-specific set, meanwhile, damages the other tasks about as much as its own or more, and for \texttt{Boolean~Logic}, \texttt{IOI} and \texttt{CopyColors~MCQA} it even has low \causalmetric.
\texttt{Addition} is the only exception, where the task-specific set removes 95\% of accuracy on its own task against 25\% on the others.
Conceptually, we hypothesize that these components are either noise from the attribution method, or that they are relevant to the task but not sufficient on their own to support it (see~\cref{sec:selective_all_k} for more details).

\paragraph{Sparse feature circuits in \texttt{Gemma-2-2B} are similar to neuron circuits.}
We extend our analysis beyond components and neurons to sparse autoencoder latents (\ie features), repeating our experiments on \texttt{Gemma-2-2B} with latents as the circuit units, following \citet{marks2025sparse} and using the Gemma Scope SAEs~\citep{lieberum-etal-2024-gemma}.
At $K$=10\% and $P$=50\%, feature circuits are only slightly more consistent than neuron circuits and both are far less consistent than component circuits.
Feature circuits also match neuron circuits in \causalmetric~on every task and are as task-specific, with an average specificity gap of 28.3\% of baseline accuracy against 29.7\% for neurons, while component circuits show no gap.
We discuss full results in \cref{sec:sae_features}.

\subsection{Case study of shared attention heads in \texttt{Llama-3.2-3B}}
\label{sec:case_study}

We now conduct a basic case study on \texttt{Llama-3.2-3B}.
Intersecting the task circuits under EAP-IG at $K$=10\% and $P$=50\%, we recover a set of 17 components, of which 14 are MLP blocks and three are attention heads.
One of the heads, L4H10, is an attention sink~\citep{xiao2024efficient}, sending 7.9 times more attention to the beginning-of-text token than to the body of the prompt, while the other two attend to task-relevant tokens regardless of the task.
This is preliminary evidence that the shared core contains general-purpose machinery, though we leave an analysis of the MLP blocks and other models to future work.
We include full details in \cref{sec:case_study_details}.

\section{Discussion}
\label{sec:discussion}

\subsection{Why do circuits overlap?}

\paragraph{MLP blocks as shared infrastructure.} In our \texttt{Llama-3.2-3B} case study, 14 of the 17 components shared by all six task circuits are MLP blocks. These layers plausibly perform general-purpose operations -- storing parametric knowledge~\citep{sun2025redeep, liu2025llm}, mapping tokens into a useful representational space, adjusting positional information -- that all downstream computation depends on, regardless of task. Indeed, activation steering typically operates on post-MLP residual stream states~\citep{subramani-etal-2022-extracting, adila2026weight} rather than individual attention heads. More recently, for activation-based confidence estimation probes, features are also extracted from post-MLP states~\citep{subramani2026the}.

\paragraph{Superposition and granularity.}
\citet{elhage2022superposition} observe that neural networks represent more features than they have dimensions.
Thus, two tasks that rely on different features in the same attention head or MLP block will inevitably share that component. Neuron-level circuits (a finer-grained unit of computation) are substantially more task-specific, though only a small fraction of individual neurons recur across examples of the same task. One might expect feature-level circuits derived from SAEs or CLTs~\citep{dunefsky2024transcoders,marks2025sparse,ameisen2025circuit} to recover both properties. However, we find preliminary evidence against this for \texttt{Gemma-2-2B}, where SAE feature circuits are as task-specific as neuron circuits and only slightly more consistent (\cref{sec:sae_features}).

\paragraph{Reuse as a feature, not a bug.}
So far, we have treated non-specificity as a limitation of circuit discovery, but this assumes that task circuits ought to be distinct. It could instead be the case that models develop (via training pressures) reusable computational motifs that support many different behaviors.
Such reuse could plausibly contribute to generalization; \eg in-context learning likely works because models can apply the same retrieval and binding operations across novel contexts~\citep{olsson2022context,arditi2026incontextlearningof}. Under this view, the shared core is a meaningful subject of study in and of itself.

\subsection{What does this mean for circuit-level analysis?}

\paragraph{Neither granularity (component or neuron) recovers both desired properties.}
Our results show that component- and neuron-level circuits seem to capture different properties of model computation. Component-level circuits are stable across inputs but overlap substantially across tasks, while neuron-level circuits are task-specific but vary across examples. A component-level circuit may therefore say little about what distinguishes one task from another, while a neuron-level circuit may not characterize the task as a whole. Our preliminary results with sparse feature circuits for \texttt{Gemma-2-2B} suggest that the same holds for feature-level circuits (\cref{sec:sae_features}). Claims about a circuit for a behavior therefore depend heavily on (among other things) the granularity at which it is studied.

\paragraph{Circuit structure depends on the attribution method.}
The main contrast between component- and neuron-level circuits holds under EAP-IG and RelP. However, the methods recover different amounts of component reuse and produce circuits with different compositions and layer distributions. This suggests that circuits obtained by different attribution methods will have qualitatively different structures. 


\paragraph{Targeted interventions require finer notions of localization.}
The broad reuse of causally important components may have implications for model editing. Methods that modify particular weight matrices to target factual associations~\citep{meng2022locating,dai2022knowledge} may affect other behaviors if the same matrices support many tasks. This is consistent with evidence that causal localization does not straightforwardly inform editing~\citep{hase2023does} and that editing techniques can have low specificity~\citep{hoelscher-obermaier-etal-2023-detecting}. Our results apply less directly to safety interventions that modify directions within activation space rather than entire components~\citep{li2024inference,zou2025representationengineeringtopdownapproach, lee2025programming}. 
\section{Related Work}
\label{sec:related_work}

\paragraph{Circuit discovery and evaluation.}
One of the first works to almost fully reverse-engineer a model behavior was \citet{wang2022interpretability}, who identified an attention-only circuit for indirect object identification (IOI) in \texttt{GPT-2 Small}.
Subsequent work has focused on scaling circuit discovery~\citep{conmy2023towards, syed-etal-2024-attribution}.
We use EAP-IG~\citep{hanna2024have} and RelP~\citep{jafari2025relp} throughout for their scalability.
On the evaluation side, \citet{shi2024hypothesis} propose statistical tests for \emph{necessity}, \emph{sufficiency}, and \emph{minimality}, while \citet{miller2024transformer} show that standard evaluation metrics can be fragile.
Our work adds \emph{consistency} and \emph{specificity} to this evaluation toolkit.

\paragraph{Circuit reuse across tasks.}
Several works have studied the concept of circuit reuse across tasks.
\citet{merullo2024circuit} show that the IOI circuit in \texttt{GPT-2 Medium} is largely reused for a Colored Objects task, with 78\% of its attention heads in common, \citet{lan-etal-2024-towards} find that related sequence continuation tasks rely on a shared sub-circuit, and \citet{mondorf-etal-2025-circuit} find that functionally similar string-edit operations have overlapping circuits that can be composed to represent more complex operations.
These works interpret shared components as evidence that the tasks share an algorithm.
However, they do not disambiguate whether a shared component reflects a shared algorithm or is simply needed for every task (\ie specificity).
Our cross-task experiments are designed to disambiguate between these cases, at two levels of granularity.
We find that component-level circuits overlap even between unrelated tasks (\eg \texttt{Addition} and \texttt{ARC}) and are not specific to their task, whereas neuron-level circuits are, which suggests that shared components need not reflect an intuitive shared algorithm.

\section{Conclusion}


We demonstrate that consistency and specificity are important properties to consider for circuit discovery.
Although circuits at the component and neuron level pass necessity tests, neither satisfies both properties.
Component-level circuits are consistent but not specific, since circuits for different tasks overlap heavily and damage other tasks as much as their own.
Neuron-level circuits are the opposite, with few neurons recurring across a task but far more damage to their own task than to any other, and sparse feature circuits in \texttt{Gemma-2-2B} behave similarly.
This complicates the assumption that circuits are stable and unique descriptions of a task, and implies that controlling model behavior via circuits is not a precise intervention.

What should the field take from this? We think circuit discovery can still be useful, but researchers should be much more careful about what claims these techniques can support (since passing necessity tests alone does not imply other desired properties, such as consistency or specificity).
Our findings also prompt several exciting questions for future work.
Can we design attribution methods that explicitly incorporate some notion of specificity?
Can high reuse on one task predict whether a model generalizes to other related tasks?
Should we rethink the definition of a circuit, for example as a "soft" or "fuzzy" subgraph that varies across inputs rather than a single fixed one?
And finally, what other properties of circuits should we care about, if we want to measure how useful circuits are as an abstraction for understanding model behavior?

\subsection*{AI use statement}

We used AI tools, specifically Claude Code, for writing the Matplotlib code used to make the plots in this paper, with human guidance on the specific style and content of said plots. We have verified that this code uses the data we collected from our experiments and creates accurate plots.
Additionally, we used large language models, specifically Claude Fable 5.1, to review our final manuscript (i.e. by prompting it with "You are reviewer 2, please provide a thorough review of the attached paper"), which helped us to improve readability of the paper, but did not specifically lead to any new experiments being run.
We otherwise have not relied on AI tools for any of the experiments or writing in this paper.
We take responsibility for the final content of this work, including text, claims or artifacts produced with the aid of generative AI.

%

\bibliography{custom}

@inproceedings{syed-etal-2024-attribution,
    title = "Attribution Patching Outperforms Automated Circuit Discovery",
    author = "Syed, Aaquib  and
      Rager, Can  and
      Conmy, Arthur",
    editor = "Belinkov, Yonatan  and
      Kim, Najoung  and
      Jumelet, Jaap  and
      Mohebbi, Hosein  and
      Mueller, Aaron  and
      Chen, Hanjie",
    booktitle = "Proceedings of the 7th BlackboxNLP Workshop: Analyzing and Interpreting Neural Networks for NLP",
    month = nov,
    year = "2024",
    address = "Miami, Florida, US",
    publisher = "Association for Computational Linguistics",
    url = "https://aclanthology.org/2024.blackboxnlp-1.25/",
    doi = "10.18653/v1/2024.blackboxnlp-1.25",
    pages = "407--416"
}

@article{elhage2022superposition,
   title={Toy Models of Superposition},
   author={Elhage, Nelson and Hume, Tristan and Olsson, Catherine and Schiefer, Nicholas and Henighan, Tom and Kravec, Shauna and Hatfield-Dodds, Zac and Lasenby, Robert and Drain, Dawn and Chen, Carol and Grosse, Roger and McCandlish, Sam and Kaplan, Jared and Amodei, Dario and Wattenberg, Martin and Olah, Christopher},
   year={2022},
   journal={Transformer Circuits Thread},
   url={https://transformer-circuits.pub/2022/to\_model/index.html}
}

@article{elhage2021mathematical,
   title={A Mathematical Framework for Transformer Circuits},
   author={Elhage, Nelson and Nanda, Neel and Olsson, Catherine and Henighan, Tom and Joseph, Nicholas and Mann, Ben and Askell, Amanda and Bai, Yuntao and Chen, Anna and Conerly, Tom and DasSarma, Nova and Drain, Dawn and Ganguli, Deep and Hatfield-Dodds, Zac and Hernandez, Danny and Jones, Andy and Kernion, Jackson and Lovitt, Liane and Ndousse, Kamal and Amodei, Dario and Brown, Tom and Clark, Jack and Kaplan, Jared and McCandlish, Sam and Olah, Chris},
   year={2021},
   journal={Transformer Circuits Thread},
   url={https://transformer-circuits.pub/2021/framework/index.html}
}

@inproceedings{wang2022interpretability,
    title={Interpretability in the Wild: a Circuit for Indirect Object Identification in {GPT}-2 Small},
    author={Kevin Ro Wang and Alexandre Variengien and Arthur Conmy and Buck Shlegeris and Jacob Steinhardt},
    booktitle={The Eleventh International Conference on Learning Representations },
    year={2023},
    url={https://openreview.net/forum?id=NpsVSN6o4ul}
}

@article{clark2018arc,
  title={Think you have Solved Question Answering? Try {ARC}, the {AI2} Reasoning Challenge},
  author={Peter Clark and Isaac Cowhey and Oren Etzioni and Tushar Khot and Ashish Sabharwal and Carissa Schoenick and Oyvind Tafjord},
  journal={ArXiv},
  year={2018},
  volume={abs/1803.05457},
  url={https://api.semanticscholar.org/CorpusID:3922816}
}

@inproceedings{merullo2024circuit,
    title={Circuit Component Reuse Across Tasks in Transformer Language Models},
    author={Jack Merullo and Carsten Eickhoff and Ellie Pavlick},
    booktitle={The Twelfth International Conference on Learning Representations},
    year={2024},
    url={https://openreview.net/forum?id=fpoAYV6Wsk}
}

@article{qwen3,
    title={Qwen3 Technical Report}, 
    author={An Yang and Anfeng Li and Baosong Yang and Beichen Zhang and Binyuan Hui and Bo Zheng and Bowen Yu and Chang Gao and Chengen Huang and Chenxu Lv and Chujie Zheng and Dayiheng Liu and Fan Zhou and Fei Huang and Feng Hu and Hao Ge and Haoran Wei and Huan Lin and Jialong Tang and Jian Yang and Jianhong Tu and Jianwei Zhang and Jianxin Yang and Jiaxi Yang and Jing Zhou and Jingren Zhou and Junyang Lin and Kai Dang and Keqin Bao and Kexin Yang and Le Yu and Lianghao Deng and Mei Li and Mingfeng Xue and Mingze Li and Pei Zhang and Peng Wang and Qin Zhu and Rui Men and Ruize Gao and Shixuan Liu and Shuang Luo and Tianhao Li and Tianyi Tang and Wenbiao Yin and Xingzhang Ren and Xinyu Wang and Xinyu Zhang and Xuancheng Ren and Yang Fan and Yang Su and Yichang Zhang and Yinger Zhang and Yu Wan and Yuqiong Liu and Zekun Wang and Zeyu Cui and Zhenru Zhang and Zhipeng Zhou and Zihan Qiu},
    journal = {arXiv preprint arXiv:2505.09388},
    year={2025}
}

@article{gemmateam2024gemma2improvingopen,
  title={Gemma 2: Improving Open Language Models at a Practical Size},
  author={{Gemma Team}},
  journal={ArXiv},
  year={2024},
  volume={abs/2408.00118},
  url={https://api.semanticscholar.org/CorpusID:270843326}
}

@article{grattafiori2024llama3herdmodels,
  title={The Llama 3 Herd of Models},
  author={{Llama Team}},
  journal={ArXiv},
  year={2024},
  url={https://api.semanticscholar.org/CorpusID:271571434}
}

@inproceedings{
walsh2025,
title={2 {OLM}o 2 Furious ({COLM}{\textquoteright}s Version)},
author={Evan Pete Walsh and Luca Soldaini and Dirk Groeneveld and Kyle Lo and Shane Arora and Akshita Bhagia and Yuling Gu and Shengyi Huang and Matt Jordan and Nathan Lambert and Dustin Schwenk and Oyvind Tafjord and Taira Anderson and David Atkinson and Faeze Brahman and Christopher Clark and Pradeep Dasigi and Nouha Dziri and Allyson Ettinger and Michal Guerquin and David Heineman and Hamish Ivison and Pang Wei Koh and Jiacheng Liu and Saumya Malik and William Merrill and Lester James Validad Miranda and Jacob Morrison and Tyler Murray and Crystal Nam and Jake Poznanski and Valentina Pyatkin and Aman Rangapur and Michael Schmitz and Sam Skjonsberg and David Wadden and Christopher Wilhelm and Michael Wilson and Luke Zettlemoyer and Ali Farhadi and Noah A. Smith and Hannaneh Hajishirzi},
booktitle={Second Conference on Language Modeling},
year={2025},
url={https://openreview.net/forum?id=2ezugTT9kU}
}

@inproceedings{
hanna2024have,
title={Have Faith in Faithfulness: Going Beyond Circuit Overlap When Finding Model Mechanisms},
author={Michael Hanna and Sandro Pezzelle and Yonatan Belinkov},
booktitle={First Conference on Language Modeling},
year={2024},
url={https://openreview.net/forum?id=TZ0CCGDcuT}
}

@inproceedings{miller2024transformer,
    title={Transformer Circuit Evaluation Metrics Are Not Robust},
    author={Joseph Miller and Bilal Chughtai and William Saunders},
    booktitle={First Conference on Language Modeling},
    year={2024},
    url={https://openreview.net/forum?id=zSf8PJyQb2}
}

@inproceedings{
shi2024hypothesis,
title={Hypothesis Testing the Circuit Hypothesis in {LLM}s},
author={Claudia Shi and Nicolas Beltran-Velez and Achille Nazaret and Carolina Zheng and Adri{\`a} Garriga-Alonso and Andrew Jesson and Maggie Makar and David Blei},
booktitle={The Thirty-eighth Annual Conference on Neural Information Processing Systems},
year={2024},
url={https://openreview.net/forum?id=5ai2YFAXV7}
}

@inproceedings{
jafari2025relp,
title={Rel{P}: Faithful and Efficient Circuit Discovery via Relevance Patching},
author={Farnoush Rezaei Jafari and Oliver Eberle and Ashkan Khakzar and Neel Nanda},
booktitle={Mechanistic Interpretability Workshop at NeurIPS 2025},
year={2025},
url={https://openreview.net/forum?id=5PKPy82sWN}
}

@inproceedings{
marks2025sparse,
title={Sparse Feature Circuits: Discovering and Editing Interpretable Causal Graphs in Language Models},
author={Samuel Marks and Can Rager and Eric J Michaud and Yonatan Belinkov and David Bau and Aaron Mueller},
booktitle={The Thirteenth International Conference on Learning Representations},
year={2025},
url={https://openreview.net/forum?id=I4e82CIDxv}
}

@inproceedings{
mueller2025mib,
title={{MIB}: A Mechanistic Interpretability Benchmark},
author={Aaron Mueller and Atticus Geiger and Sarah Wiegreffe and Dana Arad and Iv{\'a}n Arcuschin and Adam Belfki and Yik Siu Chan and Jaden Fried Fiotto-Kaufman and Tal Haklay and Michael Hanna and Jing Huang and Rohan Gupta and Yaniv Nikankin and Hadas Orgad and Nikhil Prakash and Anja Reusch and Aruna Sankaranarayanan and Shun Shao and Alessandro Stolfo and Martin Tutek and Amir Zur and David Bau and Yonatan Belinkov},
booktitle={Forty-second International Conference on Machine Learning},
year={2025},
url={https://openreview.net/forum?id=sSrOwve6vb}
}

@inproceedings{conmy2023towards,
 author = {Conmy, Arthur and Mavor-Parker, Augustine and Lynch, Aengus and Heimersheim, Stefan and Garriga-Alonso, Adri\`{a}},
 booktitle = {Advances in Neural Information Processing Systems},
 editor = {A. Oh and T. Naumann and A. Globerson and K. Saenko and M. Hardt and S. Levine},
 pages = {16318--16352},
 publisher = {Curran Associates, Inc.},
 title = {Towards Automated Circuit Discovery for Mechanistic Interpretability},
 url = {https://proceedings.neurips.cc/paper_files/paper/2023/file/34e1dbe95d34d7ebaf99b9bcaeb5b2be-Paper-Conference.pdf},
 volume = {36},
 year = {2023}
}

@inproceedings{meng2022locating,
    title={Locating and Editing Factual Associations in {GPT}},
    author={Kevin Meng and David Bau and Alex Andonian and Yonatan Belinkov},
    booktitle={Advances in Neural Information Processing Systems},
    year={2022},
    url={https://openreview.net/forum?id=-h6WAS6eE4}
}

@inproceedings{li2024inference,
    title={Inference-Time Intervention: Eliciting Truthful Answers from a Language Model},
    author={Kenneth Li and Oam Patel and Fernanda Vi\'{e}gas and Hanspeter Pfister and Martin Wattenberg},
    booktitle={Advances in Neural Information Processing Systems},
    year={2023}
}

@inproceedings{dai2022knowledge,
    title={Knowledge Neurons in Pretrained Transformers},
    author={Damai Dai and Li Dong and Yaru Hao and Zhifang Sui and Baobao Chang and Furu Wei},
    booktitle={Proceedings of the 60th Annual Meeting of the Association for Computational Linguistics},
    year={2022}
}

@inproceedings{
arora2026language,
title={Language Model Circuits Are Sparse in the Neuron Basis},
author={Aryaman Arora and Zhengxuan Wu and Jacob Steinhardt and Sarah Schwettmann},
booktitle={Forty-third International Conference on Machine Learning},
year={2026},
url={https://openreview.net/forum?id=OrviwFWcN4}
}

@article{ameisen2025circuit,
  author={Ameisen, Emmanuel and Lindsey, Jack and Pearce, Adam and Gurnee, Wes and Turner, Nicholas L. and Chen, Brian and Citro, Craig and Abrahams, David and Carter, Shan and Hosmer, Basil and Marcus, Jonathan and Sklar, Michael and Templeton, Adly and Bricken, Trenton and McDougall, Callum and Cunningham, Hoagy and Henighan, Thomas and Jermyn, Adam and Jones, Andy and Persic, Andrew and Qi, Zhenyi and Ben Thompson, T. and Zimmerman, Sam and Rivoire, Kelley and Conerly, Thomas and Olah, Chris and Batson, Joshua},
  title={Circuit Tracing: Revealing Computational Graphs in Language Models},
  journal={Transformer Circuits Thread},
  year={2025},
  url={https://transformer-circuits.pub/2025/attribution-graphs/methods.html}
}

@article{bricken2023monosemanticity,
       title={Towards Monosemanticity: Decomposing Language Models With Dictionary Learning},
       author={Bricken, Trenton and Templeton, Adly and Batson, Joshua and Chen, Brian and Jermyn, Adam and Conerly, Tom and Turner, Nick and Anil, Cem and Denison, Carson and Askell, Amanda and Lasenby, Robert and Wu, Yifan and Kravec, Shauna and Schiefer, Nicholas and Maxwell, Tim and Joseph, Nicholas and Hatfield-Dodds, Zac and Tamkin, Alex and Nguyen, Karina and McLean, Brayden and Burke, Josiah E and Hume, Tristan and Carter, Shan and Henighan, Tom and Olah, Christopher},
       year={2023},
       journal={Transformer Circuits Thread},
       note={https://transformer-circuits.pub/2023/monosemantic-features/index.html}
    }

@inproceedings{vig2020causal,
    author = {Vig, Jesse and Gehrmann, Sebastian and Belinkov, Yonatan and Qian, Sharon and Nevo, Daniel and Singer, Yaron and Shieber, Stuart},
    booktitle = {Advances in Neural Information Processing Systems},
    editor = {H. Larochelle and M. Ranzato and R. Hadsell and M.F. Balcan and H. Lin},
    pages = {12388--12401},
    publisher = {Curran Associates, Inc.},
    title = {Investigating Gender Bias in Language Models Using Causal Mediation Analysis},
    url = {https://proceedings.neurips.cc/paper_files/paper/2020/file/92650b2e92217715fe312e6fa7b90d82-Paper.pdf},
    volume = {33},
    year = {2020}
}

@inproceedings{
hase2023does,
title={Does Localization Inform Editing? Surprising Differences in Causality-Based Localization vs. Knowledge Editing in Language Models},
author={Peter Hase and Mohit Bansal and Been Kim and Asma Ghandeharioun},
booktitle={Thirty-seventh Conference on Neural Information Processing Systems},
year={2023},
url={https://openreview.net/forum?id=EldbUlZtbd}
}

@inproceedings{hoelscher-obermaier-etal-2023-detecting,
    title = "Detecting Edit Failures In Large Language Models: An Improved Specificity Benchmark",
    author = "Hoelscher-Obermaier, Jason  and
      Persson, Julia  and
      Kran, Esben  and
      Konstas, Ioannis  and
      Barez, Fazl",
    editor = "Rogers, Anna  and
      Boyd-Graber, Jordan  and
      Okazaki, Naoaki",
    booktitle = "Findings of the Association for Computational Linguistics: ACL 2023",
    month = jul,
    year = "2023",
    address = "Toronto, Canada",
    publisher = "Association for Computational Linguistics",
    url = "https://aclanthology.org/2023.findings-acl.733/",
    doi = "10.18653/v1/2023.findings-acl.733",
    pages = "11548--11559",
}

@inproceedings{subramani-etal-2022-extracting,
    title = "Extracting Latent Steering Vectors from Pretrained Language Models",
    author = "Subramani, Nishant  and
      Suresh, Nivedita  and
      Peters, Matthew",
    editor = "Muresan, Smaranda  and
      Nakov, Preslav  and
      Villavicencio, Aline",
    booktitle = "Findings of the Association for Computational Linguistics: ACL 2022",
    month = may,
    year = "2022",
    address = "Dublin, Ireland",
    publisher = "Association for Computational Linguistics",
    url = "https://aclanthology.org/2022.findings-acl.48/",
    doi = "10.18653/v1/2022.findings-acl.48",
    pages = "566--581"
}

@inproceedings{
    sun2025redeep,
    title={Re{D}e{EP}: Detecting Hallucination in Retrieval-Augmented Generation via Mechanistic Interpretability},
    author={ZhongXiang Sun and Xiaoxue Zang and Kai Zheng and Jun Xu and Xiao Zhang and Weijie Yu and Yang Song and Han Li},
    booktitle={The Thirteenth International Conference on Learning Representations},
    year={2025},
    url={https://openreview.net/forum?id=ztzZDzgfrh}
}

@article{liu2025llm,
  title={LLM Microscope: What Model Internals Reveal About Answer Correctness and Context Utilization},
  author={Liu, Jiarui and Jain, Jivitesh and Diab, Mona and Subramani, Nishant},
  journal={arXiv preprint arXiv:2510.04013},
  year={2025}
}

@article{pearl1995docalculus,
 ISSN = {00063444, 14643510},
 URL = {http://www.jstor.org/stable/2337329},
 author = {Judea Pearl},
 journal = {Biometrika},
 number = {4},
 pages = {669--688},
 publisher = {[Oxford University Press, Biometrika Trust]},
 title = {Causal Diagrams for Empirical Research},
 urldate = {2026-05-08},
 volume = {82},
 year = {1995}
}

@inproceedings{
dunefsky2024transcoders,
title={Transcoders find interpretable {LLM} feature circuits},
author={Jacob Dunefsky and Philippe Chlenski and Neel Nanda},
booktitle={The Thirty-eighth Annual Conference on Neural Information Processing Systems},
year={2024},
url={https://openreview.net/forum?id=J6zHcScAo0}
}

@inproceedings{sundararajan2017axiomatic,
author = {Sundararajan, Mukund and Taly, Ankur and Yan, Qiqi},
title = {Axiomatic attribution for deep networks},
year = {2017},
publisher = {JMLR.org},
booktitle = {Proceedings of the 34th International Conference on Machine Learning - Volume 70},
pages = {3319–3328},
numpages = {10},
location = {Sydney, NSW, Australia},
series = {ICML'17}
}

@article{bach2015pixel,
    doi = {10.1371/journal.pone.0130140},
    author = {Bach, Sebastian AND Binder, Alexander AND Montavon, Grégoire AND Klauschen, Frederick AND Müller, Klaus-Robert AND Samek, Wojciech},
    journal = {PLOS ONE},
    publisher = {Public Library of Science},
    title = {On Pixel-Wise Explanations for Non-Linear Classifier Decisions by Layer-Wise Relevance Propagation},
    year = {2015},
    month = {07},
    volume = {10},
    url = {https://doi.org/10.1371/journal.pone.0130140},
    pages = {1-46},
    number = {7},
}

@inproceedings{ali2022xai,
  author    = {Ameen Ali and
               Thomas Schnake and
               Oliver Eberle and
               Gr{\'{e}}goire Montavon and
               Klaus{-}Robert M{\"{u}}ller and
               Lior Wolf},
  title     = {{XAI} for Transformers: Better Explanations through Conservative Propagation},
  booktitle = {{ICML}},
  series    = {Proceedings of Machine Learning Research},
  volume    = {162},
  pages     = {435--451},
  publisher = {{PMLR}},
  year      = {2022}
}

@inproceedings{xiao2024efficient,
 author = {Xiao, Guangxuan and Tian, Yuandong and Chen, Beidi and Han, Song and Lewis, Mike },
 booktitle = {International Conference on Learning Representations},
 editor = {B. Kim and Y. Yue and S. Chaudhuri and K. Fragkiadaki and M. Khan and Y. Sun},
 pages = {21875--21895},
 title = {Efficient Streaming Language Models with Attention Sinks},
 url = {https://proceedings.iclr.cc/paper_files/paper/2024/file/5e5fd18f863cbe6d8ae392a93fd271c9-Paper-Conference.pdf},
 volume = {2024},
 year = {2024}
}

@inproceedings{subramani2026the,
    title={The {ACUTE} Protocol: Operationalizing Language Model Activations for Better Calibration, Utility, and Trust},
    author={Nishant Subramani and Palash Goyal and Yiwen Song and Mani Malek and Yuan Xue and Tomas Pfister and Hamid Palangi},
    booktitle={Forty-third International Conference on Machine Learning},
    year={2026},
    url={https://openreview.net/forum?id=67lpPXbYf6}
}

@misc{zou2025representationengineeringtopdownapproach,
      title={Representation Engineering: A Top-Down Approach to AI Transparency}, 
      author={Andy Zou and Long Phan and Sarah Chen and James Campbell and Phillip Guo and Richard Ren and Alexander Pan and Xuwang Yin and Mantas Mazeika and Ann-Kathrin Dombrowski and Shashwat Goel and Nathaniel Li and Michael J. Byun and Zifan Wang and Alex Mallen and Steven Basart and Sanmi Koyejo and Dawn Song and Matt Fredrikson and Zico Kolter and Dan Hendrycks},
      year={2023},
      eprint={2310.01405},
      archivePrefix={arXiv},
      primaryClass={cs.CL}
}

@inproceedings{
lee2025programming,
title={Programming Refusal with Conditional Activation Steering},
author={Bruce W. Lee and Inkit Padhi and Karthikeyan Natesan Ramamurthy and Erik Miehling and Pierre Dognin and Manish Nagireddy and Amit Dhurandhar},
booktitle={The Thirteenth International Conference on Learning Representations},
year={2025},
url={https://openreview.net/forum?id=Oi47wc10sm}
}

@article{olsson2022context,
   title={In-context Learning and Induction Heads},
   author={Olsson, Catherine and Elhage, Nelson and Nanda, Neel and Joseph, Nicholas and DasSarma, Nova and Henighan, Tom and Mann, Ben and Askell, Amanda and Bai, Yuntao and Chen, Anna and Conerly, Tom and Drain, Dawn and Ganguli, Deep and Hatfield-Dodds, Zac and Hernandez, Danny and Johnston, Scott and Jones, Andy and Kernion, Jackson and Lovitt, Liane and Ndousse, Kamal and Amodei, Dario and Brown, Tom and Clark, Jack and Kaplan, Jared and McCandlish, Sam and Olah, Chris},
   year={2022},
   journal={Transformer Circuits Thread},
   note={https://transformer-circuits.pub/2022/in-context-learning-and-induction-heads/index.html}
}

@inproceedings{arditi2026incontextlearningof,
  author = {Arditi, Andy},
  title = {In-context learning of representations can be explained by induction circuits},
  booktitle = {ICLR Blogposts 2026},
  year = {2026},
  date = {April 27, 2026},
  url  = {https://iclr-blogposts.github.io/2026/blog/2026/iclr-induction/}
}

@inproceedings{NEURIPS2025_764ff747,
 author = {Chanin, David and Wilken-Smith, James and Dulka, Tom\'{a}\v{s} and Bhatnagar, Hardik and Golechha, Satvik and Bloom, Joseph},
 booktitle = {Advances in Neural Information Processing Systems},
 doi = {10.52202/085713-2755},
 editor = {D. Belgrave and C. Zhang and H. Lin and R. Pascanu and P. Koniusz and M. Ghassemi and N. Chen},
 pages = {82318--82355},
 publisher = {Curran Associates, Inc.},
 title = {A is for Absorption: Studying Feature Splitting and Absorption in Sparse Autoencoders},
 url = {https://proceedings.neurips.cc/paper_files/paper/2025/file/764ff7477b8e24dbe01531f6791e8bdf-Paper-Conference.pdf},
 volume = {38, Main Conference},
 year = {2025}
}

@inproceedings{song-etal-2026-mechanistic,
    title = "Mechanistic Interpretability Should Prioritize Feature Consistency in Sparse Autoencoders",
    author = "Song, Xiangchen  and
      Muhamed, Aashiq  and
      Zheng, Yujia  and
      Kong, Lingjing  and
      Tang, Zeyu  and
      Diab, Mona T.  and
      Smith, Virginia  and
      Zhang, Kun",
    editor = "Liakata, Maria  and
      Moreira, Viviane P.  and
      Zhang, Jiajun  and
      Jurgens, David",
    booktitle = "Proceedings of the 64th Annual Meeting of the {A}ssociation for {C}omputational {L}inguistics (Volume 1: Long Papers)",
    month = jul,
    year = "2026",
    address = "San Diego, California, United States",
    publisher = "Association for Computational Linguistics",
    url = "https://aclanthology.org/2026.acl-long.99/",
    doi = "10.18653/v1/2026.acl-long.99",
    pages = "2172--2210",
    ISBN = "979-8-89176-390-6"
}

@inproceedings{
adila2026weight,
title={Weight Updates as Activation Shifts: A Principled Framework for Steering},
author={Dyah Adila and John Cooper and Alexander Yun and Avi Trost and Frederic Sala},
booktitle={Forty-third International Conference on Machine Learning},
year={2026},
url={https://openreview.net/forum?id=eMCp3enKXM}
}

@inproceedings{
    zhang2024towards,
    title={Towards Best Practices of Activation Patching in Language Models: Metrics and Methods},
    author={Fred Zhang and Neel Nanda},
    booktitle={The Twelfth International Conference on Learning Representations},
    year={2024},
    url={https://openreview.net/forum?id=Hf17y6u9BC}
}

@inproceedings{
    li2024optimal,
    title={Optimal ablation for interpretability},
    author={Maximilian Li and Lucas Janson},
    booktitle={The Thirty-eighth Annual Conference on Neural Information Processing Systems},
    year={2024},
    url={https://openreview.net/forum?id=opt72TYzwZ}
}

@inproceedings{
huben2024sparse,
title={Sparse Autoencoders Find Highly Interpretable Features in Language Models},
author={Robert Huben and Hoagy Cunningham and Logan Riggs Smith and Aidan Ewart and Lee Sharkey},
booktitle={The Twelfth International Conference on Learning Representations},
year={2024},
url={https://openreview.net/forum?id=F76bwRSLeK}
}

@misc{rajamanoharan2025jumping,
      title={Jumping Ahead: Improving Reconstruction Fidelity with JumpReLU Sparse Autoencoders}, 
      author={Senthooran Rajamanoharan and Tom Lieberum and Nicolas Sonnerat and Arthur Conmy and Vikrant Varma and János Kramár and Neel Nanda},
      year={2024},
      eprint={2407.14435},
      archivePrefix={arXiv},
      primaryClass={cs.LG},
      url={https://arxiv.org/abs/2407.14435}, 
}

@inproceedings{lieberum-etal-2024-gemma,
    title = "Gemma Scope: Open Sparse Autoencoders Everywhere All At Once on Gemma 2",
    author = "Lieberum, Tom  and
      Rajamanoharan, Senthooran  and
      Conmy, Arthur  and
      Smith, Lewis  and
      Sonnerat, Nicolas  and
      Varma, Vikrant  and
      Kramar, Janos  and
      Dragan, Anca  and
      Shah, Rohin  and
      Nanda, Neel",
    editor = "Belinkov, Yonatan  and
      Kim, Najoung  and
      Jumelet, Jaap  and
      Mohebbi, Hosein  and
      Mueller, Aaron  and
      Chen, Hanjie",
    booktitle = "Proceedings of the 7th BlackboxNLP Workshop: Analyzing and Interpreting Neural Networks for NLP",
    month = nov,
    year = "2024",
    address = "Miami, Florida, US",
    publisher = "Association for Computational Linguistics",
    url = "https://aclanthology.org/2024.blackboxnlp-1.19/",
    doi = "10.18653/v1/2024.blackboxnlp-1.19",
    pages = "278--300"
}

@inproceedings{
    gao2025scaling,
    title={Scaling and evaluating sparse autoencoders},
    author={Leo Gao and Tom Dupre la Tour and Henk Tillman and Gabriel Goh and Rajan Troll and Alec Radford and Ilya Sutskever and Jan Leike and Jeffrey Wu},
    booktitle={The Thirteenth International Conference on Learning Representations},
    year={2025},
    url={https://openreview.net/forum?id=tcsZt9ZNKD}
}

@inproceedings{mondorf-etal-2025-circuit,
    title = "Circuit Compositions: Exploring Modular Structures in Transformer-Based Language Models",
    author = "Mondorf, Philipp  and
      Wold, Sondre  and
      Plank, Barbara",
    editor = "Che, Wanxiang  and
      Nabende, Joyce  and
      Shutova, Ekaterina  and
      Pilehvar, Mohammad Taher",
    booktitle = "Proceedings of the 63rd Annual Meeting of the Association for Computational Linguistics (Volume 1: Long Papers)",
    month = jul,
    year = "2025",
    address = "Vienna, Austria",
    publisher = "Association for Computational Linguistics",
    url = "https://aclanthology.org/2025.acl-long.727/",
    doi = "10.18653/v1/2025.acl-long.727",
    pages = "14934--14955",
    ISBN = "979-8-89176-251-0"
}

@inproceedings{lan-etal-2024-towards,
    title = "Towards Interpretable Sequence Continuation: Analyzing Shared Circuits in Large Language Models",
    author = "Lan, Michael  and
      Torr, Philip  and
      Barez, Fazl",
    editor = "Al-Onaizan, Yaser  and
      Bansal, Mohit  and
      Chen, Yun-Nung",
    booktitle = "Proceedings of the 2024 Conference on Empirical Methods in Natural Language Processing",
    month = nov,
    year = "2024",
    address = "Miami, Florida, USA",
    publisher = "Association for Computational Linguistics",
    url = "https://aclanthology.org/2024.emnlp-main.699/",
    doi = "10.18653/v1/2024.emnlp-main.699",
    pages = "12576--12601"
}

@misc{heimersheim2024useinterpretactivationpatching,
      title={How to use and interpret activation patching}, 
      author={Stefan Heimersheim and Neel Nanda},
      year={2024},
      eprint={2404.15255},
      archivePrefix={arXiv},
      primaryClass={cs.LG},
      url={https://arxiv.org/abs/2404.15255}, 
}
\bibliographystyle{colm2026_conference}

\appendix
\clearpage
\appendix

\etocdepthtag.toc{appendix}
\etocsettagdepth{mainbody}{none}
\etocsettagdepth{appendix}{subsection}
\renewcommand{\contentsname}{Appendix Contents}
\tableofcontents
\clearpage

\section{Limitations}
\label{sec:limitations}

\textbf{Granularity.} We analyze circuits at two levels of granularity throughout the main text: attention heads and MLP blocks, and individual MLP neurons. We also run our experiments over sparse autoencoder features for \texttt{Gemma-2-2B}, which has a public SAE release, and find the same tradeoff (\cref{sec:sae_features}). We do not run additional SAE experiments with the other four models, nor with cross-layer transcoders (CLTs;~\citealp{ameisen2025circuit}).

Our primary reason is the computational cost associated with training such SAEs or CLTs (one per layer for each model we study, 108 in total for the remaining four). Furthermore, these feature bases are not necessarily "natural" to the model in the way the neuron basis is, and feature quality is not guaranteed. Indeed, prior work has identified issues such as feature splitting/absorption and (lack of) feature consistency, which may confound findings obtained from feature circuits~\citep{NEURIPS2025_764ff747, song-etal-2026-mechanistic}. 

Finally, recent work suggests that, for the purpose of circuit discovery, MLP neuron circuits are just as sparse as feature circuits, implying that they provide no added benefit~\citep{arora2026language}. Our result using \texttt{Gemma-2-2B} is consistent with theirs: feature circuits are more consistent than neuron circuits on five of six tasks but far less so than component circuits, and they are as specific. Whether this holds across models remains an open question.

\textbf{Task complexity.} Our tasks, while diverse, do not include generative or multi-step tasks. More complex tasks may show different trends for consistency and specificity.

\textbf{Model scale.} We only study models up to 4B parameters. The tradeoff we observe may persist or disappear at larger scale, where a model may have enough capacity to represent each task with a circuit that is both stable across its examples and distinct from other tasks.

\section{Attribution Method Details}
\label{sec:attribution_details}

\subsection{Edge Attribution Patching}
\label{sec:eap_details}

Given a clean input $x$ and a corrupted input $x'$, EAP computes the attribution score for each component $c$ as a first-order  approximation of the effect of patching that component's activation:
\[
\hat{e}_{c} = \left( a_c(x') - a_c(x) \right)^\top \cdot \frac{\partial L(x)}{\partial a_c}
\]
where $a_c(\cdot)$ denotes the activation of component $c$ and $L(x)$ is a scalar metric evaluated on the clean input, here the log-probability of the correct answer (\cref{sec:task_details}), and the score is summed over sequence positions. In practice, this is just a single line of PyTorch code:

\begin{center}
\texttt{score = (act\_corrupt - act\_clean) * grad}
\end{center}

\noindent where \texttt{act\_corrupt} and \texttt{act\_clean} are the component's activations under the corrupted and clean inputs respectively, and \texttt{grad} is the gradient of the metric with respect to the clean activation. Per-component scores are obtained by summing over positions and (for attention heads) the head dimension.

\subsection{EAP with Integrated Gradients}
\label{sec:eapig_details}

The first-order approximation above is unreliable wherever $\partial L / \partial a_c$ is close to zero at the clean activation, since then the score is small regardless of how much the activation itself matters~\citep{hanna2024have}. EAP-IG replaces the single gradient with an integrated gradient~\citep{sundararajan2017axiomatic} accumulated along the straight-line path between the two inputs ($x$ and $x'$). Defining $z$ and $z'$ as the token embeddings of $x$ and $x'$, the EAP-IG score of component $u$ is
$$
\hat{e}^{\,\mathrm{IG}}_{c}
= \left( a_c(x') - a_c(x) \right)^\top \cdot
\frac{1}{m} \sum_{k=0}^{m-1}
\frac{\partial L\!\left( z' + \tfrac{k}{m}\left( z - z' \right) \right)}{\partial a_c},
$$
where $m$ is the number of integration steps. We use $m = 5$, the value \citet{hanna2024have} recommend.

\subsection{Relevance Patching}
\label{sec:relp_details}

RelP modifies attribution patching to better approximate the effects of
actually patching activations~\citep{jafari2025relp}. It scores components with the same product of activation difference and gradient as attribution patching, but computes the gradient through a \emph{replacement model}, whose forward pass is the same but has a modified backwards pass, in the manner of layer-wise relevance propagation, ~\citep{bach2015pixel,ali2022xai,jafari2025relp}. The score of component $c$ is
$$
\hat{e}^{\,\mathrm{RelP}}_{c}
= \left( a_c(x) - a_c(x') \right)^\top \cdot \frac{\partial L_{\mathrm{rep}}(x')}{\partial a_c},
$$
where $L_{\mathrm{rep}}$ denotes the metric computed through the
replacement model. Gradients are computed
using the replacement rules in \Cref{tab:lrp_rules}. We evaluate the
gradient on the corrupted input, corresponding to patching clean
activations into the corrupted run. Thus, in the notation of
\citet{jafari2025relp}, $x'$ corresponds to the original input and $x$
to the patch input.

We follow the replacement rules described by
\citet{arora2026language}, using the patching direction specified above.
The normalization scale and attention pattern are treated as constants
during backpropagation. For a differentiable elementwise activation
$\sigma$ with $\sigma(0)=0$, we use
$$
\widetilde{\sigma}(u)
= u \odot \mathrm{Freeze}\!\left(r_\sigma(u)\right),
\qquad
r_\sigma(t)
=
\begin{cases}
\sigma(t)/t, & t \ne 0,\\
\sigma'(0), & t = 0,
\end{cases}
$$
where $r_\sigma$ is applied elementwise and $\mathrm{Freeze}$ treats
its argument as a constant during backpropagation.
For SiLU, this simplifies to
$u \odot \mathrm{Freeze}(\operatorname{sigmoid}(u))$.
For gated MLPs, we halve the gradients through the product using
the LRP half rule. All other operations are unchanged.

\begin{table}[h]
  \centering
  \begin{tabular}{lll}
    \toprule
    Operation & Original expression & Replacement expression \\
    \midrule
    RMSNorm
    & $\gamma_i x_i \big/ \sqrt{\epsilon + \overline{x^2}}$
    & $\gamma_i x_i \big/
       \mathrm{Freeze}\!\left(\sqrt{\epsilon + \overline{x^2}}\right)$ \\
    Attention
    & $\sum_k A_{qk} v_k$
    & $\sum_k \mathrm{Freeze}(A_{qk})\, v_k$ \\
    Activation function
    & $\sigma(u)$
    & $\widetilde{\sigma}(u)$ \\
    Gated MLP product
    & $\sigma(u) \odot v$
    & $\tfrac{1}{2}\,\widetilde{\sigma}(u) \odot v
       + \tfrac{1}{2}\,\mathrm{Freeze}\!\left(
         \widetilde{\sigma}(u) \odot v \right)$ \\
    \bottomrule
  \end{tabular}
  \vskip 1em
  \caption{\textbf{Replacement rules for RelP.}
  Each replacement preserves the original forward value while
  modifying the gradient computation.
  $\mathrm{Freeze}$ denotes the stop-gradient operator, and
  $\gamma_i$ is the learned RMSNorm gain.
  In the gated-product row, $u$ and $v$ are the gate and
  up-projections of the MLP input, and $\widetilde{\sigma}$
  includes the activation replacement defined above.}
  \label{tab:lrp_rules}
\end{table}

\section{Task Details}
\label{sec:task_details}

Activation patching and its approximations require a clean input $x$ and a corrupted input $x'$ that alters the information the model must use while preserving surface structure as much as possible. The attribution metric we use is the log-probability the model assigns to the correct answer, summed over the answer's token positions. Below we describe each task and its corruption strategy.

\paragraph{Addition.} The model is given a 2-digit arithmetic problem (\eg "Compute: 47 + 63 =") and must output the correct sum. Corrupted inputs are generated by pairing each problem with a randomly selected different problem from the same batch, so that the corrupted input has the same format but different operands and answer.

\paragraph{Boolean Logic.} The model is tasked with evaluating logical expressions composed of \texttt{and}, \texttt{or}, and \texttt{not} over boolean values (\eg "Evaluate: true and (false or true) =" $\rightarrow$ "true"). To construct a corrupted input, we randomly select a literal whose modification ((\eg \texttt{or} $\rightarrow$ \texttt{and}) would flip the expression's truth value. This ensures the corrupted input has the opposite answer while minimally changing the input.

\paragraph{Indirect Object Identification (IOI).} Following the task setup in \citet{wang2022interpretability}, the model must identify the indirect object in sentences with a specific template involving two names (\eg "Bilbo and Frodo spoke in Rivendell before Bilbo gave Sting to" $\rightarrow$ "Frodo"). We use the dataset and counterfactuals from the Mechanistic Interpretability Benchmark~\citep{mueller2025mib}. The corrupted input applies the S2-IO flip counterfactual, which swaps the subject and indirect object names so that the correct completion changes while the sentence template remains the same.

\paragraph{CopyColors MCQA.} This is a multiple-choice task from \citet{mueller2025mib} where the model is provided a passage describing objects and their colors, followed by a question asking which color corresponds to a particular object. The answer choices are given as labeled options (A, B, C, D). The corrupted input is the answer-position counterfactual from the benchmark, which permutes the order of the answer choices so that the correct answer appears at a different position, changing the correct label token while keeping the passage and question unchanged.

\paragraph{ARC Easy / ARC Challenge.} The AI2 Reasoning Challenge~\citep{clark2018arc} consists of multiple-choice science exam questions. The Easy split contains questions answerable with basic retrieval and reasoning, while the Challenge split filters for questions requiring more complex inference. We use the datasets and counterfactuals from \citet{mueller2025mib}. As with CopyColors~MCQA, the corrupted input applies the answer-position counterfactual, permuting the order of answer choices so that the correct label changes.

\section{Model Details}
\label{sec:model_details}

See \cref{tab:model_ids} for a list of model information and \cref{tab:baseline_accuracy} for each model's unablated accuracy on each task.

\begin{table*}[htbp]
\centering
\setlength{\tabcolsep}{3pt}
\begin{tabular}{llrccc}
\toprule
\textbf{Model} & \textbf{HuggingFace ID} & \textbf{Params} & $|L|$ & $|a|$ & $d_{\text{model}}$ \\
\midrule
\texttt{Gemma 2 2B} & \texttt{google/gemma-2-2b} & 2.6B & 26 & 8 & 2304 \\
\texttt{Gemma 2 2B Instruct} & \texttt{google/gemma-2-2b-it} & 2.6B & 26 & 8 & 2304 \\
\midrule
\texttt{Llama-3.2-3B} & \texttt{meta-llama/Llama-3.2-3B} & 3.2B & 28 & 24 & 3072 \\
\texttt{Llama-3.2-3B Instruct} & \texttt{meta-llama/Llama-3.2-3B-Instruct} & 3.2B & 28 & 24 & 3072 \\
\midrule
\texttt{Qwen3-4B} & \texttt{Qwen/Qwen3-4B} & 4.0B & 36 & 32 & 2560 \\
\midrule
\texttt{OLMo-2-1B} & \texttt{allenai/OLMo-2-0425-1B} & 1.2B & 16 & 16 & 2048 \\
\bottomrule
\end{tabular}%
\caption{\textbf{Models studied in this work.} We report the number of parameters, number of layers $|L|$, total number of attention heads $|a|$, and hidden dimension $d_{\text{model}}$ for each model.}
\label{tab:model_ids}
\end{table*}

\begin{table}[htbp]
\centering
\setlength{\tabcolsep}{5pt}
\resizebox{\textwidth}{!}{%
\begin{tabular}{lcccccc}
\toprule
\textbf{Model} & Addition & ARC (Chal.) & ARC (Easy) & Boolean & IOI & CopyColors MCQA \\
\midrule
\texttt{Gemma 2 2B} & 80 & 57 & 82 & 64 & 100 & 100 \\
\texttt{Gemma 2 2B Instruct} & 96 & 77 & 92 & 57 & 100 & 98 \\
\texttt{Llama-3.2-3B} & 100 & 69 & 85 & 72 & 100 & 100 \\
\texttt{Llama-3.2-3B Instruct} & 99 & 76 & 88 & 66 & 100 & 98 \\
\texttt{Qwen3-4B} & 98 & 84 & 94 & 88 & 100 & 100 \\
\bottomrule
\end{tabular}
}
\caption{Baseline accuracy (\%) of each model on each task, measured on $\mathcal{D}^\mathrm{train}_T$.}
\label{tab:baseline_accuracy}
\end{table}

\section{Robustness to the Ablation Method}
\label{sec:ablation_robustness}

Our experiments measure \causalmetric~and specificity using zero ablation, which sets the output activation of each ablated component to zero. However, zero is an arbitrary value to the model (i.e. there's no reason to expect that zero is a neutral/baseline state for component activations), so various works have explored other ablation methods~\citep{zhang2024towards,li2024optimal}.

A popular alternative is mean ablation, which replaces the activation with its mean over some reference distribution of activations, such as the counterfactual/corrupted inputs~\citep{wang2022interpretability}. Other methods include resample ablation (replacing the activation with its value on a corrupted/counterfactual input;~\citet{heimersheim2024useinterpretactivationpatching}), optimal ablation (replacing the activation with a constant chosen to minimize the model's loss;~\citet{li2024optimal}), and Gaussian noising (corrupting the input embeddings with Gaussian noise;~\citet{meng2022locating}). To understand whether our findings are robust to the choice of ablation method, we re-run the main experiments of \cref{sec:within_task,sec:cross_task} under mean ablation, using the average activation over the corrupted inputs of the target task. Alternatively, one could ablate on a per-input basis, replacing each component's activation on a given input with its activation on that input's paired counterfactual/corrupted input (i.e., resample ablation).

\subsection{Mean ablation}
\label{sec:mean_ablation_method}

Let $S \subseteq \mathcal{V}$ be the circuit to ablate, and let $\mathcal{D}^{\mathrm{corr}}_T$ be the corrupted inputs of the target task. For each $c \in S$ and each token position $i$, let $\bar{a}_{c,i}$ denote the mean activation of $c$ at position $i$ over the inputs in $\mathcal{D}^{\mathrm{corr}}_T$ of length at least $i$, and let $\bar{a}_S := (\bar{a}_{c,i})_{c \in S,i}$ denote the resulting family of mean activations.

Mean ablation is defined as the intervention $\mathrm{do}(S \leftarrow \bar{a}_S)$, which clamps the activation of every $c \in S$ at each position $i$ to $\bar{a}_{c,i}$. Positions beyond the longest corrupted input receive the mean over all positions, following \citet{wang2022interpretability}. The mean-ablated prediction and accuracy are then \cref{eq:zap,eq:acc} with $\mathrm{do}(S \leftarrow 0)$ replaced by $\mathrm{do}(S \leftarrow \bar{a}_S)$.

\subsection{Experimental Setup}

\begin{figure}[t]
    \centering
    \includegraphics[width=\textwidth]{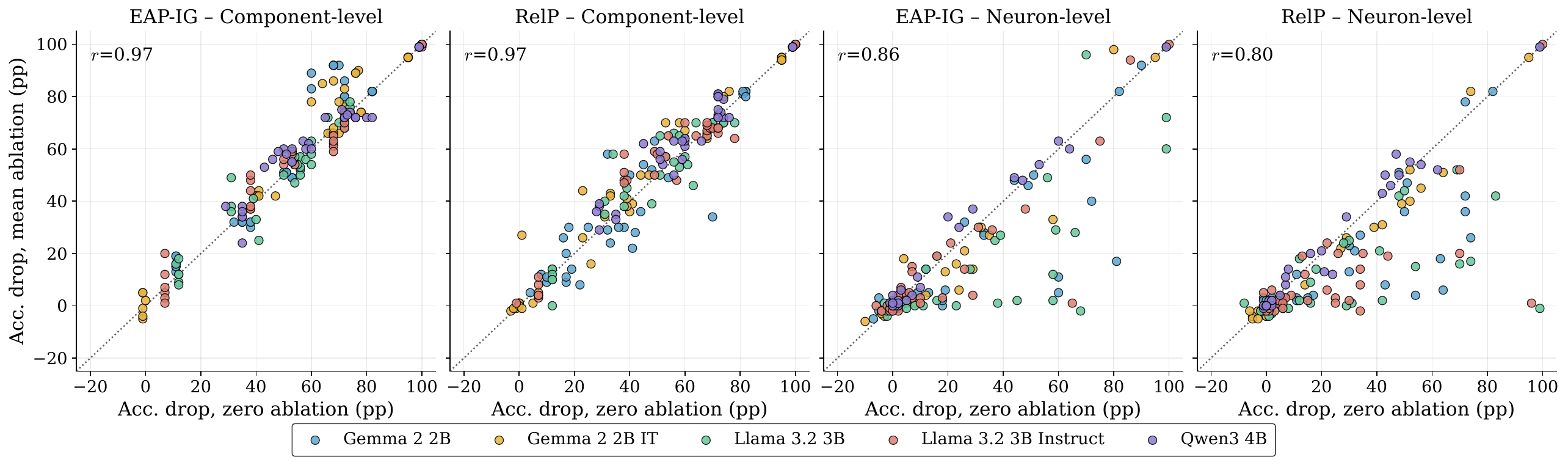}
    \caption{\textbf{Accuracy drop under zero and mean ablation.} One point per task pair (the drop on one task when the other's circuit, or its own, is ablated) at top-$K=10\%$ and $P=50\%$.}
    \label{fig:zero_vs_mean_cells}
\end{figure}

At the component level we use the EAP-IG and RelP circuits obtained at top-$K=10\%$ and $P$ of $50$, $75$, $85$ and $100\%$. At the neuron level we use a $P$ of $50$, $75$ and $85\%$, since neuron-level circuits are empty at $P=100\%$, for the five main models. In \cref{fig:zero_vs_mean_cells} we show the accuracy drops of every task pair at top-$K=10\%$ and $P=50\%$, the values used in \cref{sec:within_task,sec:cross_task}, and in \cref{tab:zero_vs_mean} we report correlations between mean and zero ablation over the full sweep of $P$.

\begin{table}[htbp]
\centering
\footnotesize
\setlength{\tabcolsep}{4pt}
\renewcommand{\arraystretch}{1.1}
\begin{tabular}{ll ccc ccc ccc ccc}
\toprule
& & \multicolumn{3}{c}{$P$=50\%} & \multicolumn{3}{c}{$P$=75\%} & \multicolumn{3}{c}{$P$=85\%} & \multicolumn{3}{c}{$P$=100\%} \\
\cmidrule(lr){3-5}\cmidrule(lr){6-8}\cmidrule(lr){9-11}\cmidrule(lr){12-14}
Granularity & Method & Drop & $\nu$ & $\sigma$ & Drop & $\nu$ & $\sigma$ & Drop & $\nu$ & $\sigma$ & Drop & $\nu$ & $\sigma$ \\
\midrule
Components & EAP-IG & 0.97 & 0.96 & -0.35 & 0.97 & 0.97 & 0.40 & 0.97 & 0.97 & 0.58 & 0.90 & 0.96 & 0.89 \\
Components & RelP & 0.97 & 0.95 & 0.86 & 0.96 & 0.91 & 0.96 & 0.96 & 0.93 & 0.96 & 0.91 & 0.93 & 0.94 \\
\addlinespace
Neurons & EAP-IG & 0.86 & 0.96 & 0.93 & 0.98 & 0.99 & 1.00 & 0.97 & 0.99 & 0.99 & -- & -- & -- \\
Neurons & RelP & 0.80 & 0.94 & 0.85 & 0.97 & 0.98 & 0.98 & 0.91 & 0.97 & 0.97 & -- & -- & -- \\
\bottomrule
\end{tabular}
\caption{\textbf{Agreement between zero and mean ablation.} Pearson correlations between accuracy drops under mean and zero ablation across task pairs ($n=180$ per condition), and between the corresponding \causalmetric~($\nu$) and specificity gap ($\sigma$) values across model--task pairs ($n=30$). All results use top-$K=10\%$.}
\label{tab:zero_vs_mean}
\end{table}

\subsection{Component-level circuits}
\label{sec:component_ablation_robustness}

The findings of \cref{sec:within_task,sec:cross_task} hold under mean ablation for component-level circuits. Accuracy drops under mean and zero ablation are highly correlated for both EAP-IG and RelP ($r=0.97$ for both at $P=50\%$, \cref{fig:zero_vs_mean_cells}, and between $0.90$ and $0.97$ at the other values of $P$), and so is \causalmetric~($r$ between $0.91$ and $0.97$ at every $P$, \cref{tab:zero_vs_mean}). \Causalmetric~is slightly higher under mean ablation on average (0.59 against 0.56 for EAP-IG, 0.14 against 0.11 for RelP, \cref{tab:zero_vs_mean_per_model}).

For the specificity gap, EAP-IG circuits are near zero under both ablations (0.7 and 1.6 points on average, \cref{tab:zero_vs_mean_per_model}), so at low $P$ the gaps carry little signal to correlate and the corresponding entries of \cref{tab:zero_vs_mean} are uninformative (but included for completeness); at $P=100\%$ they correlate ($r=0.89$). RelP circuits have a small positive gap under both ablations (5.4 and 6.9 points), and their gaps are correlated across the two ablations at every $P$ ($r$ between $0.86$ and $0.96$).

\begin{table}[htbp]
\centering
\begin{tabular}{lcccccccc}
\toprule
& \multicolumn{2}{c}{EAP-IG comp.} & \multicolumn{2}{c}{RelP comp.} & \multicolumn{2}{c}{EAP-IG neurons} & \multicolumn{2}{c}{RelP neurons} \\
\cmidrule(lr){2-3}\cmidrule(lr){4-5}\cmidrule(lr){6-7}\cmidrule(lr){8-9}
Model & zero & mean & zero & mean & zero & mean & zero & mean \\
\midrule
\multicolumn{9}{l}{\textit{\Causalmetric}} \\
Gemma 2B & 0.47 & 0.53 & 0.11 & 0.12 & 0.57 & 0.56 & 0.52 & 0.49 \\
Gemma 2B IT & 0.63 & 0.68 & 0.08 & 0.10 & 0.41 & 0.42 & 0.50 & 0.47 \\
Llama 3B & 0.49 & 0.53 & 0.07 & 0.09 & 0.51 & 0.46 & 0.48 & 0.39 \\
Llama 3B IT & 0.51 & 0.54 & 0.21 & 0.23 & 0.42 & 0.46 & 0.49 & 0.36 \\
Qwen3 4B & 0.69 & 0.68 & 0.08 & 0.15 & 0.53 & 0.55 & 0.51 & 0.51 \\
Mean & 0.56 & 0.59 & 0.11 & 0.14 & 0.49 & 0.49 & 0.50 & 0.45 \\
\midrule
\multicolumn{9}{l}{\textit{Specificity gap (accuracy points)}} \\
Gemma 2B & 0.4 & 1.4 & 13.1 & 16.2 & 26.0 & 34.2 & 17.9 & 29.3 \\
Gemma 2B IT & 1.1 & 1.7 & 13.7 & 12.8 & 31.4 & 35.6 & 34.0 & 33.9 \\
Llama 3B & -0.6 & 2.9 & 0.1 & 2.0 & 19.9 & 32.2 & 17.5 & 26.4 \\
Llama 3B IT & 0.2 & 1.7 & 1.1 & -0.3 & 26.4 & 34.4 & 27.9 & 27.7 \\
Qwen3 4B & 2.7 & 0.4 & -1.0 & 3.5 & 43.5 & 44.1 & 36.7 & 36.3 \\
Mean & 0.7 & 1.6 & 5.4 & 6.9 & 29.4 & 36.1 & 26.8 & 30.7 \\
\bottomrule
\end{tabular}
\caption{\textbf{\Causalmetric~and specificity gap under zero and mean ablation} at top-$K$=10\% and $P$=50\%, averaged over the six tasks. \Causalmetric~is \cref{eq:causal_effect} and the specificity gap is in accuracy points.}
\label{tab:zero_vs_mean_per_model}
\end{table}

\subsection{Neuron-level circuits}
\label{sec:neuron_ablation_robustness}

At the neuron level, the findings of \cref{sec:within_task,sec:cross_task} also hold under mean ablation. Accuracy drops are correlated across the two ablations ($r=0.86$ for EAP-IG and $r=0.80$ for RelP at $P=50\%$, and $0.91$ to $0.98$ at $P=75$ and $85\%$), and \causalmetric~and the specificity gap are strongly correlated for both methods at every $P$ ($r$ between $0.85$ and $1.00$, \cref{tab:zero_vs_mean}). Neuron circuits are about as necessary under mean ablation as under zero ablation (0.49 against 0.49 for EAP-IG, 0.45 against 0.50 for RelP), and somewhat more specific (gaps of 36.1 against 29.4 points for EAP-IG and 30.7 against 26.8 for RelP, \cref{tab:zero_vs_mean_per_model}).
The largest differences are in the RelP circuits of the two \texttt{Llama-3.2-3B} models, whose \causalmetric~is lower under mean ablation (0.36 and 0.39 against 0.49 and 0.48), and on \texttt{Addition}, whose specificity gap is larger under mean ablation in \texttt{Gemma-2-2B} and both \texttt{Llama-3.2-3B} models.

Overall, our findings about \causalmetric~and specificity \textbf{hold} under mean ablation at both granularities and for both attribution methods.

\section{Sparse Feature Circuits}
\label{sec:sae_features}

We repeat the experiments of \cref{sec:within_task,sec:cross_task} using sparse autoencoder (SAE) latents as the circuit units, following \citet{marks2025sparse}. An SAE is trained to reconstruct a model activation, in our case the residual stream after a layer, using an overcomplete set of learned latents~\citep{bricken2023monosemanticity,huben2024sparse}. The latent activations are trained to be sparse\footnote{See \citet{elhage2022superposition} and \citet{bricken2023monosemanticity} for the motivation behind sparsity.}, typically through an $L_1$ penalty~\citep{bricken2023monosemanticity,huben2024sparse}, although other methods exist (\eg TopK~\citep{gao2025scaling} and JumpReLU~\citep{rajamanoharan2025jumping}). Latents are commonly interpreted as features used by the model during its computation, and thus we refer to circuits using these latents as \emph{feature circuits}~\citep{marks2025sparse}. We take advantage of the publicly released Gemma Scope SAEs for \texttt{Gemma-2-2B} for our experiments~\citep{lieberum-etal-2024-gemma}.

\subsection{Setup}
\label{sec:sae_setup}

We use the Gemma Scope JumpReLU SAEs of width 16k, one for the residual stream after each of the 26 layers, which gives 425,984 latents in total. For comparison, \texttt{Gemma-2-2B} has 239,616 MLP neurons and 234 components. A JumpReLU SAE sets a latent's activation to zero when its pre-activation is below a learned per-latent threshold and passes it through unchanged otherwise. The thresholds were trained with a penalty on $L_0$, the mean number of active latents per token, using straight-through estimators~\citep{rajamanoharan2025jumping}. At each layer we choose the SAE whose $L_0$ is closest to 100. On our task prompts, the fraction of unexplained variance ranges from 0.06 at layer 0 to 0.53 at layer 22.

Following \citet{marks2025sparse}, we splice each SAE into the forward pass, so that the residual stream becomes $x = \mathrm{decode}(a) + e$, where $a$ is the vector of latent activations and $e = x - \mathrm{decode}(a)$ is the SAE error term. We compute $e$ on the clean input and hold it fixed, so the model with the SAEs spliced in gives the same outputs as the original model. To ablate a latent, we set its activation in $a$ to zero and leave $e$ unchanged.

Since latent activations are sparse, only latents whose activation differs between the clean and the corrupted input receive a nonzero attribution score, which is about 10,700 to 31,100 latents per example. As a result, we take top-$K$ to be $K$\% of the latents scored on each example (not the entire set of latents), and we draw the capacity-conserved control from the latents that are scored on at least one training example.

\subsection{Consistency and \Causalmetric}
\label{sec:sae_within}

\begin{figure}[h]
  \centering
  \includegraphics[width=\textwidth]{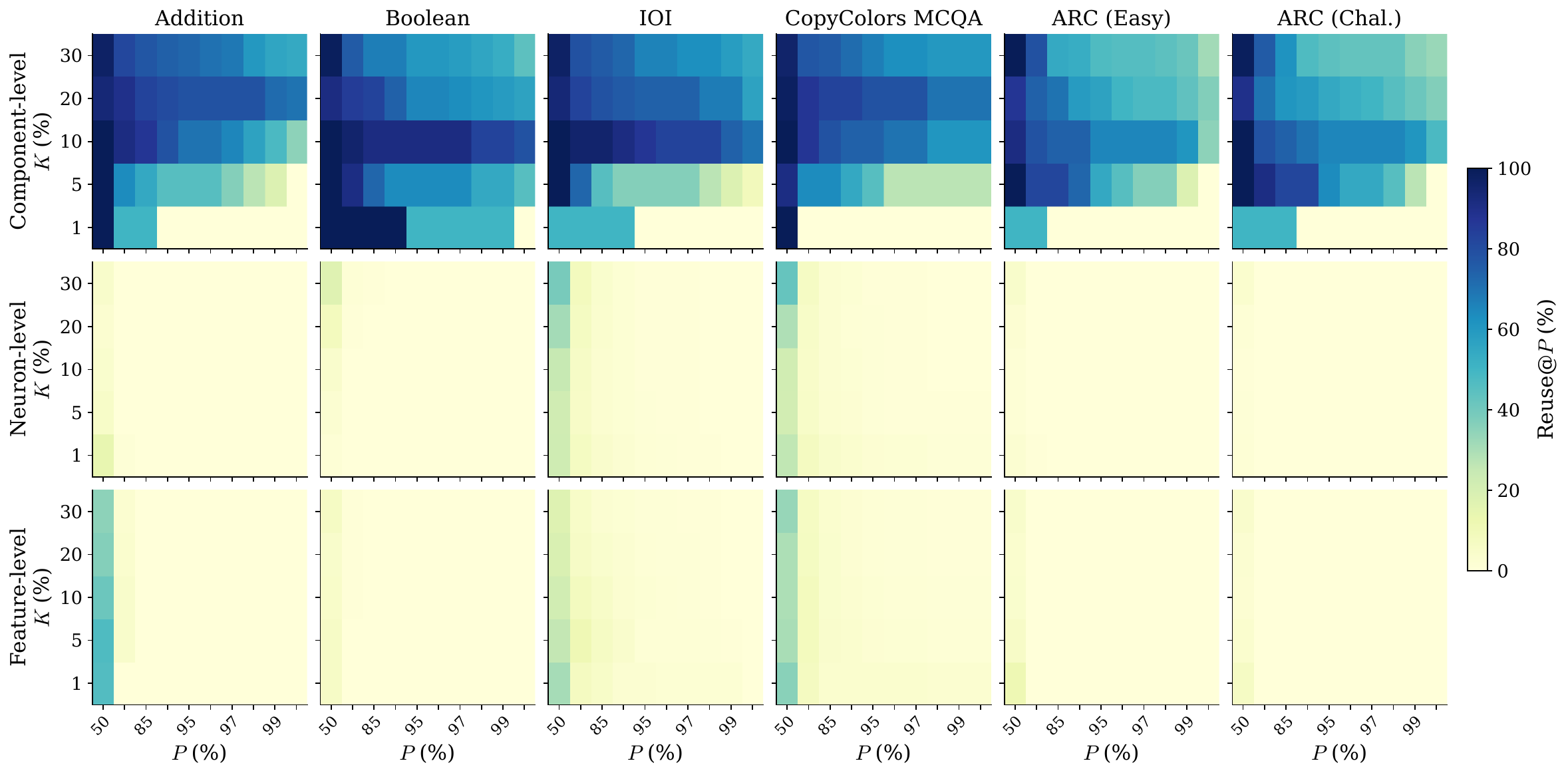}
  \caption{\textbf{\reuseat{P} over $K$ and $P$} for component, neuron and feature circuits, for \texttt{Gemma-2-2B} under EAP-IG.}
  \label{fig:sae_kp_reuse}
\end{figure}

\begin{figure}[t]
  \centering
  \includegraphics[width=\textwidth]{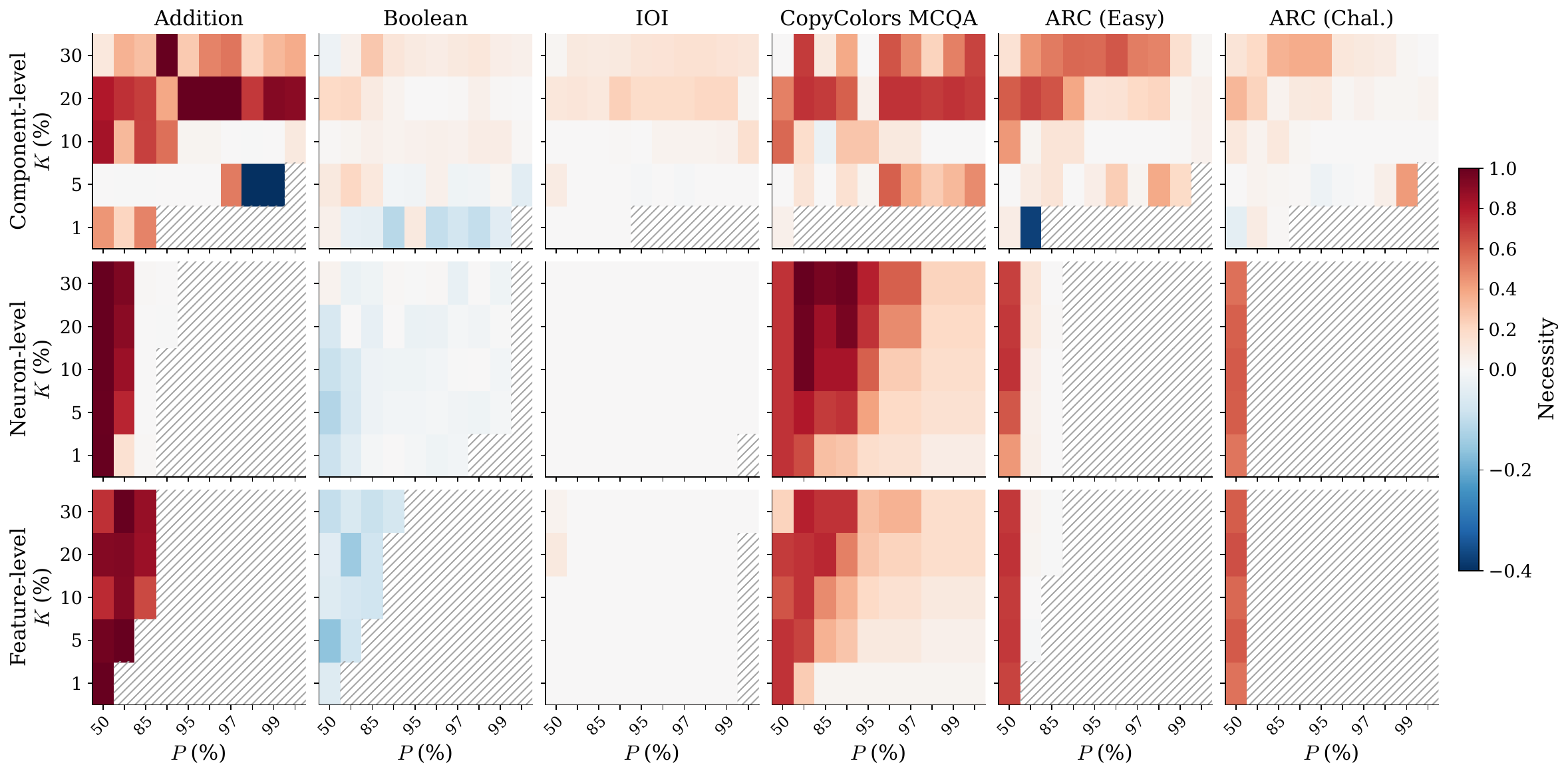}
  \caption{\textbf{\Causalmetric~over $K$ and $P$} for component, neuron and feature circuits, for \texttt{Gemma-2-2B} under EAP-IG. Hatched cells have an empty task circuit.}
  \label{fig:sae_kp_nec}
\end{figure}

In \cref{fig:sae_kp_reuse,fig:sae_kp_nec} we show \reuseat{P} and \causalmetric~for every task over all values of $K$ and $P$, where each row is a different granularity, so we can compare feature circuits with the component and neuron circuits from the main text.

We find that, for \texttt{Gemma-2-2B}, feature circuits behave much like neuron circuits. At $K$=10\% and $P$=50\%, \reuseat{P} is low for most tasks, and the tasks rank in nearly the same order as for neuron circuits, with \texttt{IOI} and \texttt{CopyColors~MCQA} (22 and 29\%) well above \texttt{Boolean~Logic} and the two \texttt{ARC} tasks (2 to 5\%). The one large difference is \texttt{Addition}, whose \reuseat{P} rises from 4\% for neurons to 41\% for latents, and this task also accounts for most of the difference in mean \reuseat{P} between the two granularities (17.3\% against 9.4\%). Both, however, remain far below component circuits, which reach a \reuseat{P} of 91 to 100\% on every task. These patterns hold across $K$, and as with neuron circuits, \reuseat{P} falls quickly as $P$ increases, with no task exceeding 8.4\% at $K$=10\% and $P$=75\%.

We also find that the \causalmetric~of feature circuits is close to that of neuron circuits on every task, across all values of $K$. At $K$=10\% and $P$=50\% it ranges from 0.57 to 0.75 for \texttt{Addition}, \texttt{CopyColors~MCQA} and the two \texttt{ARC} tasks, compared to 0.61 to 1.03 for neuron circuits, and it is near zero for \texttt{Boolean~Logic} and \texttt{IOI} at all three granularities. \texttt{Boolean~Logic} has a baseline of 0.51 on a two-way task, so its near-zero \causalmetric~is expected. For \texttt{IOI}, ablating the 256 latents shared at $P$=50\% leaves accuracy at 1.00, yet ablating each example's own top-10\% latents lowers the log-probability of the correct name by 12 to 20 nats. We therefore conclude that the latents shared across \texttt{IOI} examples encode the prompt template, while the latents that select the indirect object differ from example to example.

\subsection{Specificity}
\label{sec:sae_cross}

\begin{figure}[t]
  \centering
  \includegraphics[width=\textwidth]{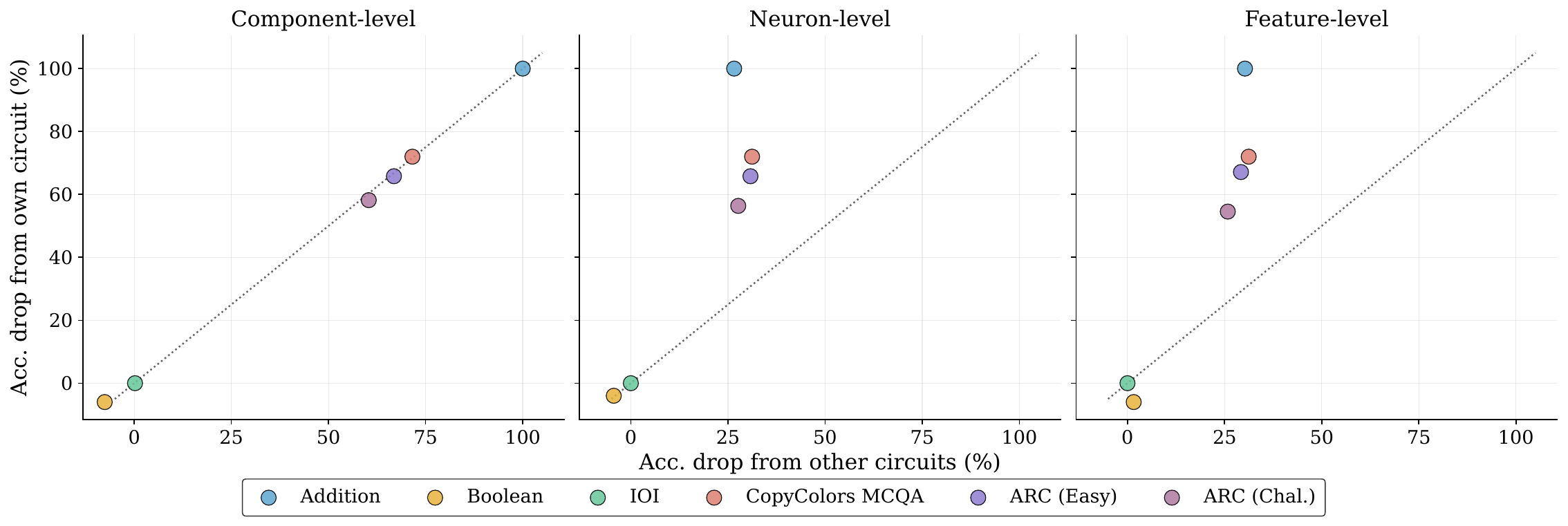}
  \caption{\textbf{Accuracy drop from ablating each task's own circuit against the mean drop from ablating the other tasks' circuits}, for \texttt{Gemma-2-2B} under EAP-IG at $K$=10\% and $P$=50\%. Drops are measured as a percentage of the task's baseline accuracy, and points above the diagonal indicate a specific circuit.}
  \label{fig:sae_own_other}
\end{figure}

In \cref{fig:sae_own_other} we plot, for each task, the accuracy drop from ablating its own circuit against the mean drop from ablating the other five tasks' circuits, as in \cref{fig:overlap_own_other}. Each accuracy drop is reported as a percentage of the task's baseline accuracy, so that tasks with low baselines can be compared with the other tasks.

We find that feature circuits are also as specific as neuron circuits, and that the two agree closely on each task. At $K$=10\% and $P$=50\%, the mean specificity gap is 28.3\% for feature circuits and 29.7\% for neuron circuits, and the gaps for individual tasks differ by at most 8 points between the two granularities. For both, ablating a task's own circuit costs \texttt{Addition} all of its accuracy and costs \texttt{CopyColors~MCQA} and the two \texttt{ARC} tasks 55 to 72\% of theirs, whereas ablating another task's circuit costs these tasks 26 to 31\%. Component circuits, in contrast, lie on the diagonal, with a mean gap of $-0.2$\%. \texttt{Boolean~Logic} and \texttt{IOI} show no gap at any granularity, which is consistent with their near-zero \causalmetric.

\subsection{Circuit size}
\label{sec:sae_size}

\begin{figure}[b]
  \centering
  \includegraphics[width=\textwidth]{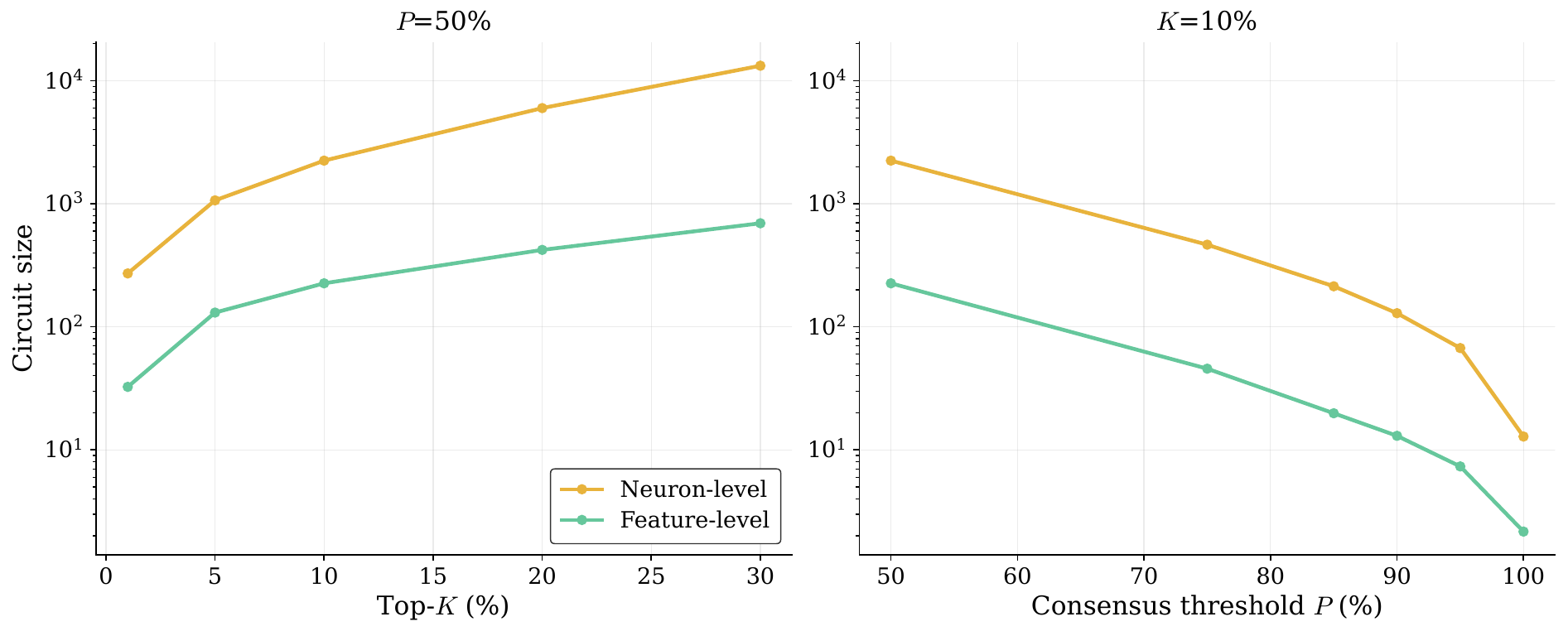}
  \caption{\textbf{Mean task circuit size of neuron and feature circuits}, against $K$ at $P$=50\% (left) and against $P$ at $K$=10\% (right), for \texttt{Gemma-2-2B} under EAP-IG.}
  \label{fig:sae_size}
\end{figure}

Since feature circuits are the same as neuron circuits in terms of \causalmetric~and specificity, we next compare how large these circuits are. In \cref{fig:sae_size} we show the mean size of the tasks circuits for neurons and SAE latents, showing the trend across $K$ at $P$=50\% and across $P$ at $K$=10\%.

We observe that feature circuits are roughly ten times smaller than neuron circuits. For example, at $K$=10\% and $P$=50\%, the average circuit has 226 latents against 2,242 neurons, and they stays between 9 and 11 times larger up to $P$=95\%. Across $K$ at $P$=50\%, the ratio grows from 8 at $K$=1\% to 19 at $K$=30\%. Nonetheless, feature circuits reach a similar \causalmetric~and specificity to neuron circuits with about a tenth of the units.

\subsection{Dictionary size and attribution method}
\label{sec:sae_robustness}

The difference in consistency between SAE latents and neurons could be due to dictionary size. In particular, larger SAEs tend to split a latent into several more specific latents~\citep{bricken2023monosemanticity}, which could lower \reuseat{P}. To test this, we recompute \reuseat{P} and \causalmetric~for width-65k SAEs on the three tasks with the highest \reuseat{P}.
The width-65k SAEs have 1,703,936 latents in total, about seven times the number of MLP neurons.

\begin{figure}[htbp]
  \centering
  \includegraphics[width=\textwidth]{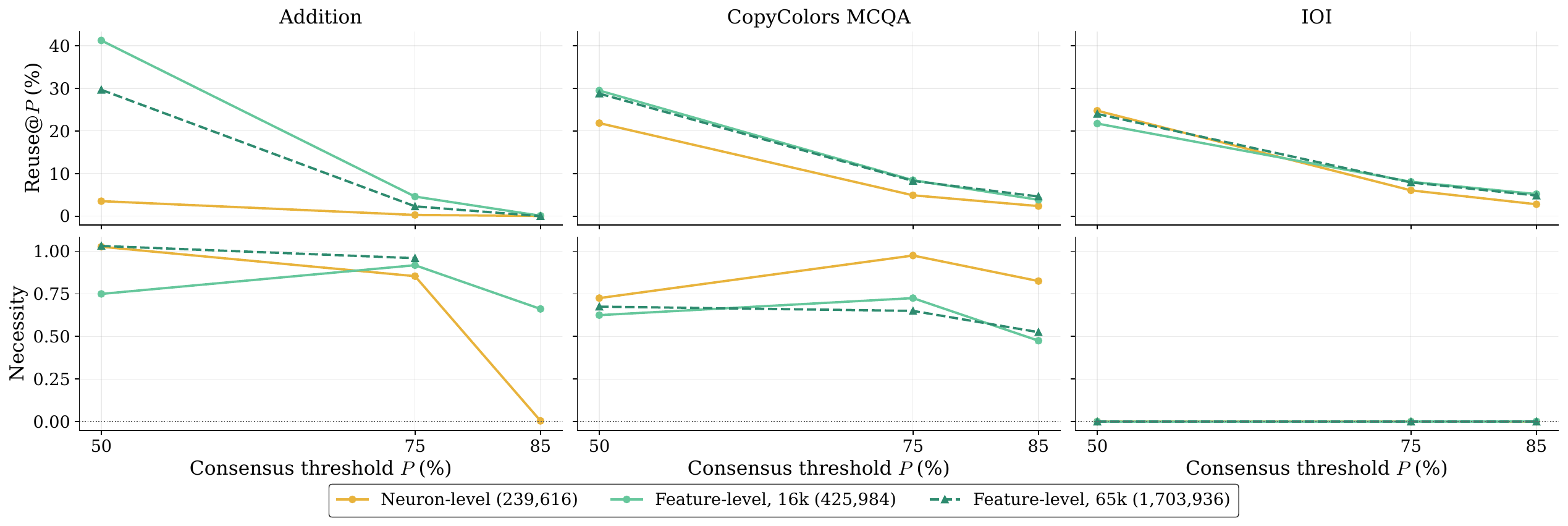}
  \caption{\textbf{Dictionary-size control.} \reuseat{P} and \causalmetric~for the width-16k and width-65k SAEs and for neurons, for \texttt{Addition}, \texttt{CopyColors~MCQA}, and \texttt{IOI} .}
  \label{fig:sae_dictionary}
\end{figure}

\cref{fig:sae_dictionary} shows that, despite using larger SAEs, \reuseat{P} stays the same for \texttt{CopyColors~MCQA} and \texttt{IOI} and actually drops slightly for \texttt{Addition}. Therefore, even at larger dictionary sizes, feature circuits remain more consistent than neuron circuits.

\begin{figure}[htbp]
  \centering
  \includegraphics[width=0.9\textwidth]{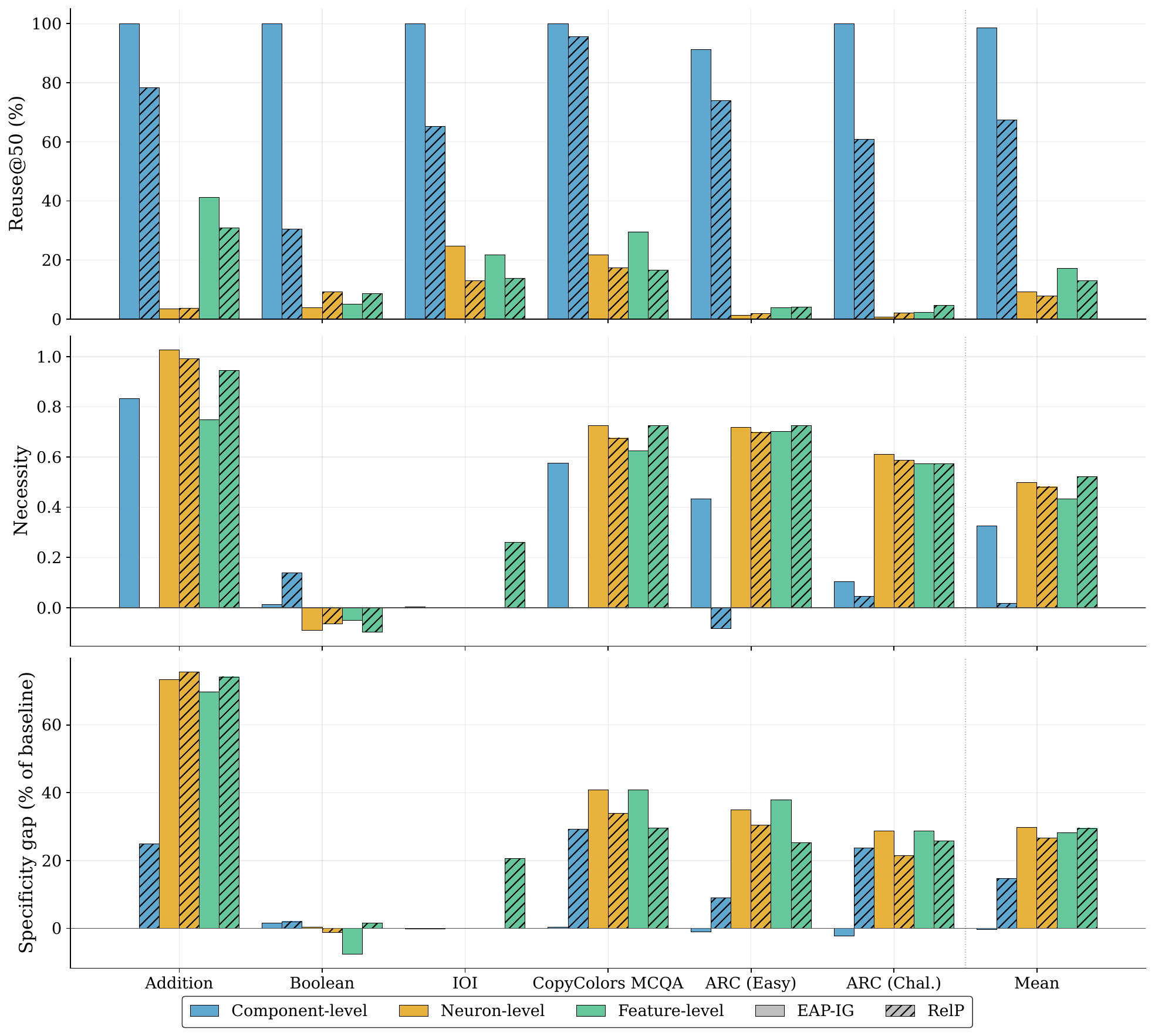}
  \caption{\textbf{EAP-IG and RelP} at $K$=10\% and $P$=50\%. Hatched bars are RelP.}
  \label{fig:sae_methods}
\end{figure}

In \cref{fig:sae_methods} we compare results between EAP-IG and RelP, at $K$=10\% and $P$=50\%. We find that the two attribution methods largely agree. The correlation between EAP-IG and RelP is $r$=0.92 for \reuseat{P}, 0.73 for \causalmetric~and 0.87 for the specificity gap. Under both methods, component circuits have the highest \reuseat{P}, the lowest \causalmetric~and the smallest specificity gap, while neuron and feature circuits are close to each other on all three properties. RelP does, however, find less \reuseat{P} than EAP-IG, especially for component circuits.

\section{Circuit Size and Composition}
\label{sec:circuit_anatomy}

\subsection{Agreement Between Attribution Methods}
\label{sec:method_agreement}

\begin{figure*}[htbp]
  \centering
  \includegraphics[width=\textwidth]{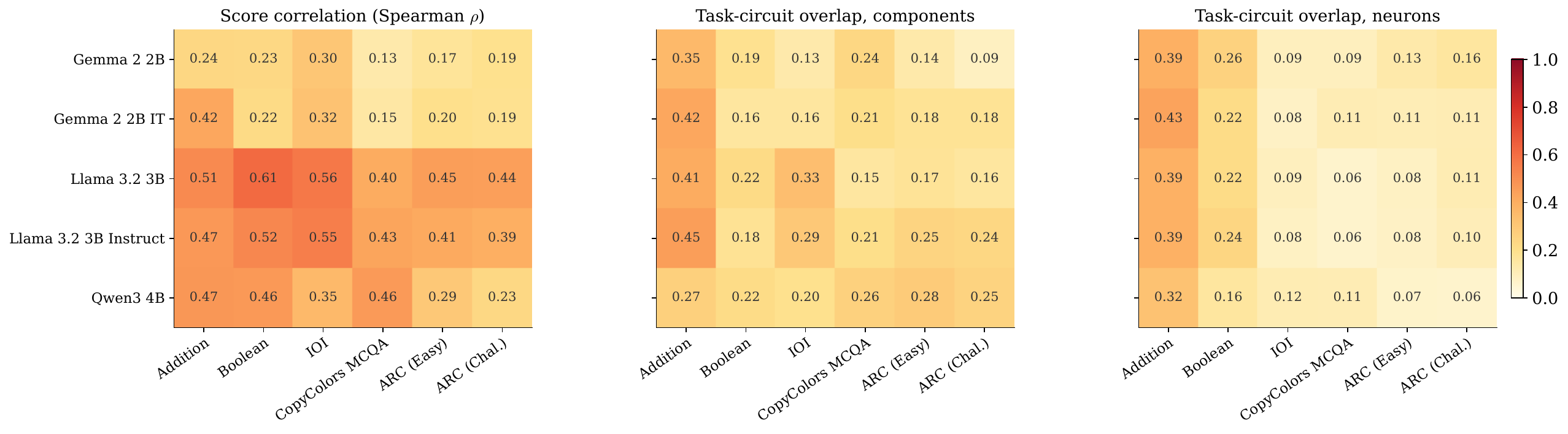}
  \caption{\textbf{Agreement between EAP-IG and RelP for each model and task.} \emph{Left:}~Spearman correlation between EAP-IG component scores and the magnitude of RelP component scores on the same example, averaged over examples. \emph{Middle and right:}~Jaccard overlap between the EAP-IG and RelP task circuits at $K$=10\% and $P$=50\%, at the component and neuron level.}
  \label{fig:method_agreement}
\end{figure*}

In \cref{fig:method_agreement} we show, for every model and task, the correlations between EAP-IG and RelP circuits. Specifically, we compute the Spearman correlation in the scores they assign to circuit units and the overlap between the task circuits they produce.
The scores agree the most in the two \texttt{Llama-3.2-3B} models (Spearman $\rho$ of 0.39 to 0.61) and the least in the \texttt{Gemma-2} models (0.13 to 0.42).
Agreement in scores does not carry over to the circuits, since even in the \texttt{Llama-3.2-3B} models the two methods' component-level task circuits overlap by only 0.15 to 0.45.
At the neuron level, only the circuits of \texttt{Addition} overlap by more than 0.3 (0.32 to 0.43).

\subsection{Circuit Size}
\label{sec:circuit_size}

\begin{figure}[htbp]
  \centering
  \includegraphics[width=\linewidth]{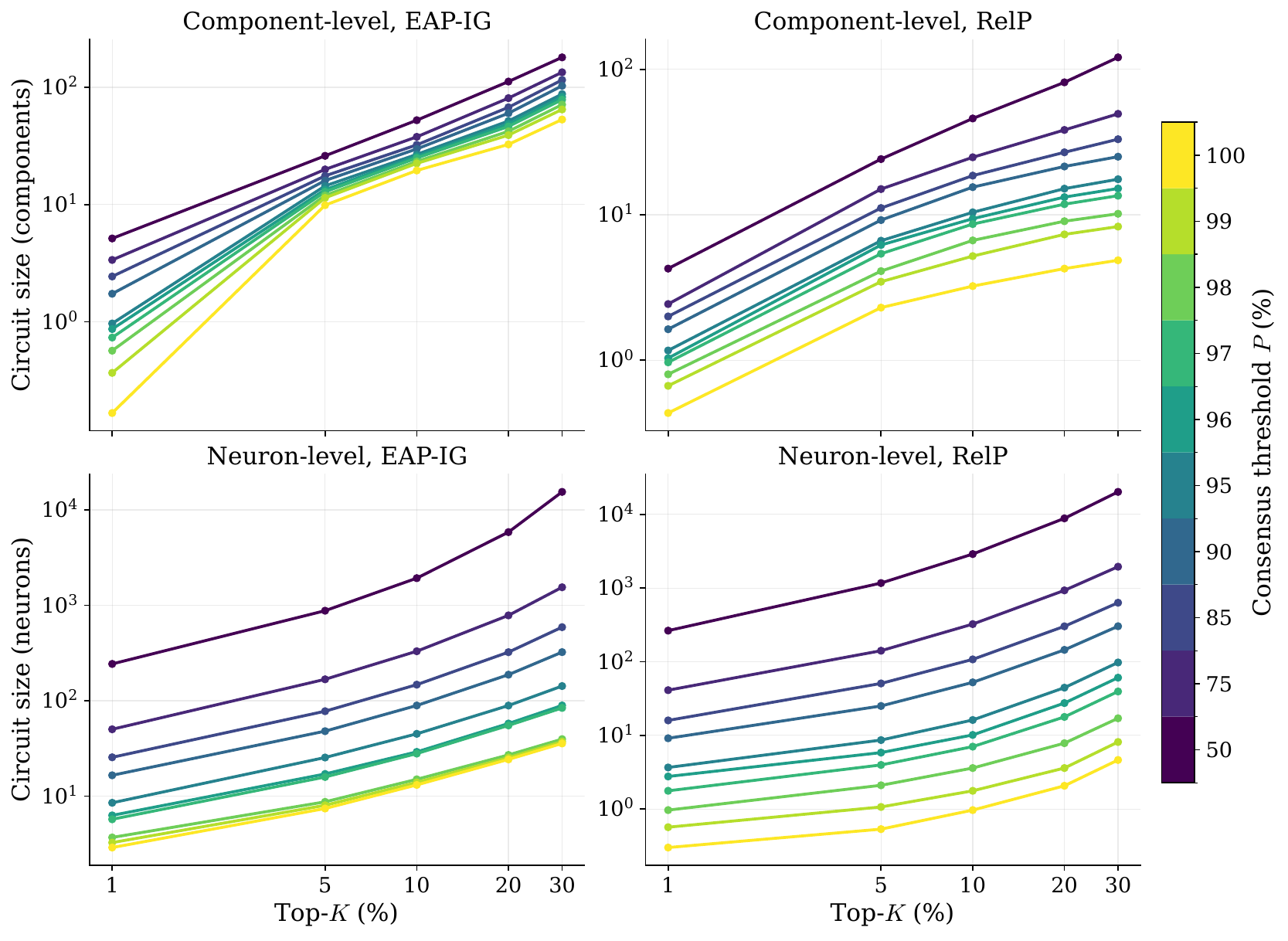}
  \caption{\textbf{Task circuit size across $K$ and $P$.} Number of components (top) and neurons (bottom) in the task circuit across top-$K$, one line per consensus threshold $P$, for EAP-IG (left) and RelP (right), averaged over models and tasks with empty circuits counted as zero. Both axes are logarithmic.}
  \label{fig:circuit_size}
\end{figure}

\Cref{fig:circuit_size} reports the absolute size of the within-task circuits across top-$K$ and consensus thresholds $P$.
The two granularities differ substantially in magnitude.
At $K$=10\% and $P$=50\%, component-level task circuits typically contain tens of attention heads and MLP blocks, whereas neuron-level task circuits contain hundreds to thousands of neurons.
The size of EAP-IG component-level circuits is set mainly by $K$, as raising $P$ from 50\% to 100\% shrinks them by a factor of 2.6 to 3.4 for $K\geq5\%$, whereas RelP component-level circuits and neuron-level circuits of both methods shrink much faster with $P$ (neuron-level circuits by a factor of 5 to 10 from $P$=50\% to 75\% alone).

\subsection{Component Types}
\label{sec:component_fraction_all}

\Cref{tab:composition_all_eap_ig_head_mlp,tab:composition_all_relp_head_mlp} report the attention head and MLP block percentages of each task circuit across all models, tasks, and top-$K$, at $P$=50\%.
The neuron granularity has no counterpart, since it decomposes the MLPs alone.

\subsection{Component Counts by Model}
\label{sec:component_tables}

Tables~\ref{tab:comp-google-gemma-2-2b}--\ref{tab:comp-qwen3-4b} report the mean number of attention heads and MLP blocks in each circuit for all models and tasks under EAP-IG, extending the summary in the main text to the full set of models.

\subsection{Cross-Task Circuit Decomposition Sizes}
\label{sec:decomposition_sizes}

The analysis above considers the size of the within-task circuit $S_P$.
For the cross-task experiments, we additionally partition pairs of task circuits into their shared core, task-only set, and partner-only set, as described in \cref{sec:cross_task}.
\Cref{tab:decomposition_all_eap_ig_head_mlp,tab:decomposition_all_relp_head_mlp,tab:decomposition_all_eap_ig_neuron,tab:decomposition_all_relp_neuron}
give the mean number of components in each partition for all models and tasks, averaged over the five partner tasks, at $P$=50\%.

The fraction of the union contained in the shared core differs across attribution methods.
Under EAP-IG it is 38--61\% at $K$=5\% and 53--62\% at $K$=30\%, with the largest values at $K$=10\%, where the Gemma models reach 83\% and 88\% compared with 45--49\% for the Llama and Qwen families.
Under RelP the shared core is smaller, between 11\% and 31\% at every $K$ and in every model, consistent with its lower cross-task overlap.


\begin{table*}[htbp]
\centering
\small
\setlength{\tabcolsep}{3pt}
\begin{tabular}{clcccccc}
\toprule
$K$ & Model & Addition & ARC (Chal.) & ARC (Easy) & Boolean & IOI & CopyColors MCQA \\
\midrule
1\% & \texttt{Gemma 2 2B} & 0/100 & 0/100 & 0/100 & 0/100 & 0/100 & 0/100 \\
 & \texttt{Gemma 2 2B Instruct} & 0/100 & 0/100 & 0/100 & -- & -- & 0/100 \\
 & \texttt{Llama-3.2-3B} & 33/67 & 0/100 & 0/100 & 0/100 & 40/60 & 33/67 \\
 & \texttt{Llama-3.2-3B Instruct} & 17/83 & 0/100 & 0/100 & 0/100 & 29/71 & 17/83 \\
 & \texttt{Qwen3-4B} & 0/100 & 0/100 & 0/100 & 0/100 & 9/91 & 45/55 \\
\midrule
5\% & \texttt{Gemma 2 2B} & 0/100 & 0/100 & 0/100 & 0/100 & 0/100 & 0/100 \\
 & \texttt{Gemma 2 2B Instruct} & 0/100 & 0/100 & 0/100 & 0/100 & 0/100 & 0/100 \\
 & \texttt{Llama-3.2-3B} & 18/82 & 38/62 & 36/64 & 42/58 & 34/66 & 50/50 \\
 & \texttt{Llama-3.2-3B Instruct} & 15/85 & 33/67 & 37/63 & 44/56 & 41/59 & 45/55 \\
 & \texttt{Qwen3-4B} & 38/62 & 48/52 & 39/61 & 47/53 & 40/60 & 42/58 \\
\midrule
10\% & \texttt{Gemma 2 2B} & 0/100 & 17/83 & 4/96 & 5/95 & 5/95 & 8/92 \\
 & \texttt{Gemma 2 2B Instruct} & 0/100 & 0/100 & 0/100 & 5/95 & 5/95 & 0/100 \\
 & \texttt{Llama-3.2-3B} & 51/49 & 71/29 & 66/34 & 74/26 & 65/35 & 67/33 \\
 & \texttt{Llama-3.2-3B Instruct} & 53/47 & 65/35 & 62/38 & 70/30 & 65/35 & 66/34 \\
 & \texttt{Qwen3-4B} & 65/35 & 67/33 & 63/37 & 67/33 & 71/29 & 67/33 \\
\midrule
20\% & \texttt{Gemma 2 2B} & 40/60 & 40/60 & 35/65 & 48/52 & 42/58 & 38/62 \\
 & \texttt{Gemma 2 2B Instruct} & 43/57 & 35/65 & 37/63 & 48/52 & 47/53 & 42/58 \\
 & \texttt{Llama-3.2-3B} & 78/22 & 84/16 & 82/18 & 85/15 & 79/21 & 80/20 \\
 & \texttt{Llama-3.2-3B Instruct} & 79/21 & 81/19 & 78/22 & 85/15 & 80/20 & 80/20 \\
 & \texttt{Qwen3-4B} & 84/16 & 83/17 & 82/18 & 84/16 & 83/17 & 84/16 \\
\midrule
30\% & \texttt{Gemma 2 2B} & 62/38 & 64/36 & 63/37 & 64/36 & 63/37 & 63/37 \\
 & \texttt{Gemma 2 2B Instruct} & 60/40 & 62/38 & 61/39 & 64/36 & 60/40 & 61/39 \\
 & \texttt{Llama-3.2-3B} & 86/14 & 89/11 & 88/12 & 89/11 & 87/13 & 87/13 \\
 & \texttt{Llama-3.2-3B Instruct} & 86/14 & 87/13 & 88/12 & 88/12 & 86/14 & 86/14 \\
 & \texttt{Qwen3-4B} & 90/10 & 90/10 & 90/10 & 90/10 & 89/11 & 89/11 \\
\bottomrule
\end{tabular}
\caption{Attention head / MLP block percentage of the shared set under EAP-IG, at $P$=50\%. A dash marks an empty shared set.}
\label{tab:composition_all_eap_ig_head_mlp}
\end{table*}

\begin{table*}[htbp]
\centering
\small
\setlength{\tabcolsep}{3pt}
\begin{tabular}{clcccccc}
\toprule
$K$ & Model & Addition & ARC (Chal.) & ARC (Easy) & Boolean & IOI & CopyColors MCQA \\
\midrule
1\% & \texttt{Gemma 2 2B} & 0/100 & 100/0 & 100/0 & -- & 100/0 & 100/0 \\
 & \texttt{Gemma 2 2B Instruct} & 0/100 & 100/0 & 100/0 & -- & 100/0 & 100/0 \\
 & \texttt{Llama-3.2-3B} & 0/100 & 75/25 & 75/25 & 67/33 & 100/0 & 83/17 \\
 & \texttt{Llama-3.2-3B Instruct} & 0/100 & 75/25 & 75/25 & 100/0 & 100/0 & 83/17 \\
 & \texttt{Qwen3-4B} & 18/82 & 58/42 & 55/45 & 50/50 & 100/0 & 60/40 \\
\midrule
5\% & \texttt{Gemma 2 2B} & 12/88 & 50/50 & 40/60 & 62/38 & 100/0 & 55/45 \\
 & \texttt{Gemma 2 2B Instruct} & 22/78 & 55/45 & 55/45 & 50/50 & 100/0 & 55/45 \\
 & \texttt{Llama-3.2-3B} & 33/67 & 71/29 & 68/32 & 60/40 & 62/38 & 75/25 \\
 & \texttt{Llama-3.2-3B Instruct} & 32/68 & 68/32 & 69/31 & 54/46 & 72/28 & 70/30 \\
 & \texttt{Qwen3-4B} & 57/43 & 58/42 & 57/43 & 69/31 & 69/31 & 72/28 \\
\midrule
10\% & \texttt{Gemma 2 2B} & 39/61 & 47/53 & 53/47 & 68/32 & 81/19 & 55/45 \\
 & \texttt{Gemma 2 2B Instruct} & 38/62 & 56/44 & 59/41 & 68/32 & 73/27 & 57/43 \\
 & \texttt{Llama-3.2-3B} & 47/53 & 59/41 & 64/36 & 67/33 & 60/40 & 82/18 \\
 & \texttt{Llama-3.2-3B Instruct} & 44/56 & 54/46 & 55/45 & 62/38 & 63/37 & 75/25 \\
 & \texttt{Qwen3-4B} & 66/34 & 61/39 & 65/35 & 74/26 & 74/26 & 82/18 \\
\midrule
20\% & \texttt{Gemma 2 2B} & 52/48 & 52/48 & 58/42 & 73/27 & 81/19 & 62/38 \\
 & \texttt{Gemma 2 2B Instruct} & 53/47 & 56/44 & 64/36 & 69/31 & 77/23 & 67/33 \\
 & \texttt{Llama-3.2-3B} & 58/42 & 67/33 & 70/30 & 81/19 & 78/22 & 86/14 \\
 & \texttt{Llama-3.2-3B Instruct} & 56/44 & 63/37 & 67/33 & 74/26 & 78/22 & 84/16 \\
 & \texttt{Qwen3-4B} & 75/25 & 78/22 & 79/21 & 85/15 & 81/19 & 88/12 \\
\midrule
30\% & \texttt{Gemma 2 2B} & 59/41 & 57/43 & 65/35 & 76/24 & 85/15 & 69/31 \\
 & \texttt{Gemma 2 2B Instruct} & 62/38 & 62/38 & 64/36 & 76/24 & 78/22 & 72/28 \\
 & \texttt{Llama-3.2-3B} & 70/30 & 77/23 & 81/19 & 85/15 & 84/16 & 91/9 \\
 & \texttt{Llama-3.2-3B Instruct} & 69/31 & 76/24 & 78/22 & 83/17 & 86/14 & 89/11 \\
 & \texttt{Qwen3-4B} & 81/19 & 85/15 & 87/13 & 89/11 & 86/14 & 91/9 \\
\bottomrule
\end{tabular}
\caption{Attention head / MLP block percentage of the shared set under RelP, at $P$=50\%. A dash marks an empty shared set.}
\label{tab:composition_all_relp_head_mlp}
\end{table*}

\begin{table}[ht]
\centering
\small
\caption{Gemma 2 2B: mean attention heads\,/\,MLPs in the top-$K\%$ circuit.}
\label{tab:comp-google-gemma-2-2b}
\begin{tabular}{lrrrr}
\toprule
\textbf{Task} & $1\%$ & $10\%$ & $20\%$ & $30\%$ \\
\midrule
Addition & 0.0\,/\,3.0 & 0.3\,/\,23.7 & 21.2\,/\,25.8 & 45.1\,/\,25.9 \\
ARC (Chal.) & 0.0\,/\,3.0 & 4.0\,/\,20.0 & 23.9\,/\,23.1 & 46.6\,/\,24.4 \\
ARC (Easy) & 0.0\,/\,3.0 & 2.7\,/\,21.3 & 22.6\,/\,24.4 & 45.8\,/\,25.2 \\
Boolean & 0.0\,/\,3.0 & 3.1\,/\,20.9 & 24.2\,/\,22.8 & 47.2\,/\,23.8 \\
IOI & 0.0\,/\,3.0 & 1.8\,/\,22.2 & 21.9\,/\,25.1 & 45.4\,/\,25.6 \\
CopyColors MCQA & 0.0\,/\,3.0 & 1.8\,/\,22.2 & 21.2\,/\,25.8 & 45.1\,/\,25.9 \\
\bottomrule
\end{tabular}
\end{table}

\begin{table}[ht]
\centering
\small
\caption{Gemma 2 2B Instruct: mean attention heads\,/\,MLPs in the top-$K\%$ circuit.}
\label{tab:comp-google-gemma-2-2b-it}
\begin{tabular}{lrrrr}
\toprule
\textbf{Task} & $1\%$ & $10\%$ & $20\%$ & $30\%$ \\
\midrule
Addition & 0.0\,/\,3.0 & 0.1\,/\,23.9 & 21.0\,/\,26.0 & 45.0\,/\,26.0 \\
ARC (Chal.) & 0.0\,/\,3.0 & 2.5\,/\,21.5 & 22.7\,/\,24.3 & 45.9\,/\,25.1 \\
ARC (Easy) & 0.0\,/\,3.0 & 1.5\,/\,22.5 & 21.8\,/\,25.2 & 45.4\,/\,25.6 \\
Boolean & 0.0\,/\,3.0 & 2.7\,/\,21.3 & 23.8\,/\,23.2 & 47.0\,/\,24.0 \\
IOI & 0.0\,/\,3.0 & 2.0\,/\,22.0 & 22.0\,/\,25.0 & 45.4\,/\,25.6 \\
CopyColors MCQA & 0.0\,/\,3.0 & 1.4\,/\,22.6 & 21.2\,/\,25.8 & 45.0\,/\,25.9 \\
\bottomrule
\end{tabular}
\end{table}

\begin{table}[ht]
\centering
\small
\caption{Llama-3.2-3B: mean attention heads\,/\,MLPs in the top-$K\%$ circuit.}
\label{tab:comp-meta-llama-llama-3.2-3b}
\begin{tabular}{lrrrr}
\toprule
\textbf{Task} & $1\%$ & $10\%$ & $20\%$ & $30\%$ \\
\midrule
Addition & 1.1\,/\,5.9 & 42.0\,/\,28.0 & 112.0\,/\,28.0 & 182.0\,/\,28.0 \\
ARC (Chal.) & 0.4\,/\,6.6 & 52.9\,/\,17.1 & 119.9\,/\,20.1 & 187.3\,/\,22.7 \\
ARC (Easy) & 0.9\,/\,6.1 & 50.9\,/\,19.1 & 117.7\,/\,22.3 & 185.6\,/\,24.4 \\
Boolean & 0.7\,/\,6.3 & 52.6\,/\,17.4 & 119.5\,/\,20.5 & 186.6\,/\,23.4 \\
IOI & 2.6\,/\,4.4 & 48.2\,/\,21.8 & 114.2\,/\,25.8 & 182.9\,/\,27.1 \\
CopyColors MCQA & 2.1\,/\,4.9 & 49.6\,/\,20.4 & 114.8\,/\,25.2 & 183.0\,/\,27.0 \\
\bottomrule
\end{tabular}
\end{table}

\begin{table}[ht]
\centering
\small
\caption{Llama-3.2-3B Instruct: mean attention heads\,/\,MLPs in the top-$K\%$ circuit.}
\label{tab:comp-meta-llama-llama-3.2-3b-instruct}
\begin{tabular}{lrrrr}
\toprule
\textbf{Task} & $1\%$ & $10\%$ & $20\%$ & $30\%$ \\
\midrule
Addition & 1.2\,/\,5.8 & 42.0\,/\,28.0 & 112.0\,/\,28.0 & 182.0\,/\,28.0 \\
ARC (Chal.) & 0.2\,/\,6.8 & 50.4\,/\,19.6 & 117.0\,/\,23.0 & 184.8\,/\,25.2 \\
ARC (Easy) & 0.5\,/\,6.5 & 48.7\,/\,21.3 & 115.6\,/\,24.4 & 184.0\,/\,26.0 \\
Boolean & 0.7\,/\,6.3 & 52.7\,/\,17.3 & 119.1\,/\,20.9 & 185.8\,/\,24.2 \\
IOI & 2.7\,/\,4.3 & 48.4\,/\,21.6 & 114.1\,/\,25.9 & 182.7\,/\,27.3 \\
CopyColors MCQA & 1.4\,/\,5.7 & 48.6\,/\,21.4 & 113.4\,/\,26.6 & 182.6\,/\,27.4 \\
\bottomrule
\end{tabular}
\end{table}

\begin{table}[ht]
\centering
\small
\caption{Qwen3-4B: mean attention heads\,/\,MLPs in the top-$K\%$ circuit.}
\label{tab:comp-qwen3-4b}
\begin{tabular}{lrrrr}
\toprule
\textbf{Task} & $1\%$ & $10\%$ & $20\%$ & $30\%$ \\
\midrule
Addition & 1.0\,/\,11.0 & 83.8\,/\,35.1 & 202.5\,/\,35.5 & 321.4\,/\,35.6 \\
ARC (Chal.) & 0.3\,/\,11.7 & 89.4\,/\,29.6 & 205.8\,/\,32.2 & 323.4\,/\,33.6 \\
ARC (Easy) & 0.8\,/\,11.2 & 86.5\,/\,32.5 & 203.8\,/\,34.2 & 322.0\,/\,35.0 \\
Boolean & 0.4\,/\,11.6 & 88.5\,/\,30.5 & 204.8\,/\,33.2 & 322.6\,/\,34.4 \\
IOI & 0.8\,/\,11.2 & 88.7\,/\,30.3 & 204.0\,/\,34.0 & 321.6\,/\,35.4 \\
CopyColors MCQA & 3.6\,/\,8.4 & 84.2\,/\,34.8 & 202.4\,/\,35.6 & 321.3\,/\,35.7 \\
\bottomrule
\end{tabular}
\end{table}

\begin{table*}[htbp]
\centering
\small
\setlength{\tabcolsep}{3pt}
\resizebox{\textwidth}{!}{%
\begin{tabular}{clcccccc}
\toprule
$K$ & Model & Addition & ARC (Chal.) & ARC (Easy) & Boolean & IOI & CopyColors MCQA \\
\midrule
1\% & \texttt{Gemma 2 2B} & 0\,/\,2\,/\,1 & 1\,/\,0\,/\,1 & 1\,/\,0\,/\,1 & 1\,/\,0\,/\,1 & 1\,/\,0\,/\,1 & 1\,/\,1\,/\,0 \\
 & \texttt{Gemma 2 2B Instruct} & 0\,/\,2\,/\,1 & 0\,/\,1\,/\,1 & 0\,/\,1\,/\,1 & 0\,/\,0\,/\,1 & 0\,/\,0\,/\,1 & 0\,/\,1\,/\,1 \\
 & \texttt{Llama-3.2-3B} & 1\,/\,5\,/\,5 & 4\,/\,3\,/\,2 & 3\,/\,3\,/\,3 & 3\,/\,4\,/\,3 & 2\,/\,3\,/\,4 & 2\,/\,4\,/\,4 \\
 & \texttt{Llama-3.2-3B Instruct} & 1\,/\,5\,/\,5 & 4\,/\,3\,/\,3 & 4\,/\,3\,/\,3 & 3\,/\,4\,/\,4 & 3\,/\,4\,/\,3 & 2\,/\,4\,/\,4 \\
 & \texttt{Qwen3-4B} & 8\,/\,3\,/\,3 & 8\,/\,2\,/\,3 & 8\,/\,3\,/\,2 & 7\,/\,3\,/\,3 & 8\,/\,3\,/\,3 & 4\,/\,7\,/\,7 \\
\midrule
5\% & \texttt{Gemma 2 2B} & 4\,/\,7\,/\,7 & 8\,/\,3\,/\,3 & 8\,/\,3\,/\,3 & 9\,/\,3\,/\,2 & 6\,/\,4\,/\,5 & 8\,/\,3\,/\,3 \\
 & \texttt{Gemma 2 2B Instruct} & 4\,/\,6\,/\,7 & 7\,/\,4\,/\,3 & 7\,/\,3\,/\,4 & 8\,/\,5\,/\,3 & 7\,/\,3\,/\,4 & 7\,/\,3\,/\,4 \\
 & \texttt{Llama-3.2-3B} & 16\,/\,18\,/\,11 & 17\,/\,7\,/\,12 & 18\,/\,7\,/\,11 & 17\,/\,9\,/\,12 & 16\,/\,13\,/\,12 & 19\,/\,13\,/\,8 \\
 & \texttt{Llama-3.2-3B Instruct} & 17\,/\,16\,/\,11 & 19\,/\,5\,/\,11 & 20\,/\,7\,/\,9 & 18\,/\,9\,/\,11 & 18\,/\,14\,/\,10 & 20\,/\,9\,/\,9 \\
 & \texttt{Qwen3-4B} & 32\,/\,23\,/\,18 & 34\,/\,12\,/\,19 & 38\,/\,13\,/\,13 & 33\,/\,16\,/\,19 & 32\,/\,15\,/\,20 & 38\,/\,22\,/\,11 \\
\midrule
10\% & \texttt{Gemma 2 2B} & 20\,/\,4\,/\,3 & 19\,/\,4\,/\,4 & 20\,/\,3\,/\,3 & 20\,/\,2\,/\,3 & 20\,/\,2\,/\,3 & 21\,/\,4\,/\,2 \\
 & \texttt{Gemma 2 2B Instruct} & 20\,/\,3\,/\,3 & 20\,/\,2\,/\,2 & 21\,/\,2\,/\,2 & 20\,/\,2\,/\,3 & 20\,/\,2\,/\,3 & 21\,/\,3\,/\,2 \\
 & \texttt{Llama-3.2-3B} & 26\,/\,31\,/\,32 & 32\,/\,23\,/\,27 & 34\,/\,22\,/\,25 & 29\,/\,33\,/\,29 & 28\,/\,32\,/\,30 & 32\,/\,28\,/\,26 \\
 & \texttt{Llama-3.2-3B Instruct} & 27\,/\,33\,/\,33 & 36\,/\,21\,/\,24 & 37\,/\,21\,/\,23 & 31\,/\,25\,/\,29 & 30\,/\,32\,/\,29 & 35\,/\,29\,/\,23 \\
 & \texttt{Qwen3-4B} & 49\,/\,51\,/\,48 & 63\,/\,31\,/\,35 & 66\,/\,29\,/\,33 & 58\,/\,37\,/\,40 & 54\,/\,44\,/\,44 & 61\,/\,44\,/\,36 \\
\midrule
20\% & \texttt{Gemma 2 2B} & 32\,/\,11\,/\,10 & 30\,/\,10\,/\,12 & 32\,/\,8\,/\,10 & 29\,/\,13\,/\,13 & 30\,/\,15\,/\,11 & 32\,/\,10\,/\,10 \\
 & \texttt{Gemma 2 2B Instruct} & 33\,/\,13\,/\,10 & 32\,/\,8\,/\,12 & 33\,/\,8\,/\,11 & 31\,/\,13\,/\,12 & 31\,/\,16\,/\,12 & 34\,/\,11\,/\,9 \\
 & \texttt{Llama-3.2-3B} & 68\,/\,59\,/\,64 & 80\,/\,47\,/\,51 & 86\,/\,42\,/\,46 & 77\,/\,61\,/\,53 & 70\,/\,59\,/\,61 & 83\,/\,53\,/\,47 \\
 & \texttt{Llama-3.2-3B Instruct} & 70\,/\,61\,/\,57 & 79\,/\,36\,/\,52 & 85\,/\,40\,/\,43 & 73\,/\,57\,/\,54 & 68\,/\,62\,/\,59 & 83\,/\,53\,/\,43 \\
 & \texttt{Qwen3-4B} & 123\,/\,100\,/\,90 & 135\,/\,69\,/\,81 & 139\,/\,65\,/\,77 & 129\,/\,91\,/\,83 & 124\,/\,87\,/\,91 & 122\,/\,100\,/\,90 \\
\midrule
30\% & \texttt{Gemma 2 2B} & 51\,/\,17\,/\,19 & 55\,/\,18\,/\,14 & 54\,/\,16\,/\,16 & 51\,/\,16\,/\,19 & 48\,/\,23\,/\,22 & 54\,/\,17\,/\,16 \\
 & \texttt{Gemma 2 2B Instruct} & 48\,/\,17\,/\,19 & 51\,/\,18\,/\,15 & 50\,/\,16\,/\,16 & 51\,/\,16\,/\,16 & 46\,/\,19\,/\,20 & 52\,/\,15\,/\,14 \\
 & \texttt{Llama-3.2-3B} & 139\,/\,68\,/\,70 & 152\,/\,57\,/\,56 & 157\,/\,52\,/\,51 & 145\,/\,67\,/\,63 & 137\,/\,65\,/\,73 & 154\,/\,59\,/\,54 \\
 & \texttt{Llama-3.2-3B Instruct} & 129\,/\,72\,/\,81 & 159\,/\,55\,/\,48 & 166\,/\,50\,/\,41 & 153\,/\,58\,/\,55 & 144\,/\,58\,/\,66 & 154\,/\,53\,/\,55 \\
 & \texttt{Qwen3-4B} & 221\,/\,126\,/\,123 & 250\,/\,99\,/\,94 & 253\,/\,96\,/\,90 & 236\,/\,111\,/\,108 & 224\,/\,113\,/\,123 & 215\,/\,124\,/\,131 \\
\bottomrule
\end{tabular}
}
\caption{Mean shared core / task-only / partner-only sizes over attention heads and MLP blocks under EAP-IG, at $P$=50\%, averaged over the five partner tasks.}
\label{tab:decomposition_all_eap_ig_head_mlp}
\end{table*}

\begin{table*}[htbp]
\centering
\small
\setlength{\tabcolsep}{3pt}
\resizebox{\textwidth}{!}{%
\begin{tabular}{clcccccc}
\toprule
$K$ & Model & Addition & ARC (Chal.) & ARC (Easy) & Boolean & IOI & CopyColors MCQA \\
\midrule
1\% & \texttt{Gemma 2 2B} & 0\,/\,2\,/\,1 & 0\,/\,1\,/\,1 & 0\,/\,1\,/\,1 & 0\,/\,0\,/\,1 & 0\,/\,2\,/\,1 & 0\,/\,1\,/\,1 \\
 & \texttt{Gemma 2 2B Instruct} & 0\,/\,2\,/\,1 & 0\,/\,1\,/\,1 & 0\,/\,1\,/\,1 & 0\,/\,0\,/\,1 & 0\,/\,2\,/\,1 & 0\,/\,1\,/\,1 \\
 & \texttt{Llama-3.2-3B} & 1\,/\,7\,/\,4 & 2\,/\,2\,/\,3 & 2\,/\,2\,/\,3 & 1\,/\,5\,/\,4 & 0\,/\,3\,/\,6 & 2\,/\,4\,/\,3 \\
 & \texttt{Llama-3.2-3B Instruct} & 1\,/\,6\,/\,3 & 2\,/\,2\,/\,3 & 2\,/\,2\,/\,3 & 0\,/\,2\,/\,5 & 0\,/\,3\,/\,5 & 2\,/\,4\,/\,2 \\
 & \texttt{Qwen3-4B} & 4\,/\,7\,/\,5 & 6\,/\,6\,/\,3 & 6\,/\,5\,/\,3 & 1\,/\,7\,/\,9 & 2\,/\,3\,/\,9 & 5\,/\,5\,/\,4 \\
\midrule
5\% & \texttt{Gemma 2 2B} & 2\,/\,6\,/\,6 & 5\,/\,3\,/\,4 & 5\,/\,5\,/\,3 & 1\,/\,7\,/\,7 & 1\,/\,6\,/\,8 & 5\,/\,6\,/\,3 \\
 & \texttt{Gemma 2 2B Instruct} & 3\,/\,6\,/\,7 & 6\,/\,5\,/\,3 & 6\,/\,5\,/\,3 & 2\,/\,4\,/\,8 & 2\,/\,7\,/\,7 & 6\,/\,5\,/\,3 \\
 & \texttt{Llama-3.2-3B} & 12\,/\,18\,/\,16 & 14\,/\,10\,/\,14 & 17\,/\,11\,/\,11 & 11\,/\,14\,/\,18 & 13\,/\,16\,/\,15 & 17\,/\,15\,/\,10 \\
 & \texttt{Llama-3.2-3B Instruct} & 12\,/\,19\,/\,14 & 15\,/\,7\,/\,13 & 16\,/\,10\,/\,11 & 11\,/\,15\,/\,16 & 12\,/\,13\,/\,15 & 16\,/\,14\,/\,10 \\
 & \texttt{Qwen3-4B} & 20\,/\,27\,/\,28 & 30\,/\,13\,/\,19 & 31\,/\,16\,/\,17 & 22\,/\,30\,/\,25 & 23\,/\,26\,/\,25 & 29\,/\,21\,/\,19 \\
\midrule
10\% & \texttt{Gemma 2 2B} & 8\,/\,10\,/\,12 & 10\,/\,5\,/\,10 & 12\,/\,7\,/\,8 & 8\,/\,14\,/\,11 & 8\,/\,13\,/\,11 & 13\,/\,9\,/\,6 \\
 & \texttt{Gemma 2 2B Instruct} & 9\,/\,12\,/\,11 & 12\,/\,6\,/\,9 & 12\,/\,5\,/\,9 & 8\,/\,14\,/\,12 & 9\,/\,13\,/\,11 & 14\,/\,9\,/\,6 \\
 & \texttt{Llama-3.2-3B} & 23\,/\,24\,/\,30 & 29\,/\,12\,/\,26 & 30\,/\,15\,/\,23 & 25\,/\,35\,/\,25 & 30\,/\,30\,/\,21 & 28\,/\,32\,/\,22 \\
 & \texttt{Llama-3.2-3B Instruct} & 26\,/\,24\,/\,28 & 33\,/\,15\,/\,21 & 35\,/\,18\,/\,18 & 27\,/\,28\,/\,25 & 30\,/\,27\,/\,22 & 28\,/\,25\,/\,25 \\
 & \texttt{Qwen3-4B} & 29\,/\,38\,/\,59 & 48\,/\,22\,/\,40 & 51\,/\,27\,/\,36 & 45\,/\,63\,/\,35 & 41\,/\,57\,/\,42 & 47\,/\,42\,/\,37 \\
\midrule
20\% & \texttt{Gemma 2 2B} & 16\,/\,15\,/\,22 & 22\,/\,9\,/\,15 & 23\,/\,10\,/\,14 & 19\,/\,22\,/\,17 & 16\,/\,27\,/\,19 & 25\,/\,15\,/\,11 \\
 & \texttt{Gemma 2 2B Instruct} & 18\,/\,20\,/\,24 & 27\,/\,12\,/\,15 & 27\,/\,15\,/\,14 & 21\,/\,18\,/\,20 & 20\,/\,23\,/\,20 & 28\,/\,17\,/\,13 \\
 & \texttt{Llama-3.2-3B} & 39\,/\,28\,/\,57 & 46\,/\,17\,/\,50 & 51\,/\,23\,/\,43 & 46\,/\,72\,/\,40 & 47\,/\,65\,/\,40 & 50\,/\,60\,/\,37 \\
 & \texttt{Llama-3.2-3B Instruct} & 35\,/\,28\,/\,54 & 45\,/\,18\,/\,44 & 49\,/\,27\,/\,37 & 41\,/\,52\,/\,41 & 42\,/\,63\,/\,38 & 46\,/\,59\,/\,34 \\
 & \texttt{Qwen3-4B} & 44\,/\,58\,/\,122 & 90\,/\,51\,/\,68 & 94\,/\,60\,/\,61 & 81\,/\,109\,/\,67 & 67\,/\,94\,/\,87 & 84\,/\,98\,/\,65 \\
\midrule
30\% & \texttt{Gemma 2 2B} & 24\,/\,22\,/\,28 & 29\,/\,8\,/\,25 & 33\,/\,15\,/\,19 & 28\,/\,31\,/\,22 & 26\,/\,33\,/\,24 & 34\,/\,24\,/\,16 \\
 & \texttt{Gemma 2 2B Instruct} & 29\,/\,26\,/\,31 & 40\,/\,16\,/\,20 & 41\,/\,18\,/\,19 & 33\,/\,25\,/\,26 & 30\,/\,29\,/\,29 & 41\,/\,27\,/\,16 \\
 & \texttt{Llama-3.2-3B} & 52\,/\,42\,/\,88 & 67\,/\,30\,/\,73 & 73\,/\,44\,/\,63 & 69\,/\,100\,/\,57 & 66\,/\,94\,/\,61 & 72\,/\,88\,/\,55 \\
 & \texttt{Llama-3.2-3B Instruct} & 48\,/\,41\,/\,88 & 67\,/\,32\,/\,68 & 72\,/\,41\,/\,60 & 59\,/\,85\,/\,66 & 64\,/\,97\,/\,58 & 69\,/\,96\,/\,52 \\
 & \texttt{Qwen3-4B} & 71\,/\,85\,/\,178 & 131\,/\,81\,/\,107 & 141\,/\,104\,/\,90 & 122\,/\,156\,/\,103 & 93\,/\,122\,/\,144 & 131\,/\,163\,/\,90 \\
\bottomrule
\end{tabular}
}
\caption{Mean shared core / task-only / partner-only sizes over attention heads and MLP blocks under RelP, at $P$=50\%, averaged over the five partner tasks.}
\label{tab:decomposition_all_relp_head_mlp}
\end{table*}

\begin{table*}[htbp]
\centering
\small
\setlength{\tabcolsep}{3pt}
\resizebox{\textwidth}{!}{%
\begin{tabular}{clcccccc}
\toprule
$K$ & Model & Addition & ARC (Chal.) & ARC (Easy) & Boolean & IOI & CopyColors MCQA \\
\midrule
1\% & \texttt{Gemma 2 2B} & 1\,/\,347\,/\,285 & 14\,/\,23\,/\,334 & 20\,/\,53\,/\,321 & 3\,/\,329\,/\,286 & 4\,/\,222\,/\,307 & 23\,/\,740\,/\,180 \\
 & \texttt{Gemma 2 2B Instruct} & 1\,/\,405\,/\,228 & 15\,/\,25\,/\,287 & 19\,/\,48\,/\,278 & 3\,/\,262\,/\,255 & 6\,/\,248\,/\,254 & 21\,/\,499\,/\,186 \\
 & \texttt{Llama-3.2-3B} & 3\,/\,301\,/\,208 & 10\,/\,18\,/\,256 & 15\,/\,41\,/\,246 & 4\,/\,291\,/\,209 & 2\,/\,128\,/\,244 & 16\,/\,531\,/\,146 \\
 & \texttt{Llama-3.2-3B Instruct} & 1\,/\,305\,/\,183 & 7\,/\,13\,/\,235 & 11\,/\,29\,/\,227 & 3\,/\,112\,/\,220 & 3\,/\,195\,/\,203 & 11\,/\,539\,/\,124 \\
 & \texttt{Qwen3-4B} & 1\,/\,363\,/\,201 & 17\,/\,30\,/\,248 & 20\,/\,48\,/\,241 & 7\,/\,417\,/\,183 & 6\,/\,134\,/\,241 & 20\,/\,310\,/\,189 \\
\midrule
5\% & \texttt{Gemma 2 2B} & 8\,/\,618\,/\,1098 & 35\,/\,57\,/\,1178 & 56\,/\,164\,/\,1132 & 23\,/\,1105\,/\,983 & 29\,/\,831\,/\,1031 & 86\,/\,3145\,/\,500 \\
 & \texttt{Gemma 2 2B Instruct} & 9\,/\,751\,/\,893 & 33\,/\,54\,/\,1004 & 53\,/\,173\,/\,956 & 19\,/\,799\,/\,871 & 24\,/\,940\,/\,837 & 70\,/\,2345\,/\,501 \\
 & \texttt{Llama-3.2-3B} & 22\,/\,603\,/\,953 & 32\,/\,49\,/\,1053 & 48\,/\,123\,/\,1019 & 37\,/\,1044\,/\,848 & 27\,/\,643\,/\,939 & 75\,/\,2801\,/\,450 \\
 & \texttt{Llama-3.2-3B Instruct} & 15\,/\,597\,/\,810 & 27\,/\,48\,/\,905 & 39\,/\,103\,/\,879 & 18\,/\,332\,/\,859 & 23\,/\,943\,/\,730 & 57\,/\,2532\,/\,372 \\
 & \texttt{Qwen3-4B} & 9\,/\,836\,/\,755 & 46\,/\,81\,/\,862 & 56\,/\,153\,/\,836 & 25\,/\,1191\,/\,665 & 19\,/\,452\,/\,820 & 62\,/\,1735\,/\,512 \\
\midrule
10\% & \texttt{Gemma 2 2B} & 25\,/\,826\,/\,2413 & 63\,/\,110\,/\,2511 & 112\,/\,348\,/\,2404 & 77\,/\,2785\,/\,1959 & 76\,/\,1748\,/\,2168 & 200\,/\,6674\,/\,1034 \\
 & \texttt{Gemma 2 2B Instruct} & 29\,/\,1065\,/\,2015 & 74\,/\,125\,/\,2149 & 116\,/\,359\,/\,2051 & 58\,/\,1938\,/\,1805 & 71\,/\,2085\,/\,1761 & 174\,/\,5218\,/\,1010 \\
 & \texttt{Llama-3.2-3B} & 51\,/\,865\,/\,2277 & 74\,/\,126\,/\,2397 & 105\,/\,279\,/\,2329 & 106\,/\,2628\,/\,1858 & 89\,/\,1464\,/\,2111 & 200\,/\,6567\,/\,958 \\
 & \texttt{Llama-3.2-3B Instruct} & 37\,/\,858\,/\,1835 & 57\,/\,103\,/\,1962 & 77\,/\,207\,/\,1918 & 52\,/\,712\,/\,1846 & 81\,/\,2099\,/\,1534 & 146\,/\,5826\,/\,711 \\
 & \texttt{Qwen3-4B} & 26\,/\,1308\,/\,1816 & 92\,/\,184\,/\,1963 & 114\,/\,349\,/\,1903 & 64\,/\,2611\,/\,1510 & 50\,/\,1054\,/\,1838 & 144\,/\,4551\,/\,1027 \\
\midrule
20\% & \texttt{Gemma 2 2B} & 99\,/\,1409\,/\,6911 & 218\,/\,403\,/\,6969 & 368\,/\,1091\,/\,6651 & 374\,/\,9836\,/\,4895 & 346\,/\,4711\,/\,5953 & 747\,/\,16953\,/\,3024 \\
 & \texttt{Gemma 2 2B Instruct} & 117\,/\,1810\,/\,5986 & 260\,/\,475\,/\,6081 & 393\,/\,1106\,/\,5795 & 294\,/\,7275\,/\,4680 & 344\,/\,5854\,/\,4904 & 688\,/\,13823\,/\,2898 \\
 & \texttt{Llama-3.2-3B} & 164\,/\,1463\,/\,7503 & 391\,/\,801\,/\,7362 & 540\,/\,1533\,/\,7037 & 533\,/\,9951\,/\,5362 & 429\,/\,4220\,/\,6633 & 989\,/\,18945\,/\,3016 \\
 & \texttt{Llama-3.2-3B Instruct} & 113\,/\,1476\,/\,5192 & 221\,/\,441\,/\,5269 & 302\,/\,812\,/\,5098 & 223\,/\,2724\,/\,4810 & 383\,/\,5826\,/\,3998 & 622\,/\,14971\,/\,1882 \\
 & \texttt{Qwen3-4B} & 134\,/\,2543\,/\,6989 & 405\,/\,898\,/\,6993 & 513\,/\,1648\,/\,6714 & 384\,/\,10951\,/\,5008 & 262\,/\,4188\,/\,6507 & 706\,/\,15661\,/\,3679 \\
\midrule
30\% & \texttt{Gemma 2 2B} & 409\,/\,2984\,/\,16349 & 961\,/\,2033\,/\,15877 & 1381\,/\,3655\,/\,15048 & 1748\,/\,25108\,/\,10318 & 1382\,/\,10654\,/\,13647 & 2768\,/\,34100\,/\,7295 \\
 & \texttt{Gemma 2 2B Instruct} & 494\,/\,4289\,/\,14952 & 1103\,/\,2153\,/\,14649 & 1511\,/\,3780\,/\,13833 & 1458\,/\,20430\,/\,10567 & 1430\,/\,12962\,/\,12094 & 2635\,/\,29768\,/\,7287 \\
 & \texttt{Llama-3.2-3B} & 529\,/\,2517\,/\,19590 & 2411\,/\,6785\,/\,16478 & 2737\,/\,7545\,/\,15935 & 2602\,/\,26052\,/\,12396 & 1686\,/\,9211\,/\,16863 & 4197\,/\,37369\,/\,8218 \\
 & \texttt{Llama-3.2-3B Instruct} & 351\,/\,2516\,/\,12869 & 952\,/\,2163\,/\,12218 & 1228\,/\,3339\,/\,11652 & 1070\,/\,10134\,/\,10482 & 1392\,/\,12205\,/\,9682 & 2258\,/\,31359\,/\,4812 \\
 & \texttt{Qwen3-4B} & 767\,/\,6082\,/\,22044 & 2160\,/\,5439\,/\,20500 & 2497\,/\,6962\,/\,19792 & 2569\,/\,35211\,/\,14056 & 1659\,/\,13735\,/\,19443 & 3696\,/\,40125\,/\,11721 \\
\bottomrule
\end{tabular}
}
\caption{Mean shared core / task-only / partner-only sizes over MLP neurons under EAP-IG, at $P$=50\%, averaged over the five partner tasks.}
\label{tab:decomposition_all_eap_ig_neuron}
\end{table*}

\begin{table*}[htbp]
\centering
\small
\setlength{\tabcolsep}{3pt}
\resizebox{\textwidth}{!}{%
\begin{tabular}{clcccccc}
\toprule
$K$ & Model & Addition & ARC (Chal.) & ARC (Easy) & Boolean & IOI & CopyColors MCQA \\
\midrule
1\% & \texttt{Gemma 2 2B} & 3\,/\,321\,/\,198 & 17\,/\,29\,/\,240 & 22\,/\,58\,/\,227 & 4\,/\,514\,/\,158 & 2\,/\,82\,/\,247 & 24\,/\,252\,/\,187 \\
 & \texttt{Gemma 2 2B Instruct} & 4\,/\,364\,/\,199 & 27\,/\,50\,/\,235 & 33\,/\,82\,/\,221 & 3\,/\,393\,/\,195 & 0\,/\,67\,/\,263 & 32\,/\,331\,/\,172 \\
 & \texttt{Llama-3.2-3B} & 2\,/\,212\,/\,238 & 28\,/\,51\,/\,240 & 32\,/\,71\,/\,231 & 5\,/\,574\,/\,163 & 2\,/\,38\,/\,273 & 34\,/\,369\,/\,169 \\
 & \texttt{Llama-3.2-3B Instruct} & 0\,/\,190\,/\,141 & 18\,/\,35\,/\,151 & 22\,/\,63\,/\,141 & 0\,/\,180\,/\,143 & 1\,/\,74\,/\,163 & 21\,/\,293\,/\,96 \\
 & \texttt{Qwen3-4B} & 3\,/\,595\,/\,456 & 56\,/\,119\,/\,488 & 62\,/\,152\,/\,475 & 10\,/\,1243\,/\,318 & 8\,/\,81\,/\,553 & 54\,/\,512\,/\,412 \\
\midrule
5\% & \texttt{Gemma 2 2B} & 17\,/\,579\,/\,906 & 42\,/\,79\,/\,976 & 56\,/\,135\,/\,949 & 25\,/\,2221\,/\,568 & 12\,/\,352\,/\,957 & 70\,/\,1625\,/\,633 \\
 & \texttt{Gemma 2 2B Instruct} & 22\,/\,651\,/\,1074 & 99\,/\,185\,/\,1075 & 148\,/\,415\,/\,970 & 28\,/\,1862\,/\,825 & 11\,/\,349\,/\,1148 & 155\,/\,2228\,/\,599 \\
 & \texttt{Llama-3.2-3B} & 15\,/\,443\,/\,1142 & 80\,/\,138\,/\,1125 & 114\,/\,288\,/\,1055 & 38\,/\,2475\,/\,708 & 16\,/\,263\,/\,1177 & 130\,/\,2243\,/\,644 \\
 & \texttt{Llama-3.2-3B Instruct} & 7\,/\,433\,/\,743 & 63\,/\,120\,/\,738 & 81\,/\,238\,/\,693 & 11\,/\,620\,/\,700 & 14\,/\,326\,/\,755 & 81\,/\,2193\,/\,301 \\
 & \texttt{Qwen3-4B} & 44\,/\,1106\,/\,2381 & 243\,/\,476\,/\,2268 & 290\,/\,755\,/\,2156 & 104\,/\,5628\,/\,1404 & 41\,/\,396\,/\,2527 & 297\,/\,3894\,/\,1519 \\
\midrule
10\% & \texttt{Gemma 2 2B} & 43\,/\,894\,/\,2372 & 83\,/\,156\,/\,2473 & 131\,/\,391\,/\,2368 & 94\,/\,5229\,/\,1445 & 43\,/\,867\,/\,2378 & 201\,/\,4883\,/\,1386 \\
 & \texttt{Gemma 2 2B Instruct} & 60\,/\,1028\,/\,3004 & 290\,/\,508\,/\,2832 & 456\,/\,1279\,/\,2479 & 123\,/\,4610\,/\,2212 & 46\,/\,936\,/\,3039 & 506\,/\,6566\,/\,1361 \\
 & \texttt{Llama-3.2-3B} & 45\,/\,653\,/\,2927 & 186\,/\,299\,/\,2828 & 286\,/\,733\,/\,2622 & 137\,/\,5837\,/\,1779 & 56\,/\,727\,/\,2899 & 357\,/\,6241\,/\,1434 \\
 & \texttt{Llama-3.2-3B Instruct} & 27\,/\,689\,/\,1980 & 148\,/\,286\,/\,1916 & 217\,/\,685\,/\,1753 & 46\,/\,1499\,/\,1796 & 44\,/\,856\,/\,1927 & 234\,/\,6023\,/\,665 \\
 & \texttt{Qwen3-4B} & 119\,/\,1533\,/\,5769 & 661\,/\,1241\,/\,5176 & 812\,/\,2049\,/\,4834 & 378\,/\,11470\,/\,3470 & 90\,/\,924\,/\,5925 & 910\,/\,10904\,/\,2946 \\
\midrule
20\% & \texttt{Gemma 2 2B} & 195\,/\,2083\,/\,7441 & 319\,/\,616\,/\,7587 & 610\,/\,1875\,/\,6985 & 585\,/\,14099\,/\,4570 & 228\,/\,2779\,/\,7263 & 1041\,/\,16031\,/\,3637 \\
 & \texttt{Gemma 2 2B Instruct} & 364\,/\,3220\,/\,9700 & 1393\,/\,2373\,/\,8635 & 2026\,/\,4945\,/\,7360 & 845\,/\,12594\,/\,7249 & 264\,/\,3053\,/\,9853 & 2378\,/\,20450\,/\,3837 \\
 & \texttt{Llama-3.2-3B} & 180\,/\,1333\,/\,8910 & 763\,/\,1213\,/\,8235 & 1171\,/\,2811\,/\,7426 & 725\,/\,15531\,/\,5417 & 265\,/\,2427\,/\,8590 & 1583\,/\,18963\,/\,3701 \\
 & \texttt{Llama-3.2-3B Instruct} & 146\,/\,1476\,/\,6749 & 639\,/\,1203\,/\,6212 & 926\,/\,2770\,/\,5554 & 285\,/\,5173\,/\,5843 & 209\,/\,2901\,/\,6388 & 1099\,/\,19270\,/\,2047 \\
 & \texttt{Qwen3-4B} & 526\,/\,3235\,/\,16249 & 2739\,/\,4593\,/\,13322 & 3348\,/\,7390\,/\,12031 & 1985\,/\,24765\,/\,10193 & 385\,/\,2885\,/\,16489 & 3944\,/\,31842\,/\,6426 \\
\midrule
30\% & \texttt{Gemma 2 2B} & 1005\,/\,6381\,/\,16477 & 1588\,/\,2725\,/\,16509 & 2545\,/\,6283\,/\,14649 & 2385\,/\,27678\,/\,10561 & 1072\,/\,7664\,/\,16140 & 3918\,/\,31553\,/\,7947 \\
 & \texttt{Gemma 2 2B Instruct} & 1866\,/\,8837\,/\,20608 & 4774\,/\,7394\,/\,17407 & 6067\,/\,11556\,/\,15022 & 3383\,/\,25327\,/\,15489 & 1375\,/\,8175\,/\,21329 & 7217\,/\,37100\,/\,8534 \\
 & \texttt{Llama-3.2-3B} & 742\,/\,3617\,/\,19538 & 2840\,/\,4385\,/\,16867 & 3927\,/\,8330\,/\,14774 & 2832\,/\,29932\,/\,11768 & 1090\,/\,6658\,/\,18512 & 5246\,/\,36161\,/\,7625 \\
 & \texttt{Llama-3.2-3B Instruct} & 639\,/\,3992\,/\,16163 & 2335\,/\,4241\,/\,14078 & 3150\,/\,8042\,/\,12341 & 1424\,/\,13800\,/\,13260 & 961\,/\,8171\,/\,14941 & 3904\,/\,37984\,/\,5447 \\
 & \texttt{Qwen3-4B} & 2699\,/\,9784\,/\,33677 & 8828\,/\,13432\,/\,25593 & 10171\,/\,18371\,/\,22993 & 6970\,/\,44046\,/\,21700 & 1912\,/\,9238\,/\,34731 & 11740\,/\,57173\,/\,13350 \\
\bottomrule
\end{tabular}
}
\caption{Mean shared core / task-only / partner-only sizes over MLP neurons under RelP, at $P$=50\%, averaged over the five partner tasks.}
\label{tab:decomposition_all_relp_neuron}
\end{table*}

\begin{figure*}[htbp]
  \centering
  \includegraphics[width=\textwidth]{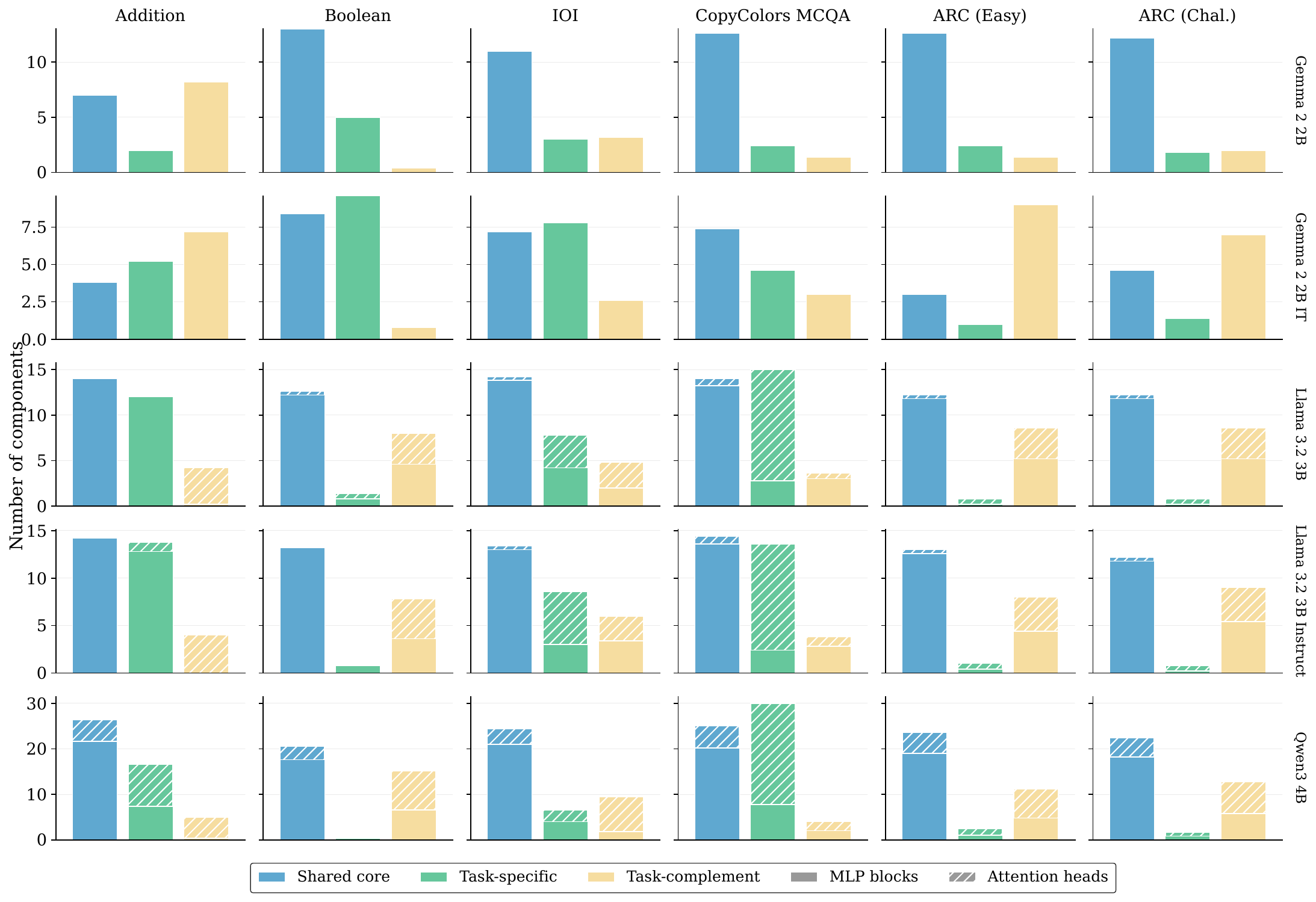}
  \caption{\textbf{MLP vs.\ attention head composition of circuit decompositions at $K$=10\%.} Each group of three bars shows the mean number of MLP blocks (solid) and attention heads (hatched) in the shared core, task-specific, and task-complement sets. MLP blocks account for the vast majority of the shared core across all models and tasks, while attention heads appear primarily in the task-specific and task-complement sets at larger circuit sizes.}
  \label{fig:decomposition_mlp_head}
\end{figure*}

\clearpage

\subsection{Layer Distribution}
\label{sec:layer_distribution_cdf}

\begin{figure*}[htbp]
  \centering
  \includegraphics[width=\textwidth]{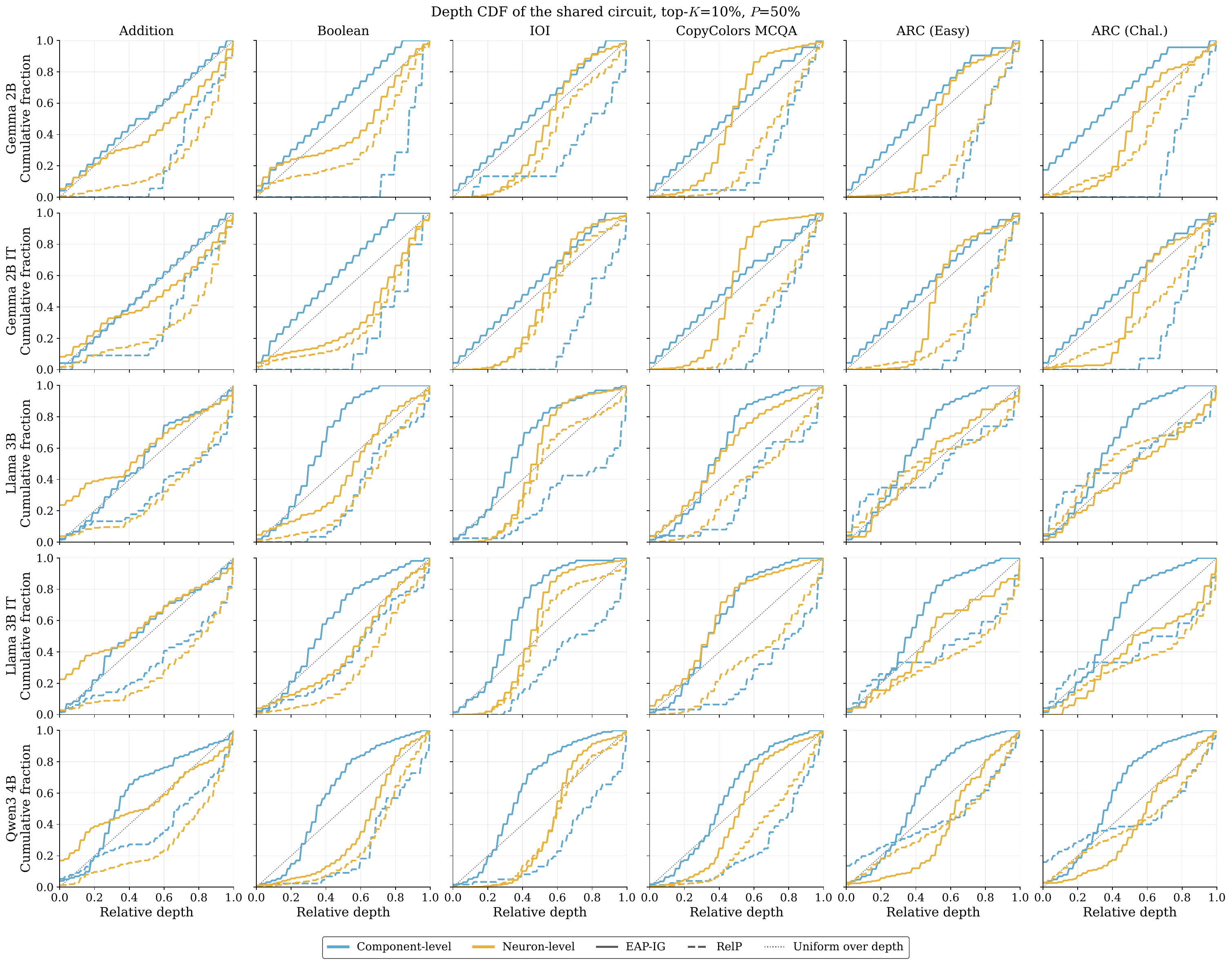}
  \caption{\textbf{Depth CDF of the task circuit for every model and task, at $K$=10\% and $P$=50\%.} Each panel gives the cumulative fraction of the task circuit at or below a given relative depth, for all four combinations of attribution method and granularity. The dotted diagonal is a circuit spread uniformly over depth.}
  \label{fig:depth_cdf_grid}
\end{figure*}

\Cref{fig:depth_cdf_grid} expands \cref{fig:circuit_anatomy} into thirty model--task cells, shown for all four combinations of attribution method and granularity.
The ordering described in the main text appears in almost every cell: component-level EAP-IG is generally earliest, while component-level RelP is latest.
For EAP-IG, the neuron-level circuit is earlier than the component-level circuit in only three of the thirty cells, including \texttt{Addition} for both Llama models.
The Gemma models show the least variation across depth. Their component-level EAP-IG circuits differ from the uniform diagonal by a mean of 0.15 and 0.11 across tasks, compared with 0.28--0.31 for the Llama and Qwen models.
These Gemma circuits consist of 93.5\% and 98.5\% MLP blocks, respectively. \texttt{Gemma~2~2B} contains only 26 such blocks, so a $K=10\%$ circuit selects roughly one MLP block per unit of depth across tasks.

\begin{figure*}[htbp]
  \centering

  \begin{minipage}[t]{0.48\textwidth}
    \centering
    \includegraphics[width=\textwidth]{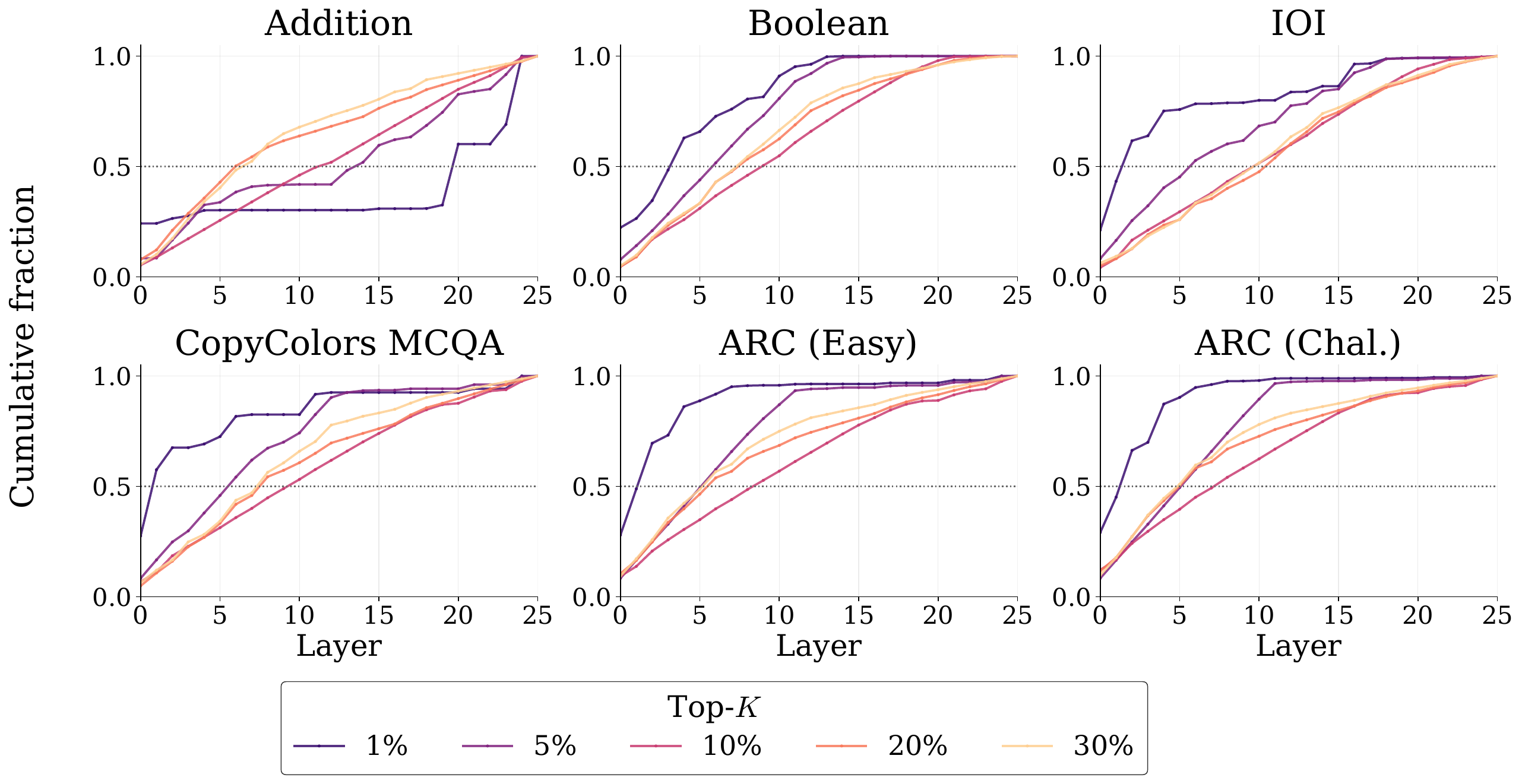}

    \small \texttt{Gemma-2-2B}
  \end{minipage}
  \hfill
  \begin{minipage}[t]{0.48\textwidth}
    \centering
    \includegraphics[width=\textwidth]{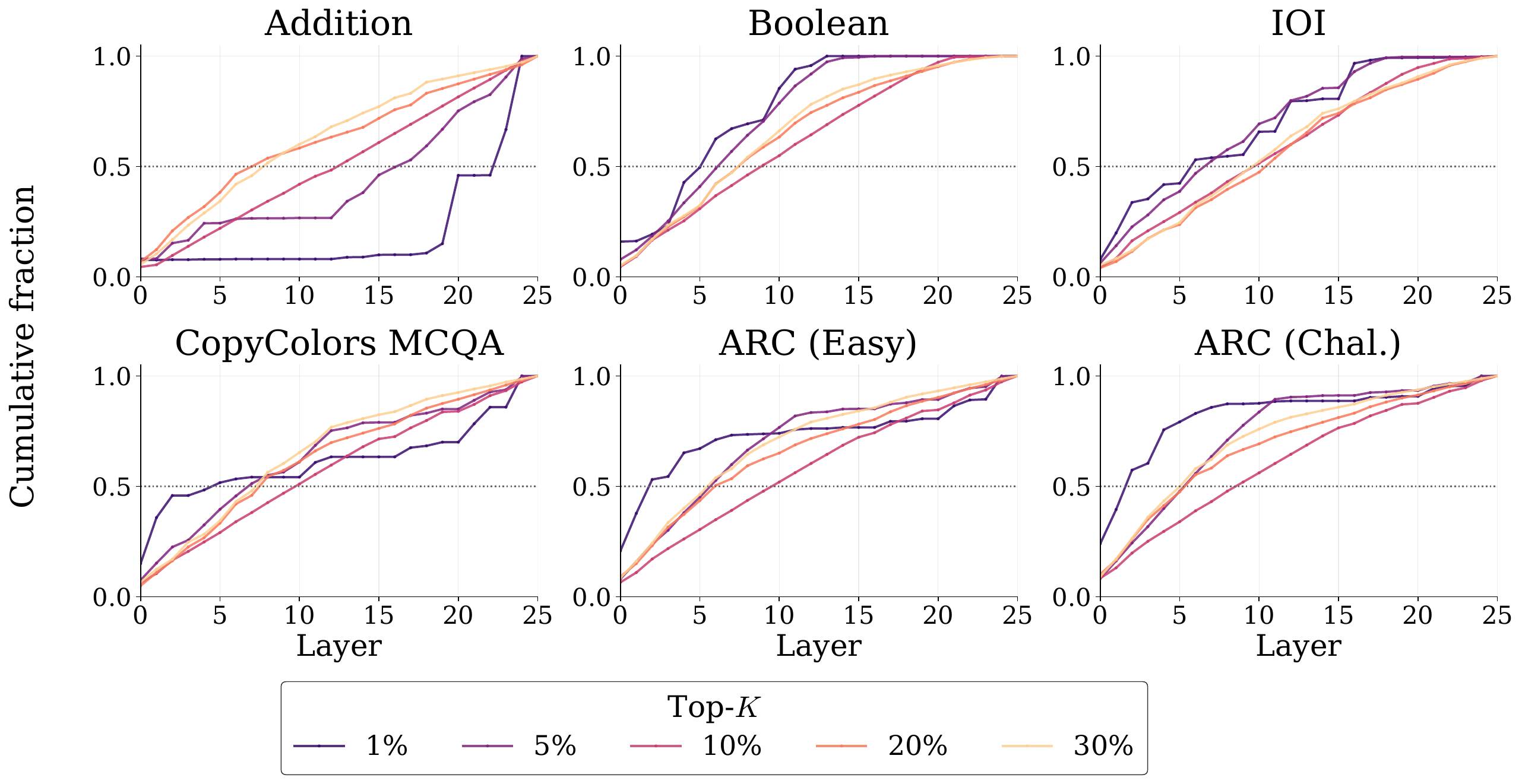}

    \small \texttt{Gemma-2-2B-IT}
  \end{minipage}

  \vspace{1em}

  \begin{minipage}[t]{0.48\textwidth}
    \centering
    \includegraphics[width=\textwidth]{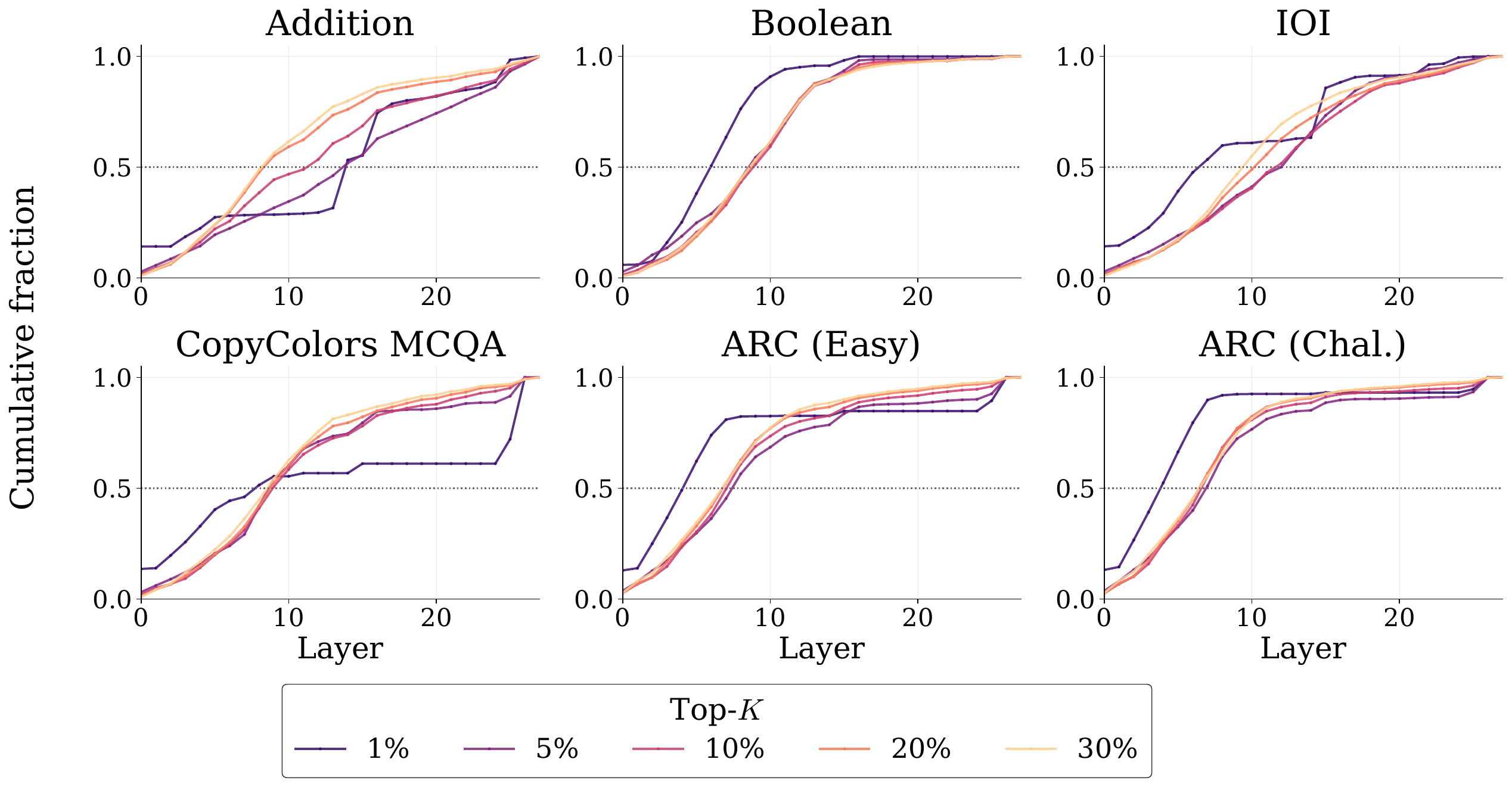}

    \small \texttt{Llama-3.2-3B}
  \end{minipage}
  \hfill
  \begin{minipage}[t]{0.48\textwidth}
    \centering
    \includegraphics[width=\textwidth]{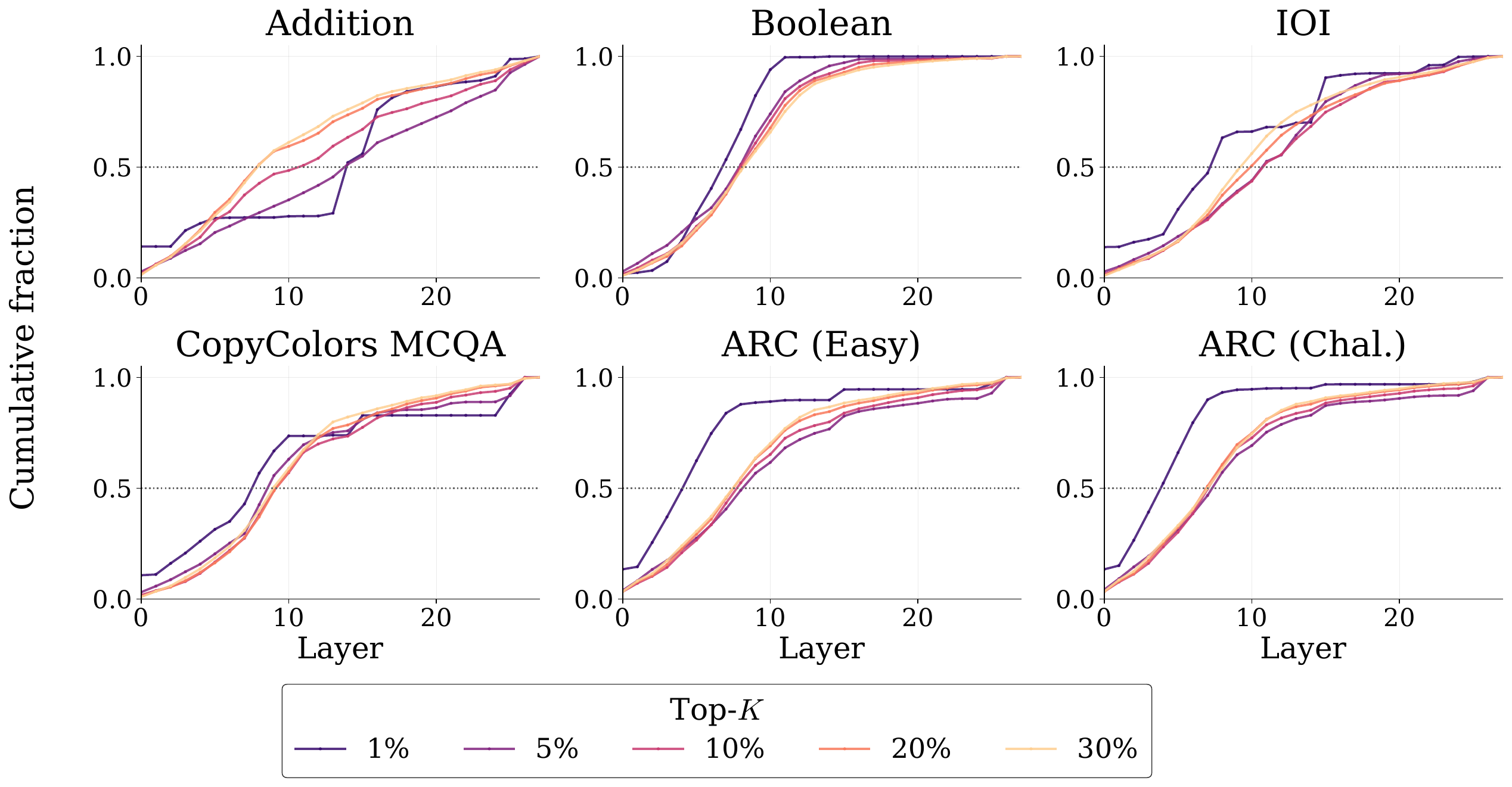}

    \small \texttt{Llama-3.2-3B-Instruct}
  \end{minipage}

  \vspace{1em}

  \centering
  \begin{minipage}[t]{0.48\textwidth}
    \centering
    \includegraphics[width=\textwidth]{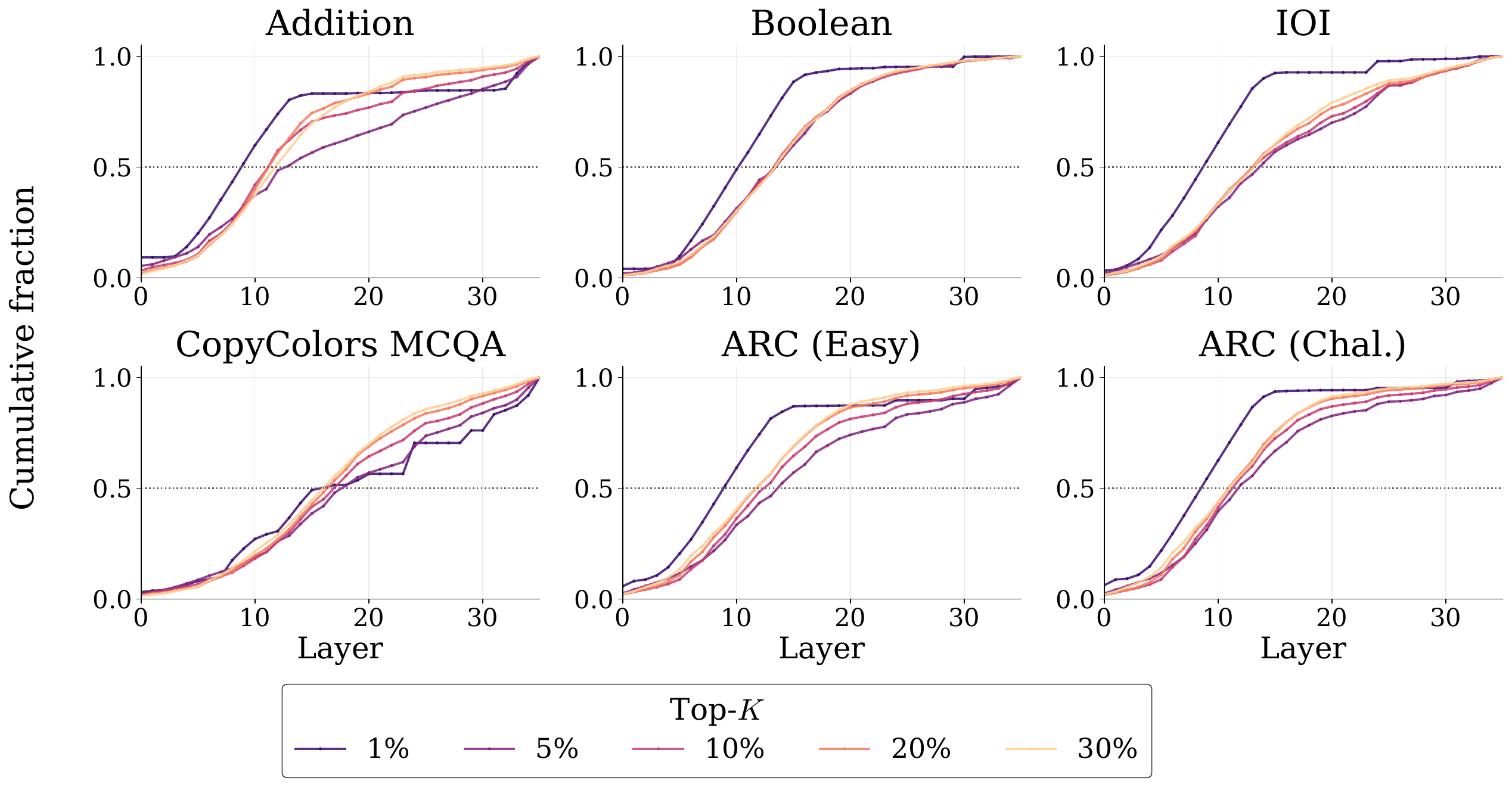}

    \small \texttt{Qwen3-4B}
  \end{minipage}

  \caption{\textbf{Cumulative layer distribution of circuit components.} Each line shows the cumulative fraction of components in the top-$K$\% circuit at or below a given layer. At small $K$, the CDF is shifted left (toward earlier layers) in every model, except for \texttt{Addition} in the \texttt{Gemma-2} models.}
  \label{fig:layer_cdf_all}
\end{figure*}

\Cref{fig:layer_cdf_all} shows the same quantity as a function of circuit size top-$K$ rather than of method and granularity, under EAP-IG at the component granularity.
In every model, the CDF at $K$=1\% is shifted toward earlier layers than at larger $K$, indicating that the highest-attribution components are concentrated in early-to-middle layers.
The exception is \texttt{Addition} in the two \texttt{Gemma-2} models, whose $K$=1\% circuit lies mostly in the last few layers.
At larger $K$, the distribution becomes more uniform across layers in all models.

\section{Selective Ablation Across K}
\label{sec:selective_all_k}

\begin{figure*}[htbp]
  \centering

  \begin{minipage}{0.9\textwidth}
    \centering
    \textbf{$K=1\%$}\\[2pt]
    \includegraphics[width=\textwidth]{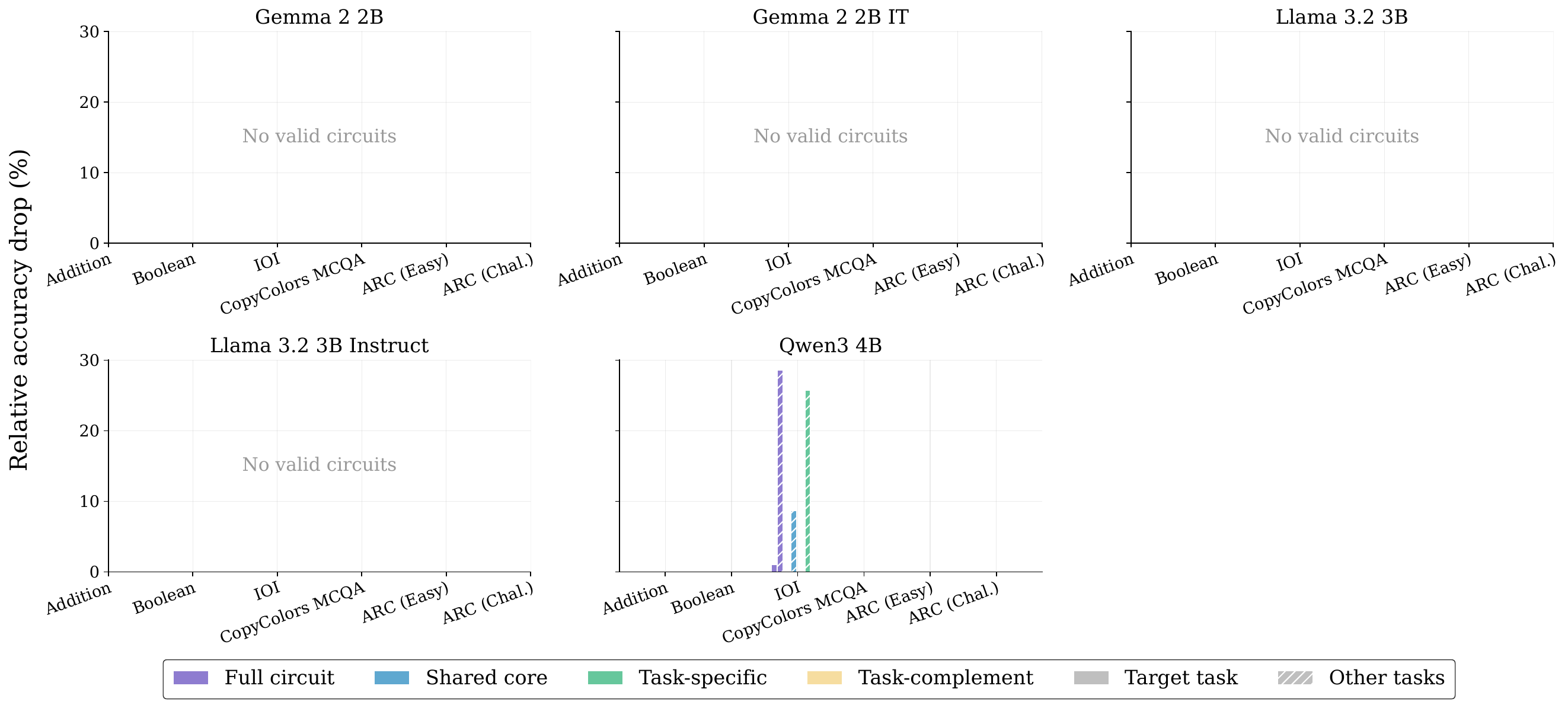}
  \end{minipage}\\[6pt]

  \begin{minipage}{0.9\textwidth}
    \centering
    \textbf{$K=5\%$}\\[2pt]
    \includegraphics[width=\textwidth]{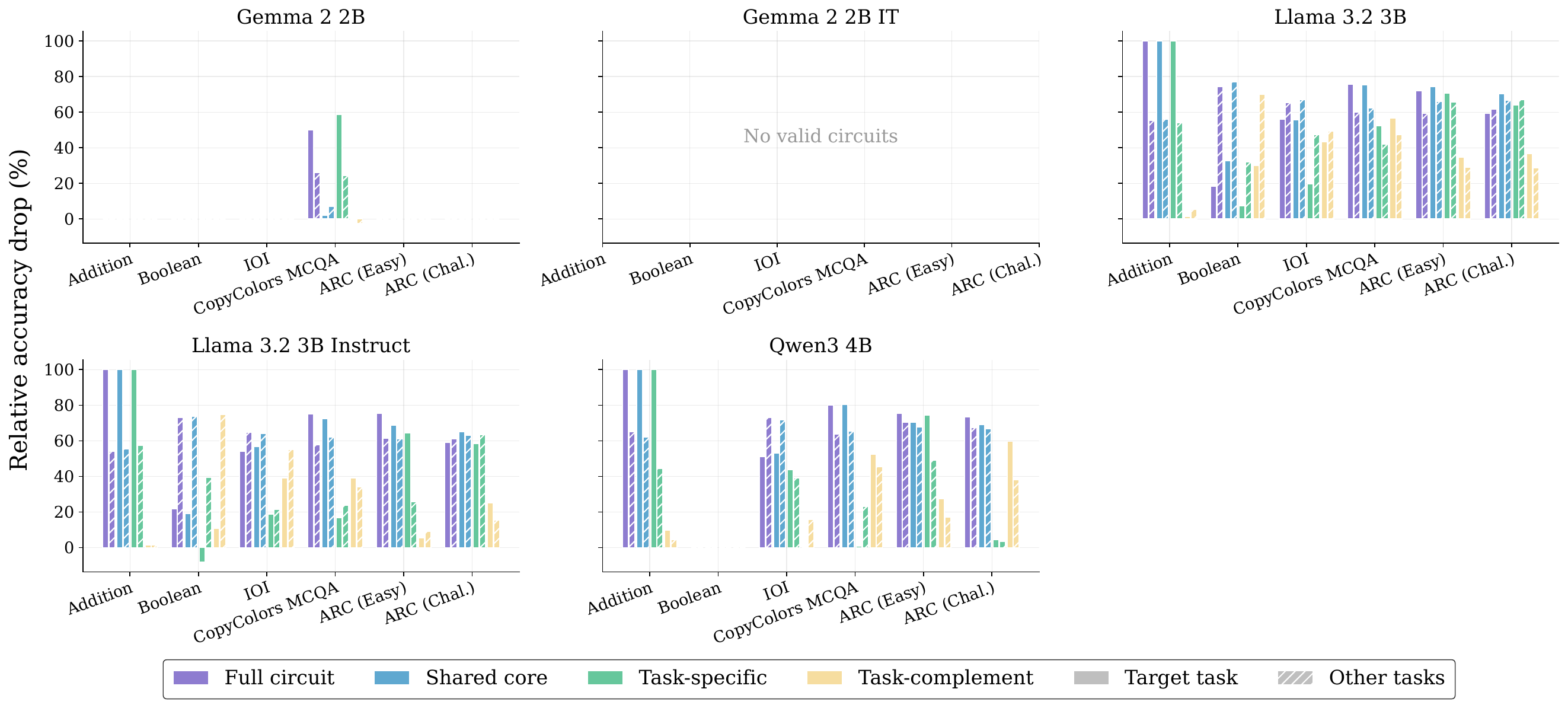}
  \end{minipage}\\[6pt]

  \begin{minipage}{0.9\textwidth}
    \centering
    \textbf{$K=10\%$}\\[2pt]
    \includegraphics[width=\textwidth]{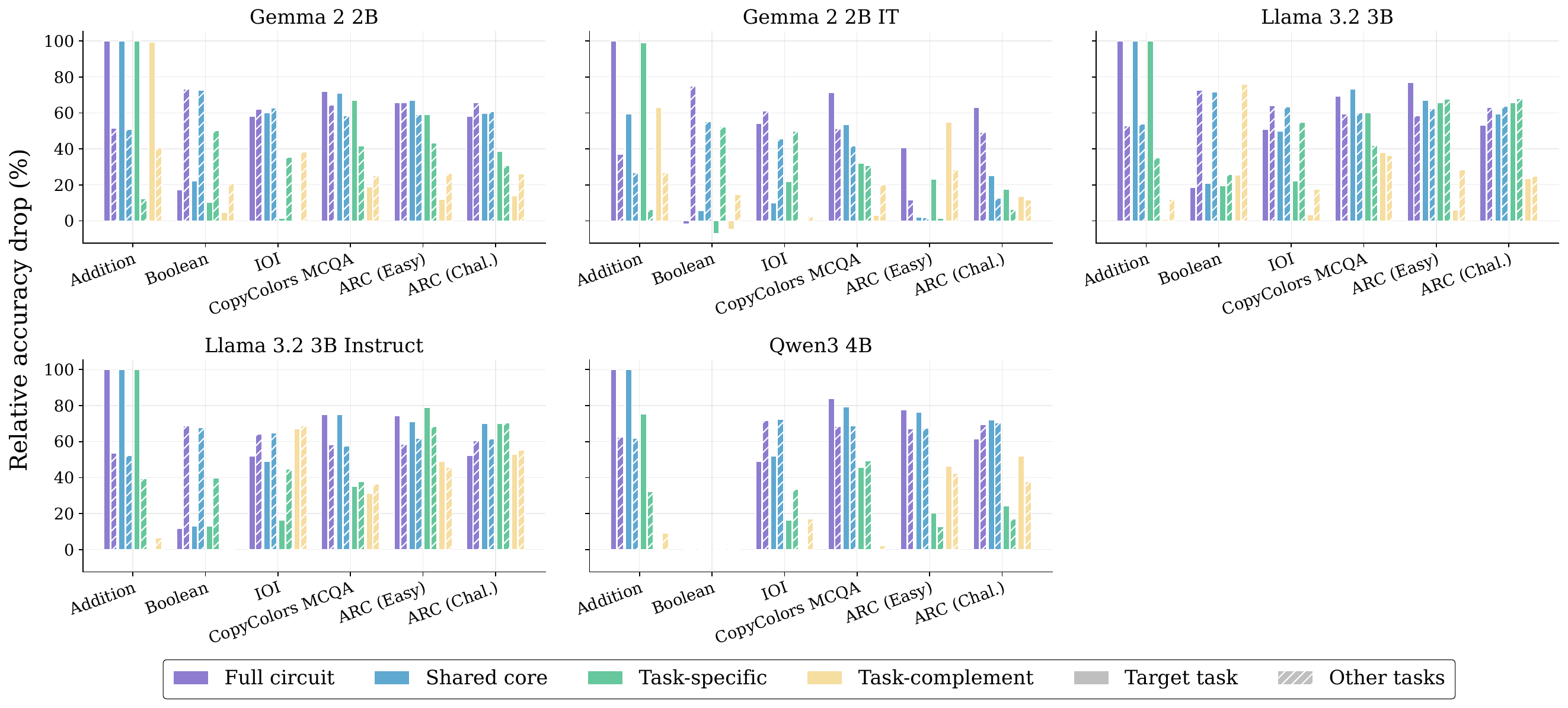}
  \end{minipage}

  \caption{Selective ablation: relative accuracy drop by component set for $K \in \{1, 5, 10\}\%$.}
  \label{fig:selective_1_5_10_k}
\end{figure*}

\begin{figure*}[htbp]
  \centering

  \begin{minipage}{0.9\textwidth}
    \centering
    \textbf{$K=20\%$}\\[2pt]
    \includegraphics[width=\textwidth]{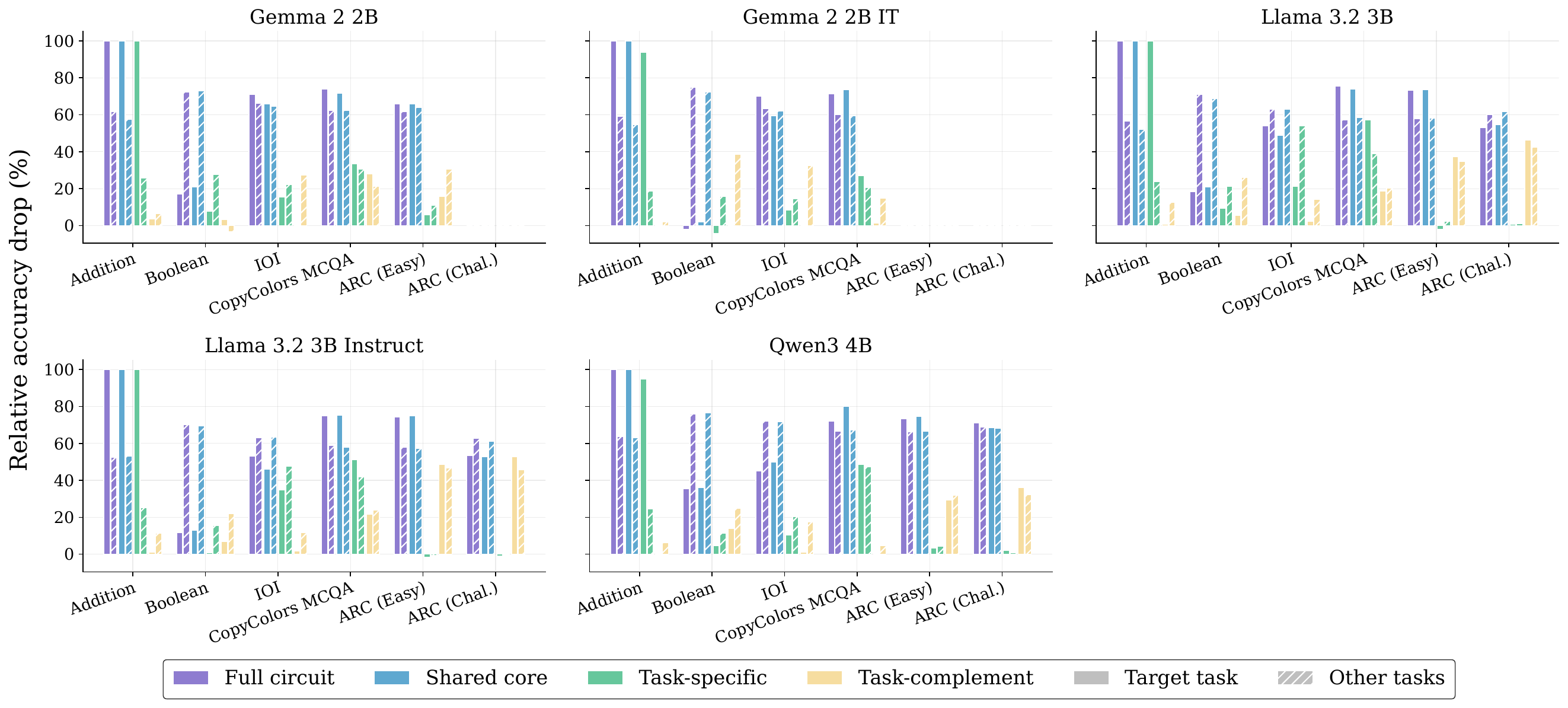}
  \end{minipage}\\[6pt]

  \begin{minipage}{0.9\textwidth}
    \centering
    \textbf{$K=30\%$}\\[2pt]
    \includegraphics[width=\textwidth]{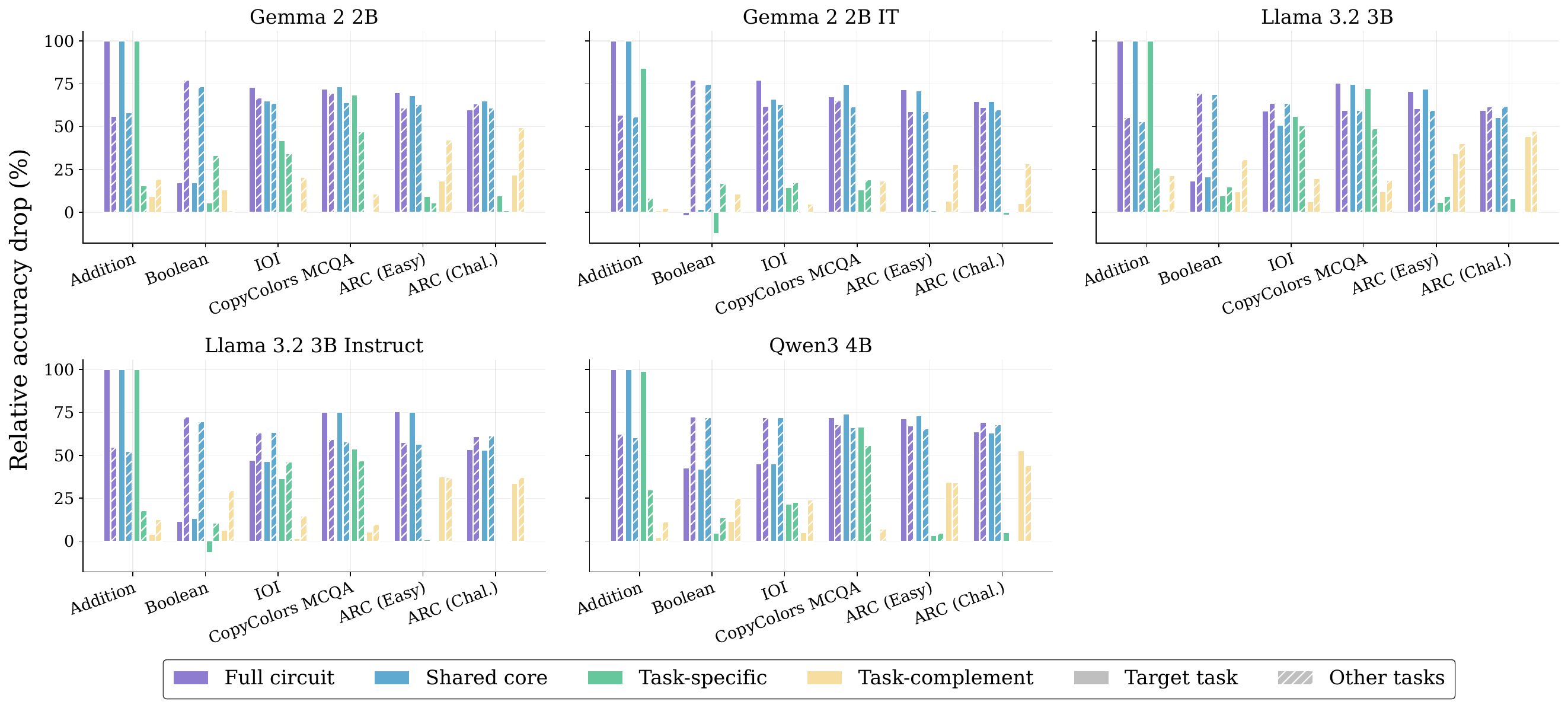}
  \end{minipage}\\[6pt]

  \caption{Selective ablation: relative accuracy drop by component set for $K \in \{20, 30\}\%$.}
  \label{fig:selective_20_30_k}
\end{figure*}

\Cref{fig:selective_1_5_10_k,fig:selective_20_30_k} show the selective ablation results for $K \in \{1, 5, 10, 20, 30\}\%$, complementing the $K$=10\% results in the main text.

\section{Case Study of Shared Components in \texttt{Llama-3.2-3B}}
\label{sec:case_study_details}

\begin{figure*}[htbp]
  \centering
  \includegraphics[width=\textwidth]{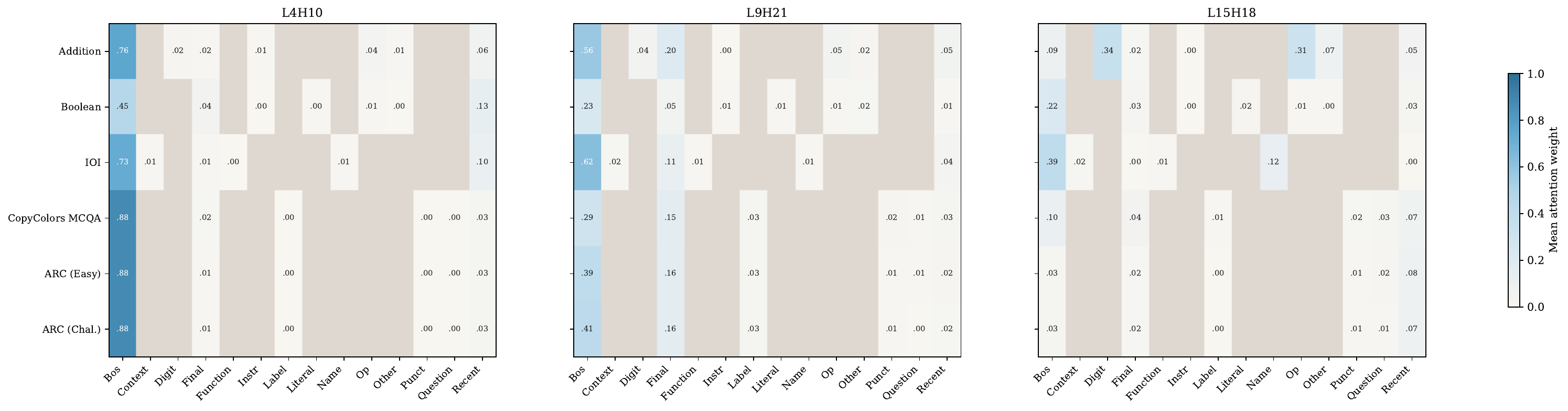}
  \caption{\textbf{Attention from the final query position for the three heads shared by all six tasks in \texttt{Llama-3.2-3B}.} Rows are tasks, columns are template roles, and each cell is the mean attention weight per token of that role, over 16 prompts. Grey cells mark roles that do not occur in that task. Because these are per-token means rather than totals, they are comparable down a column but should not be summed across a row.}
  \label{fig:case_study}
\end{figure*}

To characterize the three attention heads shared by all six tasks in \texttt{Llama-3.2-3B}, L4H10, L9H21 and L15H18, we ran 16 prompts from each task through the model and recorded the attention from the final query position, aggregating it by the template role of each key position (\cref{fig:case_study}).
Summing over every token of a role and averaging over tasks, the beginning-of-text token receives 0.762, 0.420 and 0.144 of the attention of L4H10, L9H21 and L15H18, against 0.096, 0.389 and 0.737 for the body of the prompt and 0.018, 0.137 and 0.020 for the final token.
L4H10 therefore behaves as an attention sink~\citep{xiao2024efficient}.
L15H18 sends more of its attention to the body of the prompt than 99.9\% of the heads in the model, and in \texttt{Addition}, \texttt{Boolean~Logic} and \texttt{IOI} a large share of it goes to the tokens that carry the task, with 34\% on the digits in \texttt{Addition}, 47\% on the literals in \texttt{Boolean~Logic} and 47\% on the names in \texttt{IOI}, each above the 97th percentile of all heads.
L9H21 divides its attention between the beginning-of-text token and the body of the prompt, and within the body it attends to the answer labels of \texttt{CopyColors~MCQA} and the two \texttt{ARC} tasks more than 89 to 92\% of heads do.
We restrict this analysis to the component granularity, since the neuron basis contains no attention, and to the EAP-IG circuits studied in \cref{sec:case_study}.

\clearpage

\section{Pretraining Dynamics}
\label{sec:pretraining_appendix}
\begin{figure*}[htbp]
  \centering
  \includegraphics[width=\textwidth]{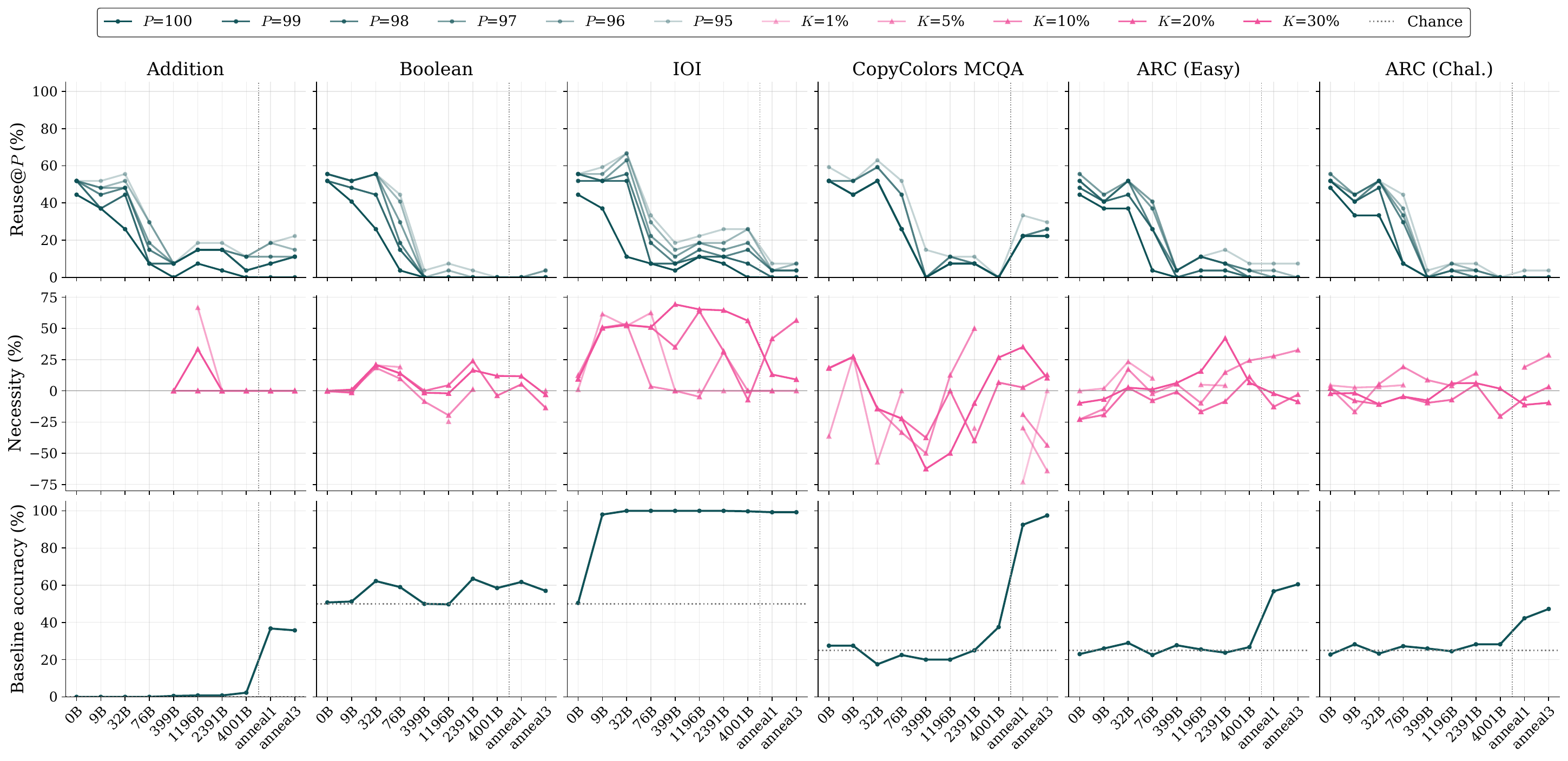}
  \caption{\textbf{Circuit reuse, \causalmetric~and baseline accuracy across \texttt{OLMo-2-1B} pretraining checkpoints.} \emph{Top:}~\reuseat{P} at top-$K$=10\% for $P$ from 95\% (lightest) to 100\% (darkest). \emph{Middle:}~\causalmetric~of the $P$=95\% task circuit for top-$K$ from 1\% (lightest) to 30\% (darkest), missing where the baseline accuracy is zero or the circuit is empty. \emph{Bottom:}~baseline accuracy, with chance accuracy as a grey dotted line. The vertical dotted line separates the stage-1 checkpoints from the two stage-2 anneals.}
  \label{fig:pretraining}
\end{figure*}

We track reuse and \causalmetric~across ten checkpoints of \texttt{OLMo-2-1B}'s $\sim$4T-token training run, eight from stage 1 and two stage-2 anneals (each $\sim$51B tokens of curated data with learning-rate decay on top of stage 1), using EAP-IG at the component granularity with 500 examples per task (\cref{fig:pretraining}).
\reuseat{95} at $K$=10\% is highest in the first 32B tokens (44--67\% across tasks) and then declines, with most of the drop between the 76B and 399B checkpoints.
For the rest of stage 1 it lies between 7 and 19\% on Addition, 0 and 15\% on ARC and CopyColors~MCQA, 19 and 26\% on IOI, and 0 and 7\% on Boolean, whose $K$=10\% task circuit is empty at 4001B.
The anneals raise reuse on CopyColors~MCQA (to 30--33\%) and lower it on IOI (to 7\%), the two tasks whose baseline accuracy changes most during training.

\Causalmetric~is hard to read for most of stage 1 because baseline accuracy on Addition, ARC and CopyColors~MCQA stays near chance, leaving little room for ablation to do measurable damage.
During the anneal phase, baselines on Addition (2$\rightarrow$37\%), ARC~(Easy) (27$\rightarrow$60\%) and CopyColors~MCQA (38$\rightarrow$98\%) jump sharply, likely because anneal mixtures emphasize downstream-task data.
With only 272 components, \texttt{OLMo-2-1B} is also sensitive to the random control, since removing a handful of random components already costs substantial accuracy.
On CopyColors~MCQA at the anneals, ablating the $K$=30\% task circuit reduces accuracy from 93--98\% to 28--30\%, against 38--63\% for a capacity-matched random ablation, while at $K$=10\% the control does as much damage as the circuit.
The clearest case is \texttt{IOI}, whose $K$=30\% circuit is necessary from 9B tokens to the end of stage 1 (\causalmetric~of 50 to 70\%) and much less so after the anneals (9 to 13\%).

\clearpage
\section{Full Within-Task Results}
\label{sec:full_within}

\begin{figure*}[htbp]
    \centering
    \includegraphics[width=\textwidth]{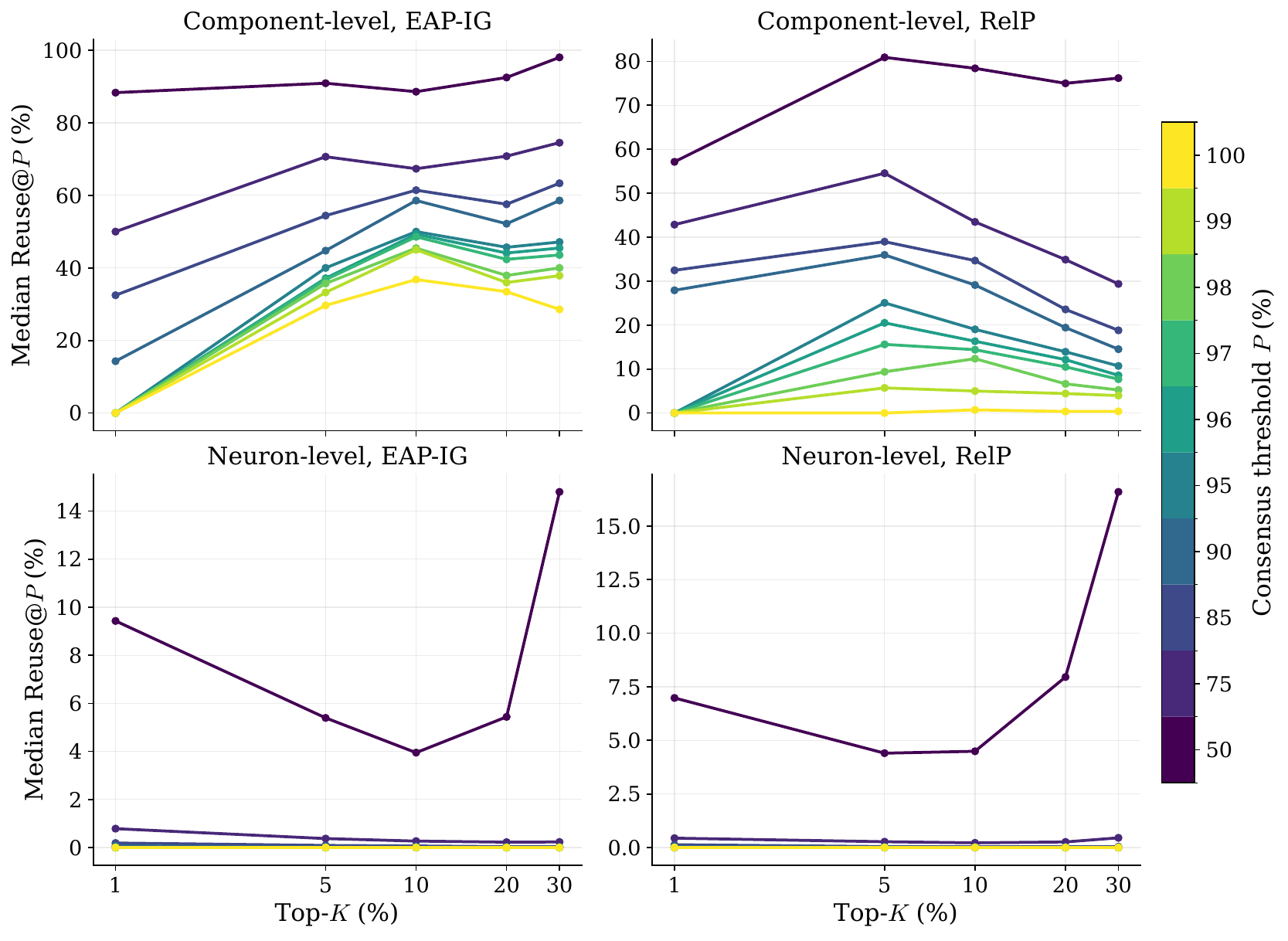}
    \caption{Median within-task~\reuseat{P} over models and tasks across top-$K$, one line per consensus threshold $P$, for component-level (top) and neuron-level (bottom) circuits under EAP-IG (left) and RelP (right).}
    \label{fig:reuse_sweep}
\end{figure*}

\begin{figure*}[htbp]
    \centering
    \includegraphics[width=\textwidth]{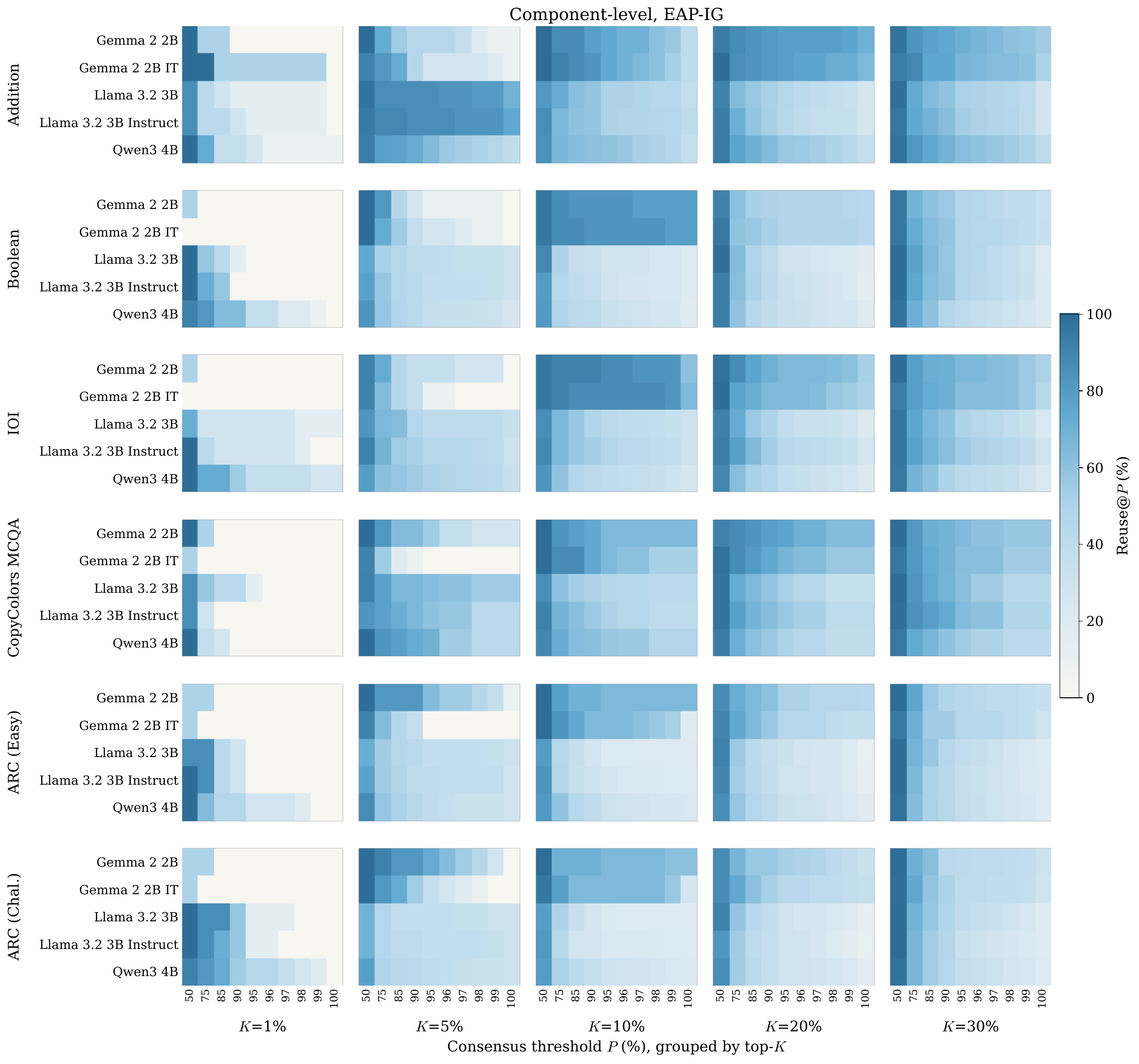}
    \caption{Within-task~\reuseat{P} of component-level circuits under EAP-IG over the full grid of top-$K$ and consensus threshold $P$. Columns are grouped by top-$K$, and rows are grouped by task, with one row per model. A value of 0 includes settings where the task circuit is empty.}
    \label{fig:reuse_all_eap_ig_head_mlp}
\end{figure*}

\begin{figure*}[htbp]
    \centering
    \includegraphics[width=\textwidth]{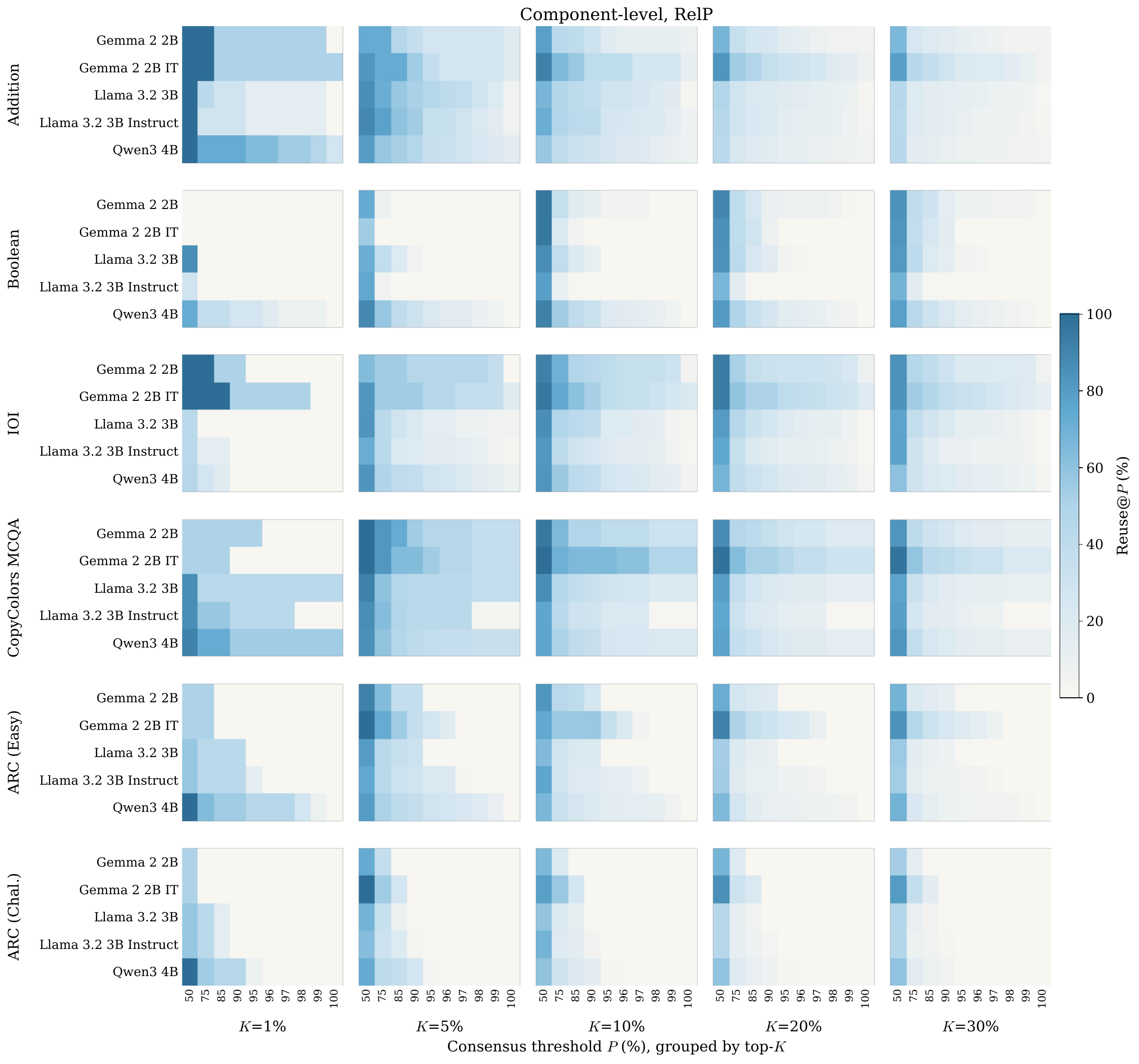}
    \caption{Within-task~\reuseat{P} of component-level circuits under RelP over the full grid of top-$K$ and consensus threshold $P$. Columns are grouped by top-$K$, and rows are grouped by task, with one row per model. A value of 0 includes settings where the task circuit is empty.}
    \label{fig:reuse_all_relp_head_mlp}
\end{figure*}

\begin{figure*}[htbp]
    \centering
    \includegraphics[width=\textwidth]{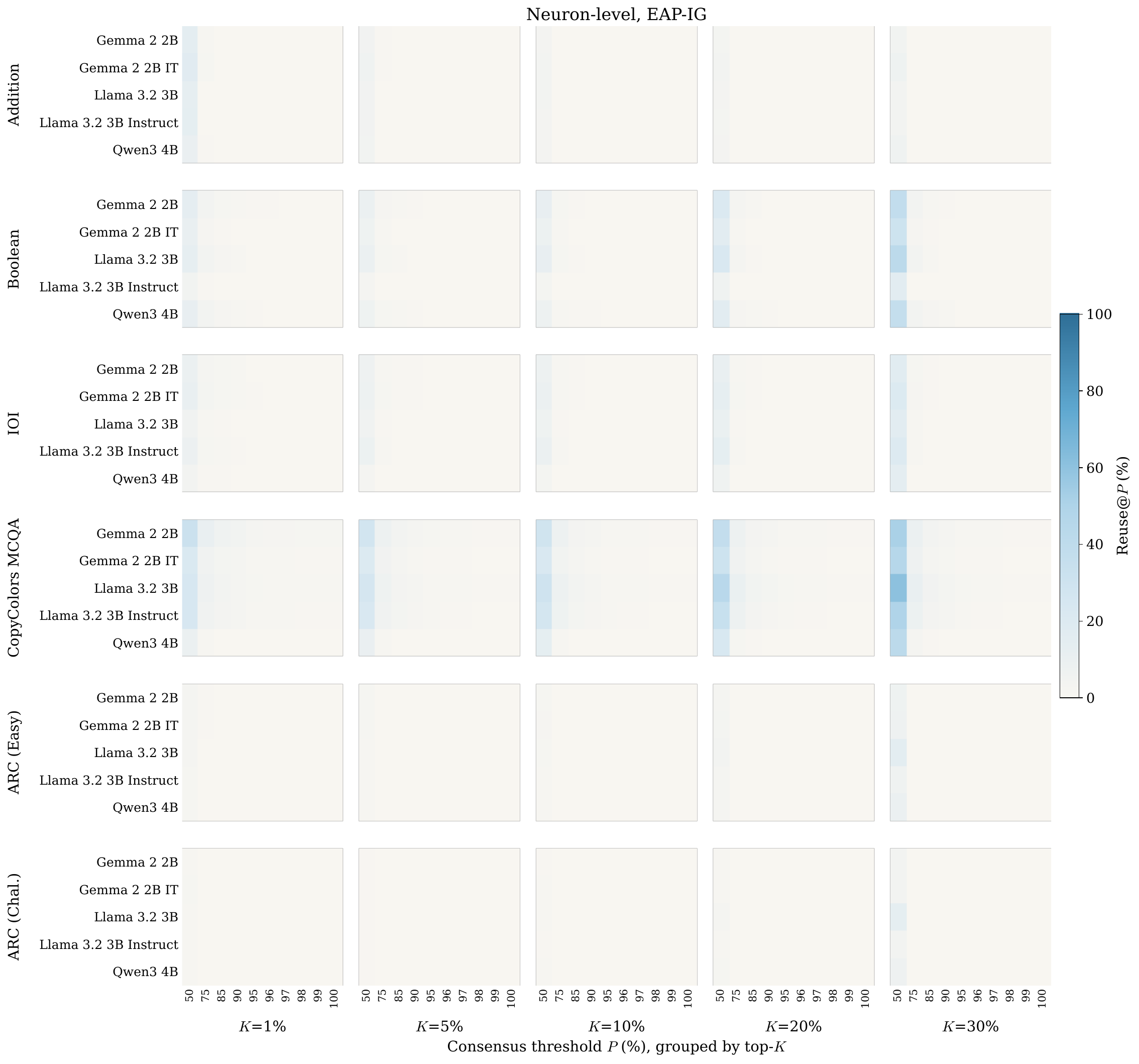}
    \caption{Within-task~\reuseat{P} of neuron-level circuits under EAP-IG over the full grid of top-$K$ and consensus threshold $P$. Columns are grouped by top-$K$, and rows are grouped by task, with one row per model. A value of 0 includes settings where the task circuit is empty.}
    \label{fig:reuse_all_eap_ig_neuron}
\end{figure*}

\begin{figure*}[htbp]
    \centering
    \includegraphics[width=\textwidth]{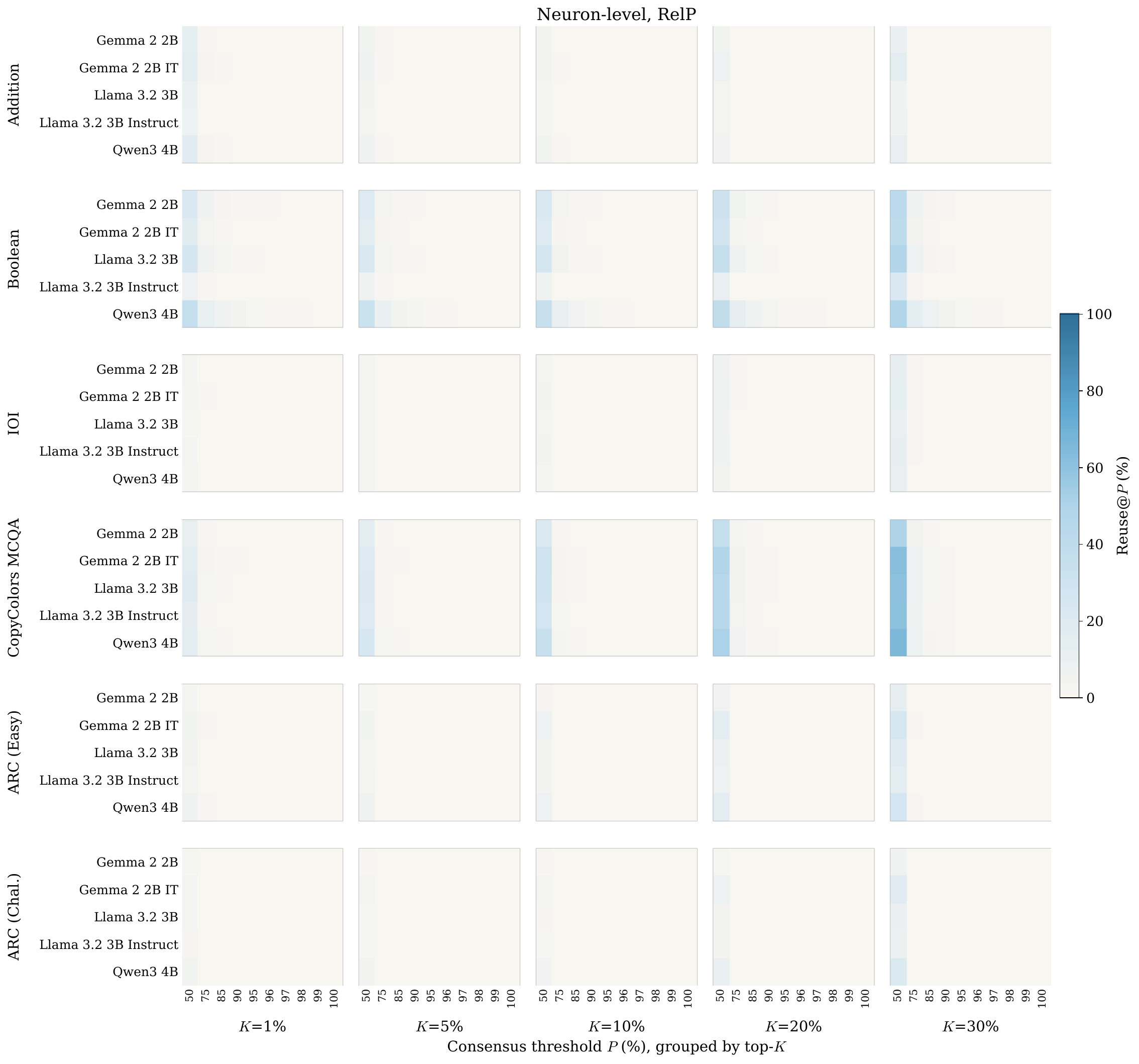}
    \caption{Within-task~\reuseat{P} of neuron-level circuits under RelP over the full grid of top-$K$ and consensus threshold $P$. Columns are grouped by top-$K$, and rows are grouped by task, with one row per model. A value of 0 includes settings where the task circuit is empty.}
    \label{fig:reuse_all_relp_neuron}
\end{figure*}

\begin{figure*}[htbp]
    \centering
    \includegraphics[width=\textwidth]{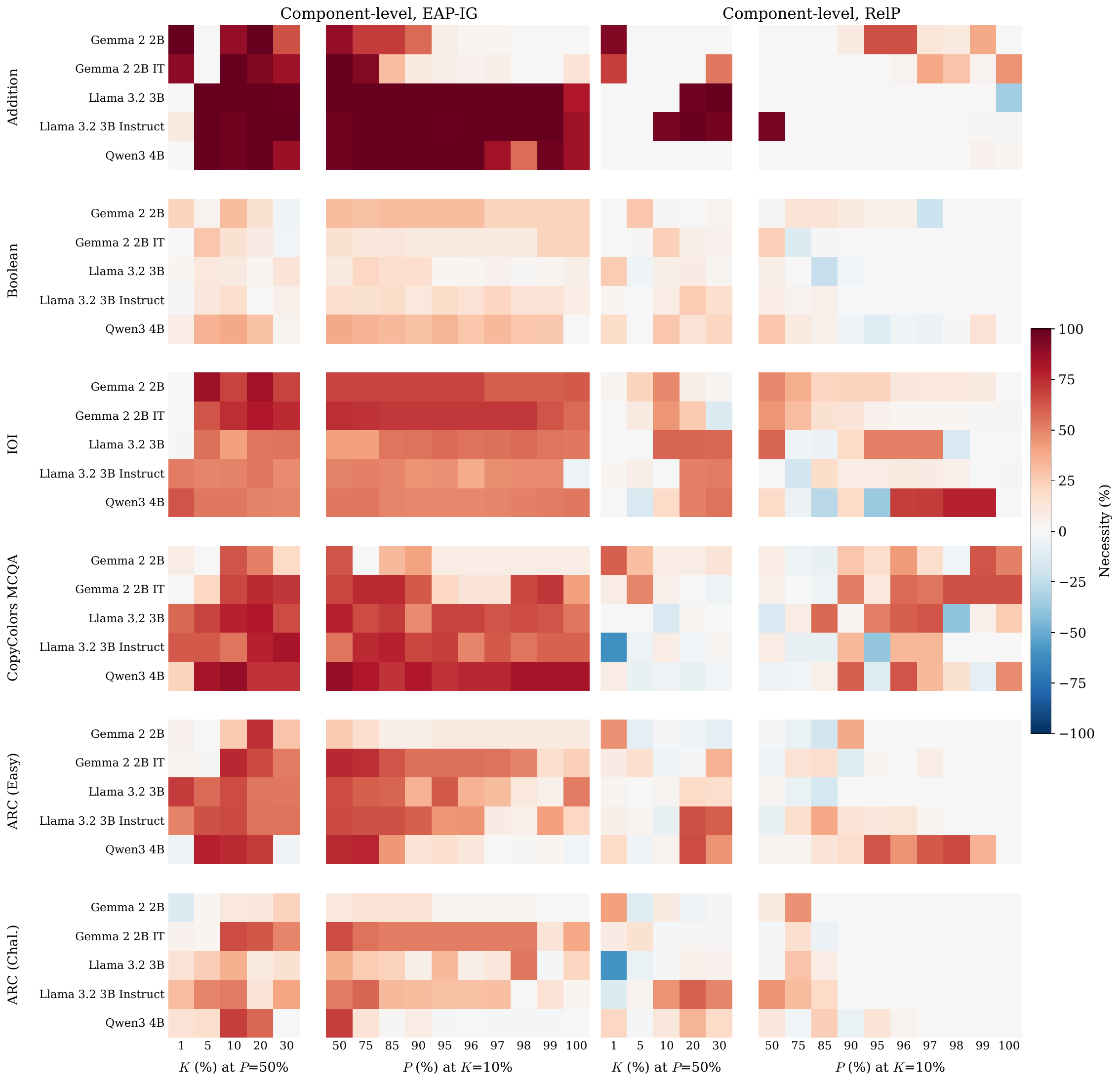}
    \caption{Within-task~\causalmetric~of component-level circuits under EAP-IG (left) and RelP (right), along top-$K$ at $P=50\%$ and along $P$ at $K=10\%$. Rows are grouped by task, with one row per model.}
    \label{fig:necessity_sweep_head_mlp}
\end{figure*}

\begin{figure*}[htbp]
    \centering
    \includegraphics[width=\textwidth]{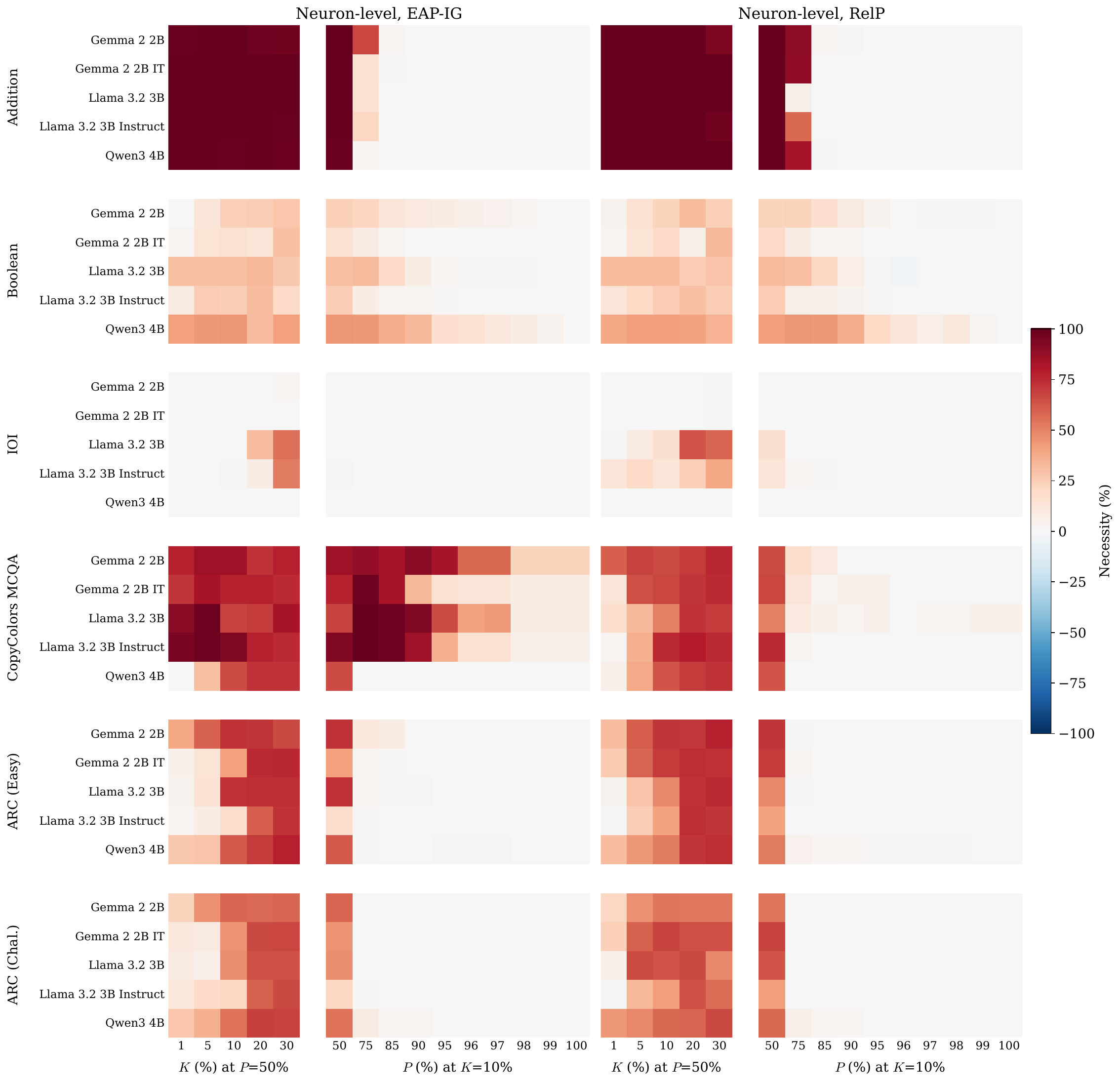}
    \caption{Within-task~\causalmetric~of neuron-level circuits under EAP-IG (left) and RelP (right), along top-$K$ at $P=50\%$ and along $P$ at $K=10\%$. Rows are grouped by task, with one row per model.}
    \label{fig:necessity_sweep_neuron}
\end{figure*}

We report~\reuseat{P} across the full sweep of top-$K$ and consensus threshold $P$ for every model and task, separately for each attribution method and granularity.
\Cref{fig:reuse_sweep} summarizes the effect of varying $K$ and $P$ with the median over models and tasks, while
\cref{fig:reuse_all_eap_ig_head_mlp,fig:reuse_all_relp_head_mlp,fig:reuse_all_eap_ig_neuron,fig:reuse_all_relp_neuron}
show the full $K\times P$ sweeps for each extraction configuration.
\Cref{fig:necessity_sweep_head_mlp,fig:necessity_sweep_neuron}
give \causalmetric~along top-$K$ at $P$=50\% and along $P$ at $K$=10\%, the two sweeps on which we measure it.

\clearpage
\section{Full Cross-Task Results}
\label{sec:full_cross}

\begin{figure*}[htbp]
  \centering
  \includegraphics[width=\textwidth]{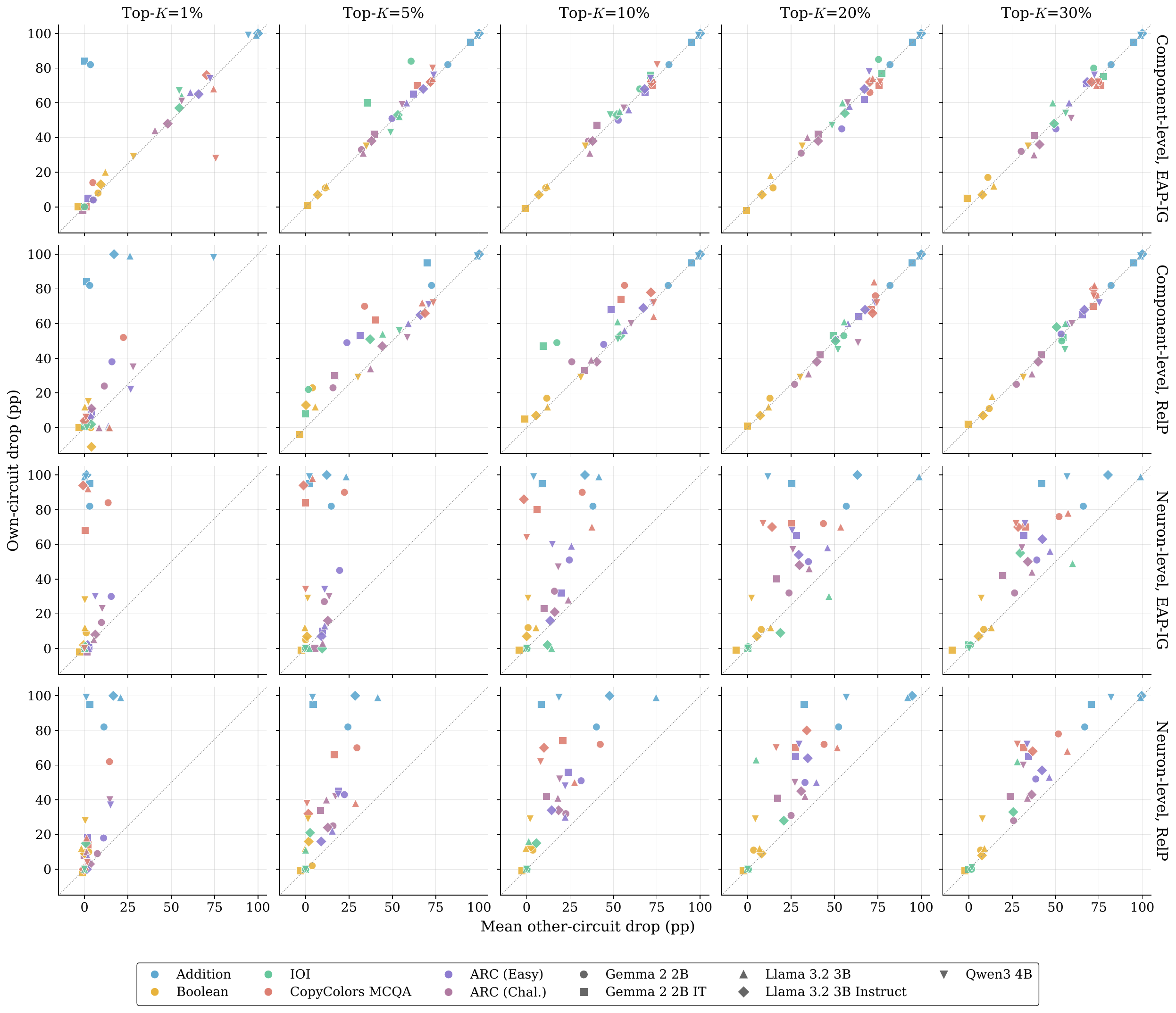}
  \caption{\textbf{Own-circuit against mean other-circuit accuracy drop at every top-$K$, $P$=50\%.} Each point is one model (marker) and task (color). Points above the dotted diagonal are tasks damaged more by their own circuit than by the other tasks' circuits. Rows are the attribution method and granularity, and columns are top-$K$.}
  \label{fig:own_other_all}
\end{figure*}

In \cref{fig:own_other_all} we show the accuracy drop on each task from ablating its own circuit against the mean drop from ablating the other five tasks' circuits, for every model, task and top-$K$ at $P$=50\%, separately for each attribution method and granularity.
Component-level circuits lie close to the diagonal from $K$=5\% upwards, except for \texttt{IOI} in the \texttt{Gemma-2} models at $K$=5\% under EAP-IG and the RelP circuits of the \texttt{Gemma-2} models at $K$=5 and 10\%, while at $K$=1\% the circuit of \texttt{Addition} is specific in most models.
Neuron-level circuits lie above the diagonal at every $K$, although less so as $K$ grows, with the mean specificity gap falling from 29.4 (EAP-IG) and 26.8 (RelP) points at $K$=10\% to 18.1 and 14.8 points at $K$=30\%.

\begin{figure*}[htbp]
  \centering
  \includegraphics[width=\textwidth]{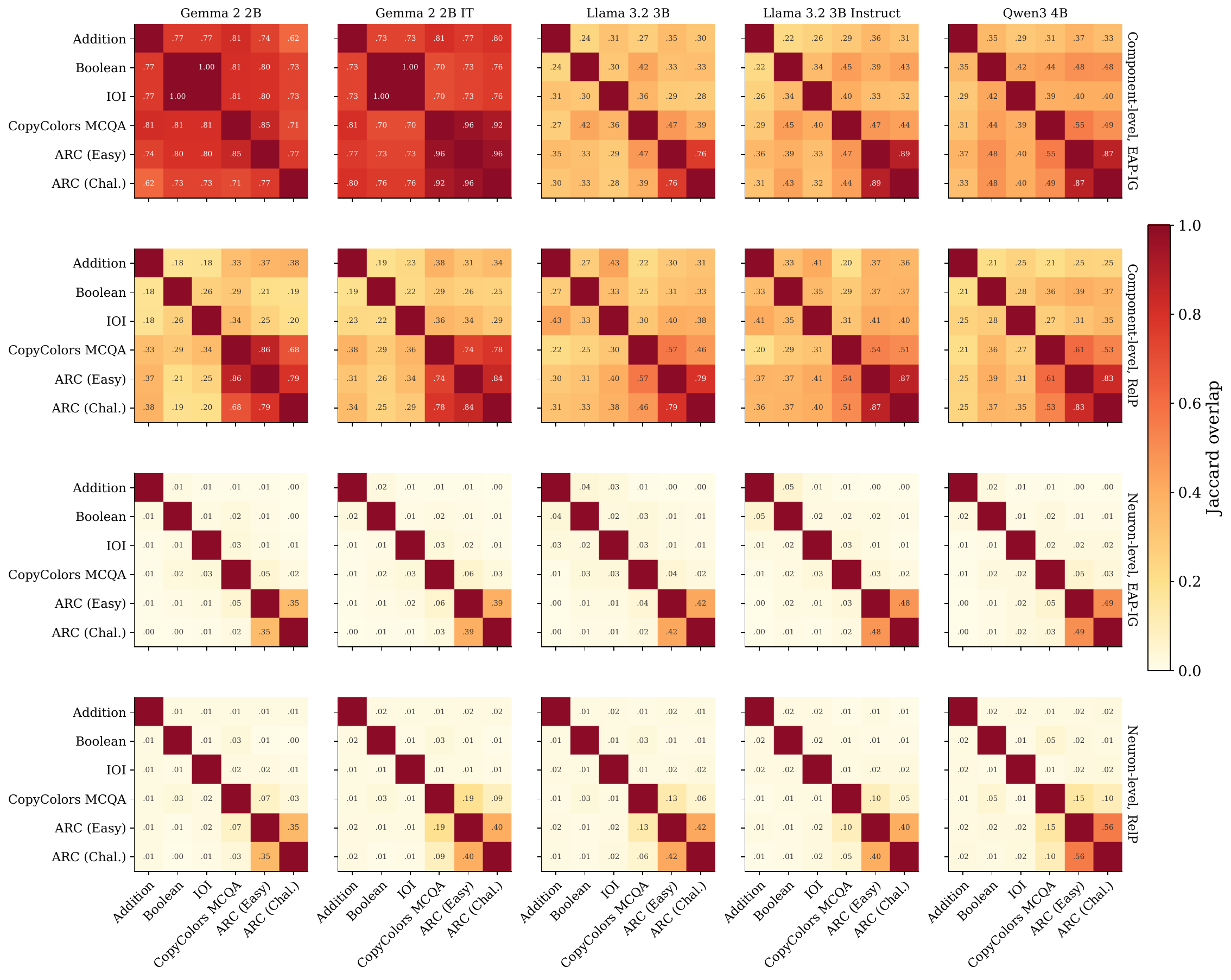}
  \caption{\textbf{Jaccard overlap between task circuits for each model at $K$=10\%, $P$=50\%.} Rows are the attribution method and granularity, and columns are models.}
  \label{fig:overlap_per_model}
\end{figure*}

In \cref{fig:overlap_per_model} we show the Jaccard overlap between every pair of task circuits for each model at $K$=10\% and $P$=50\%, the per-model version of \cref{fig:overlap_own_other} (top).
We observe that component-level circuits overlap substantially in every model, most of all under EAP-IG in the \texttt{Gemma-2} models, where any two tasks' circuits share most of their components (Jaccard overlap of 0.62 to 1.00, with identical circuits for \texttt{Boolean~Logic} and \texttt{IOI}).
Overlap is also consistently highest between related tasks, notably the two ARC tasks (0.76 to 0.96 at the component level) and, under RelP, \texttt{CopyColors~MCQA} and the ARC tasks (0.46 to 0.86), which share the multiple-choice format.
Neuron-level circuits, in contrast, barely overlap in any model (at most 0.19), with the two ARC tasks again the only exception (0.35 to 0.56).

\begin{figure*}[htbp]
  \centering
  \includegraphics[width=\textwidth]{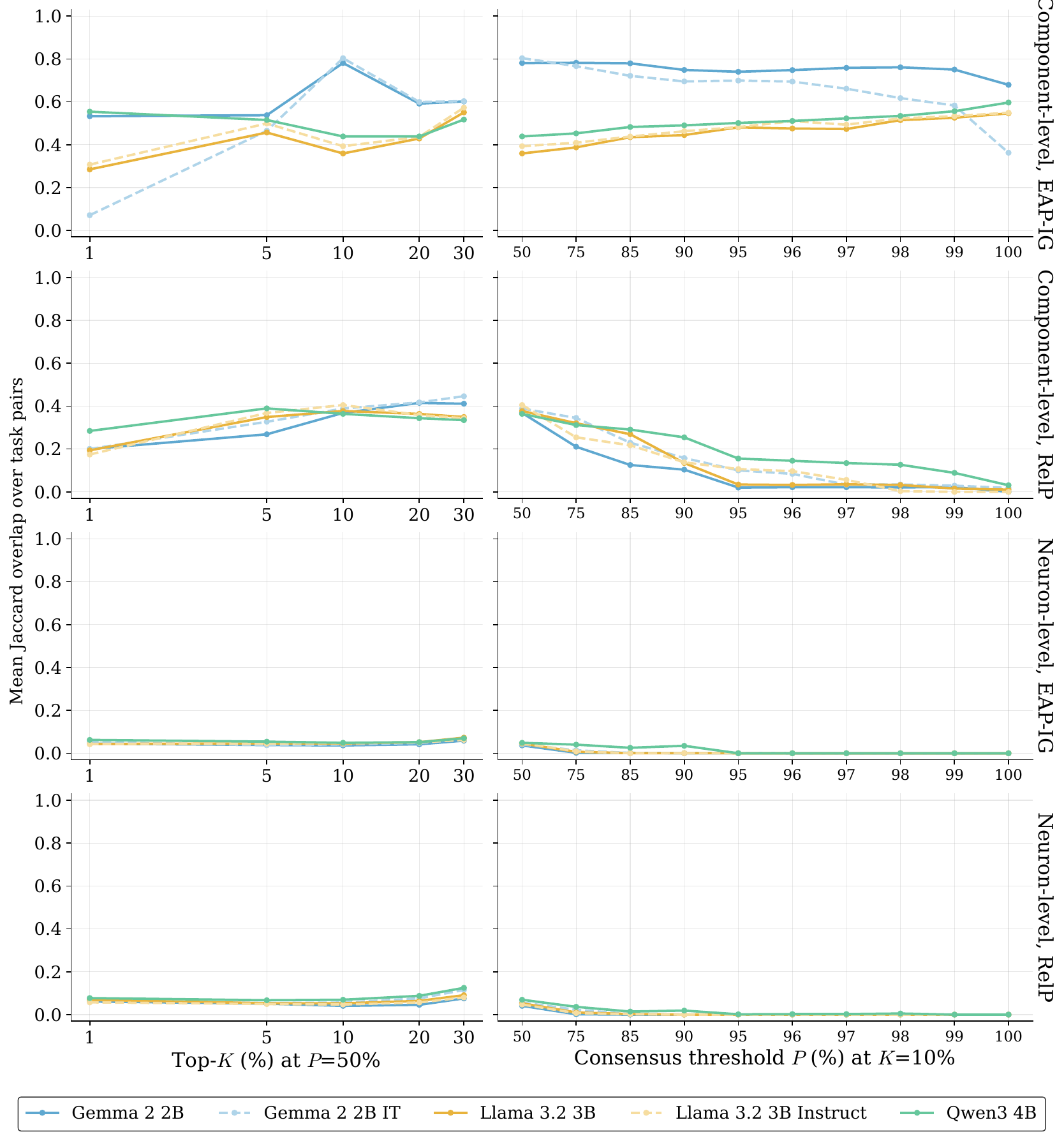}
  \caption{\textbf{Mean Jaccard overlap over the 15 task pairs for each model}, along top-$K$ at $P$=50\% (left) and along $P$ at $K$=10\% (right). Instruction-tuned models are dashed. Pairs where both circuits are empty are left out.}
  \label{fig:overlap_sweeps}
\end{figure*}

\cref{fig:overlap_sweeps} shows how the mean overlap over task pairs changes along top-$K$ at $P$=50\% and along $P$ at $K$=10\%, to check that these trends are not specific to one circuit size.
The overlap of component-level EAP-IG circuits is stable across both sweeps (0.36 to 0.80 for $K\geq5\%$), with the exception of \texttt{Gemma-2-2B-IT} at $P$=100\% (0.36).
RelP overlap, however, shrinks as $P$ grows, from about 0.38 at $P$=50\% to nearly zero at $P$=100\%, as its task circuits shrink (\cref{fig:circuit_size}).
Neuron-level overlap stays low at every $K$ and $P$ (at most 0.12).

\end{document}